\documentclass[fleqn,11pt]{wlscirep}
\usepackage[utf8]{inputenc}
\usepackage[T1]{fontenc}
\usepackage{textcomp}
\usepackage{marvosym}
\usepackage{longtable}
\usepackage{placeins}
\usepackage{bbm}
\usepackage{graphicx, caption, subcaption}
\usepackage{booktabs}
\usepackage{multirow}
\usepackage{listings}
\usepackage{xcolor}
\newcommand{\rev}[1]{\textcolor{black}{#1}} 
\newcommand{\nrev}[1]{\textcolor{black}{#1}} 
\title{Scaling Recurrence-aware Foundation Models for Clinical Records via Next-Visit Prediction}

\author[1,†]{Xiang Gao}
\author[1,†]{Haresh Rengaraj Rajamohan}
\author[1]{Weicheng Zhu}
\author[1]{Shih-Lun Huang}
\author[1]{Long Chen}
\author[1]{Gabe Schulman}
\author[1]{Huizhen Jin}
\author[1]{Shengduo Li}
\author[1]{Yixuan Wang}
\author[1]{Huidi Yang}
\author[1]{Kyunghyun Cho}
\author[2,4]{Cem M. Deniz}
\author[2,3,*]{Narges Razavian}
\affil[1]{Center for Data Science, New York University, New York, NY, USA}
\affil[2]{Department of Radiology, NYU Grossman School of Medicine, New York, NY, USA}
\affil[3]{Department of Population Health, NYU Grossman School of Medicine, New York, NY, USA}
\affil[4]{Center for Biomedical Imaging, NYU Grossman School of Medicine, New York, NY, USA}
\affil[*]{\Letter~Corresponding author: narges.razavian@nyulangone.org}
\affil[$\dagger$]{These authors contributed equally to this work}

\begin{abstract}
While large-scale pretraining has revolutionized language modeling, its potential remains underexplored in healthcare with structured electronic health records (EHRs). We present RAVEN, a novel generative pretraining strategy for sequential EHR data based on Recurrence-Aware next-Visit EveNt prediction. Leveraging a dataset of over one million unique individuals, our model learns to autoregressively generate tokenized clinical events for the next visit conditioned on patient history. We introduce regularization on predicting repeated events and highlight a key pitfall in EHR-based foundation model evaluations: repeated event tokens can inflate performance metrics when new onsets are not distinguished from subsequent occurrences. Furthermore, we empirically investigate the scaling behaviors in a data-constrained, compute-saturated regime, showing that simply increasing model size is suboptimal without commensurate increases in data volume. We evaluate our model via zero-shot prediction for forecasting the incidence of a diverse set of diseases, where it rivals fully fine-tuned representation-based Transformer models and outperforms both standard simulation-based next-token approaches and a prompted medical large language model baseline. Finally, without additional parameter updates, we show that RAVEN can generalize to an external patient cohort under lossy clinical code mappings and feature coverage gaps.

\end{abstract}

\begin{document}

\flushbottom
\maketitle
\thispagestyle{empty}

\newpage

\section*{Introduction}

Early detection and progression forecasting for chronic conditions such as cardiovascular and cerebrovascular events, diabetes complications, cognitive impairment, osteoarthritis, cancer, among others, can significantly improve healthcare outcomes and further optimize clinical trial design\cite{dubois2015timely,zhu2024predicting,arnold2022current,karsdal2016disease}. Electronic health records (EHRs), which are readily available at point of care and increasingly accessible directly to patients, provide actionable information at inference time and abundant longitudinal data for training. But they present challenges to modeling because they are sequential, high-dimensional, irregularly sampled, and heterogeneous across diagnoses, medications, laboratory measurements, and demographics \cite{nordo2019use,choi2017using,xiao2018opportunities,shickel2017deep}. 

Despite the rapid developments in artificial intelligence, many clinical prediction pipelines still train disease-specific discriminative models for a fixed endpoint and horizon. This approach is costly to scale, statistically inefficient, and poorly matched to health-system deployment, where many outcomes must be monitored simultaneously. Such models repeatedly learn representations from limited task-specific supervision with the same underlying longitudinal patient data \cite{rajkomar2018scalable}; therefore, they fail to exploit shared temporal structures across clinical outcomes. This inefficiency is pronounced for long-horizon disease-onset prediction tasks, where the sample size of disease-specific cohorts is small by construction due to the need for extended follow-up, increasing the risk of overfitting and unstable generalization. By discretizing and tokenizing clinical concepts, foundation models offer a more attractive alternative because they can learn a unified representation of longitudinal EHR data and capture the full joint distribution of clinical events over time, enabling flexible transfer across diseases, endpoints, and prediction horizons \cite{renc2024zero, pang2025cehr}. 

Prior work in training foundation models on EHR data with masked pretraining or encoder-decoder architecture requires an additional fine-tuning stage for adapting the model to particular downstream tasks \cite{li2020behrt, pang2021cehr, yang2023transformehr}. This requires curating additional task-specific datasets. More recently, generative pretraining for EHRs has gained traction due to success in other data modalities and the flexibility for zero-shot inference \cite{mcdermott2023event, renc2024zero, pang2025cehr, waxler2025generative}. However, many of these approaches are based on the next-token paradigm or trained on specific clinical records like those from intensive care units (ICU) \cite{renc2024zero}, which fail to demonstrate the challenges we outline below in learning longitudinal records from clinical visits. In this work, we propose a novel approach for developing and evaluating foundation models in the space of longitudinal EHR called the Recurrence-Aware next-Visit EveNt prediction (RAVEN).

One of the key challenges in modeling longitudinal EHR data with foundation models is that clinical events that occurred within a single patient visit lack fine-grained temporal order, making the standard next-token prediction insufficient. We believe that next-visit event prediction via multi-label generations provides a natural and scalable pretraining objective for structured EHRs. Longitudinal clinical data are inherently organized around visits, within which events are usually unordered but semantically related, while meaningful temporal structures exist across visits. Predicting the full set of events at the next encounter allows RAVEN to reason jointly over heterogeneous clinical variables without imposing an arbitrary within-visit token order, and it enables efficient downstream inference through a single forward pass compared to simulation-based inference.

Moreover, unlike text or image tokens, recurring clinical event tokens, especially chronic conditions, can reappear throughout a patient’s record. Once diagnoses such as hypertension or diabetes appear, they often reoccur in subsequent visits. Hence, the marginal distribution of clinical events is highly skewed toward repeated tokens, with new disease onsets constituting a relatively rare but clinically critical subset of events. When training generative models with the naive next-visit objective, the model can achieve lower loss by simply repeating a condition at future visits once a chronic condition appears in the patient history. 
In particular, a model that merely echoes previously observed diagnoses may score well on aggregate metrics \cite{kraljevic2024foresight}, yet degrades its performance on predicting new onsets for early disease detection or risk forecasting.

\begin{figure}[ht]
\centering
\includegraphics[width=\linewidth,trim=27bp 30bp 27bp 33bp,clip]{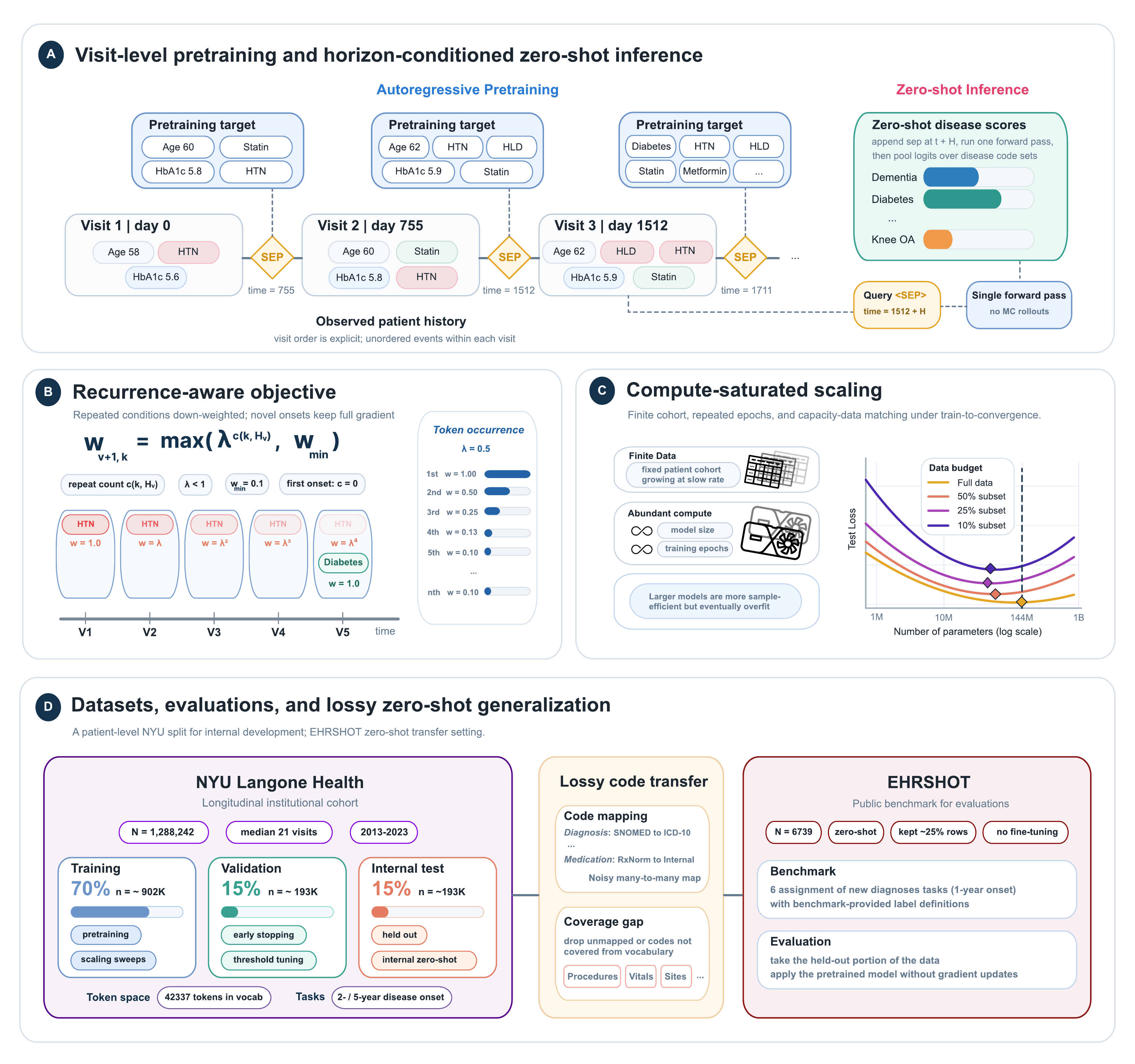}
\caption[Study overview of recurrence-aware next-visit foundation modeling]{\textbf{Study overview of recurrence-aware next-visit foundation modeling on longitudinal EHRs.} \textbf{a,} Patient trajectories are represented as temporally ordered visits containing unordered clinical events. During pretraining, a time-coded separator token \texttt{[SEP]} predicts the full event set at the next visit; during zero-shot inference, the same interface queries disease risk at future horizons by appending a separator token at time $t+H$ and pooling logits over condition-specific code sets. \textbf{b,} Repeated chronic targets are downweighted during pretraining according to their prior count so that rare first onsets retain training signal. \textbf{c,} In a finite EHR corpus revisited for many epochs, model size must match the corresponding available data as larger models eventually overfit, and the selected 144M model lies near the full-data optimum. \textbf{d,} Model development uses a patient-level 70\%/15\%/15\% split on patient data from NYU Langone; a schematic external zero-shot transfer setting based on Stanford EHRSHOT highlights ontology harmonization between benchmark concepts and the institutional token space, which can introduce information loss and drop of certain features.}
\label{fig:pretraining_zeroshot}
\end{figure}
\FloatBarrier

To address this, we equip RAVEN with a history-dependent regularization mechanism that explicitly downweights the contribution of repeatedly observed clinical events during pretraining. By penalizing predictions of tokens in proportion to their frequency in a patient's history, we encourage the model to focus on informative changes in the clinical trajectory rather than simple repetitions. This inductive bias promotes the ability to predict emerging conditions while preserving the capacity to model chronic disease persistence, leading to models that better align with clinically meaningful disease onset prediction tasks.

We carefully investigate how RAVEN performs with different choices of model and sample sizes. Most scaling laws for foundation models, including large language models (LLMs) and EHR foundation models, are often studied under compute-constrained regimes by assuming limited compute and near-infinite data source, where models are trained for a fixed compute budget which then determines the model size and the number of tokens for a single training pass \cite{kaplan2020scaling, hoffmann2022training, zhang2025exploring, waxler2025generative}. In such settings, they provide meaningful guidance on how to optimally trade off model size and data exposure under a fixed compute budget; others have also shown data-constrained scaling on the relationship between model size and the number of repeated training runs with a fixed compute budget \cite{muennighoff2023scaling}. 

Nevertheless, they do not fully reflect the nature of modeling structured EHR data, where the total number of patient trajectories is orders of magnitude smaller than the internet text tokens used in language modeling and repeated training on the same records is inevitable. Hence, similar to recent work in LLMs \cite{kim2026pre}, where the concern for running out of data is rising, we study the scaling behaviors of EHR foundation models in a data-limited, compute-saturated regime. We assume that we have infinite compute and limited data, where compute is more naturally spent on optimizing convergence of large models over the available data. We train RAVEN under varying parameter counts and data budgets for multiple epochs until validation convergence. We find that the model size eventually becomes too large to avoid overfitting, and model capacity must be matched to the available cohort size.

In this work, RAVEN autoregressively learns the joint state of medications, labs, and diagnoses for the next encounter, conditioned on patient history and the time of the next visit. Also, we show how repeated event tokens can inflate model performance through pretraining evaluations and propose recurrence-aware regularization for repeated events to encourage the learning of new events. We evaluate RAVEN with rigorously constructed downstream disease onset tasks and demonstrate strong zero-shot generalization in forecasting different types of diseases with varying prediction horizons, compared to a fully fine-tuned BERT-based foundation model baseline, several popular next-token-based approaches, and a prompted medical LLM. Additionally, we apply RAVEN pretrained on our own dataset to EHRSHOT, a public benchmark of longitudinal EHR sourced from a different health system \cite{wornow2023ehrshot}. Despite the coding differences for clinical concepts and gaps for coverage of features between the health systems, we showcase RAVEN's ability to generalize beyond its training distribution without additional few-shot samples.

\section*{Results}

We developed RAVEN, a generative foundation model for structured longitudinal electronic health records that learns to predict the full set of clinical events at a patient's next visit, conditioned on their prior history and the timing of the encounter (Fig.~\ref{fig:pretraining_zeroshot}a). The model was pretrained on de-identified EHR trajectories from approximately 1.29 million patients at NYU Langone Health, spanning a decade of inpatient and outpatient encounters (2013–2023) and comprising over 42,000 unique clinical tokens across diagnoses, medications, laboratory results, and demographics. All data splits were performed at the patient level to prevent information leakage, yielding training (70\%, n = 901,769), validation (15\%, n = 193,236), and test (15\%, n = 193,237) sets. To address the tendency of models to rely on repeating conditions rather than learning to anticipate new disease onsets, we introduced a recurrence-aware regularization mechanism that downweights recurrent events during training (Fig.~\ref{fig:pretraining_zeroshot}b). We systematically investigated model scaling under a data-constrained, compute-saturated regime to identify the capacity–data trade-off appropriate for finite clinical cohorts (Fig.~\ref{fig:pretraining_zeroshot}c). We evaluated RAVEN's ability to perform zero-shot disease-onset forecasting at multiple horizons on held-out internal data, and further tested its generalization capability through zero-shot transfer to EHRSHOT, a public benchmark derived from a separate health system with different coding conventions and feature coverage (Fig.~\ref{fig:pretraining_zeroshot}d).

\subsection*{Zero-shot disease forecasting}
Without any task-specific fine-tuning, we evaluated the zero-shot ability of RAVEN to forecast future long-term disease incidence across seven diverse clinical conditions: dementia, knee osteoarthritis (OA), pancreatic cancer, prostate cancer, acute myocardial infarction (MI), congestive heart failure (CHF), and chronic obstructive pulmonary disease (COPD). The ground truth label was defined by the first occurrence of the condition, which is determined by a set of diagnosis and medication codes within the prediction window. For each condition, we constructed evaluation examples using rolling prediction windows over patient trajectories: a past input history serves as context, and the model predicts whether the first onset of the condition occurs within the specified horizon. Here, we chose to evaluate at both 2-year and 5-year horizons. To ensure the model is predicting future new onset based on prior history, rather than simply repeating the conditions if already present, we exclude prediction windows if their input history windows contain any label code, or if the disease onset occurred within one year following the prediction time point. This ensures the model forecasts medium- to long-term risk rather than imminent events and reduces the risks of label leakage.

We benchmarked RAVEN against three variants of popular autoregressive next-token baselines that estimate risk via multiple simulations of future trajectories. Particularly, we include a standard next-token cross-entropy model (Multiclass), a next-token set-based loss model (SeqLoss), and a joint gap-and-event generation model (EGE). For these baselines, long-horizon risk is estimated via inference autoregressively with $R{=}\rev{64}$ rollouts per patient window, whereas RAVEN uses a single forward pass per example. Additionally, we evaluate MedGemma-27B \cite{sellergren2026medgemma} as a zero-shot prompted large language model baseline that operates directly on serialized patient histories represented as text, providing a comparison against general-purpose medical LLMs. We also compare against a strong supervised baseline via a BERT-based foundation model pretrained using the same training data \cite{zhu2024predicting} but also fine-tuned for each particular condition and horizon. Full details of the evaluation protocol, including endpoint definitions and baseline descriptions, are provided in the Methods section.

\begin{table}[ht]
\centering
\caption[Zero-shot forecasting at the 2-year horizon]{\textbf{Zero-shot forecasting at the 2-year horizon (AUROC / AUPRC).} All methods except BERT (FT) operate without task-specific fine-tuning. RAVEN uses a single forward pass; simulation-based models aggregate $R{=}\rev{64}$ rollouts. Baseline prevalence levels and task-specific statistics can be found in the Supplementary Table \ref{tab:eval_cohort_stats}. Bold denotes the best AUROC per condition.}
\label{tab:zs_2y_all_baselines}
\small
\setlength{\tabcolsep}{3.5pt}
\resizebox{\linewidth}{!}{%
\begin{tabular}{lcccccc}
\toprule
\textbf{Condition} & \textbf{RAVEN} & \textbf{Multiclass} & \textbf{SeqLoss} & \textbf{EGE} & \textbf{MedGemma} & \textbf{BERT (FT)} \\
\midrule
Dementia          & \textbf{0.789 / 0.037} & 0.687 / 0.026 & 0.704 / 0.031 & 0.677 / 0.026 & 0.620 / 0.011 & 0.731 / 0.050 \\
Knee OA           & 0.726 / 0.057          & 0.627 / 0.039 & 0.672 / 0.053 & 0.606 / 0.038 & 0.642 / 0.039 & \textbf{0.744 / 0.064} \\
COPD              & 0.691 / 0.052          & 0.544 / 0.036 & 0.629 / 0.044 & 0.602 / 0.040 & 0.589 / 0.030 & \textbf{0.704 / 0.055} \\
CHF               & 0.857 / 0.078          & 0.708 / 0.047 & 0.733 / 0.048 & 0.656 / 0.045 & 0.736 / 0.036 & \textbf{0.862 / 0.085} \\
Acute MI          & 0.793 / 0.031          & 0.603 / 0.015 & 0.579 / 0.013 & 0.604 / 0.017 & 0.748 / 0.026 & \textbf{0.818 / 0.049} \\
Pancreatic Cancer & \textbf{0.693 / 0.002} & 0.496 / 0.000 & 0.494 / 0.000 & 0.502 / 0.001 & 0.620 / 0.001 & 0.607 / 0.001 \\
Prostate Cancer   & \textbf{0.910 / 0.022} & 0.828 / 0.018 & 0.820 / 0.017 & 0.736 / 0.010 & 0.659 / 0.005 & 0.905 / 0.021 \\
\midrule
\textbf{Macro average} & \textbf{0.780 / 0.040} & 0.642 / 0.026 & 0.662 / 0.029 & 0.626 / 0.025 & 0.659 / 0.021 & 0.767 / 0.046 \\
\bottomrule
\end{tabular}%
}
\end{table}

\begin{table}[ht]
\centering
\caption[Zero-shot forecasting at the 5-year horizon]{\textbf{Zero-shot forecasting at the 5-year horizon (AUROC / AUPRC).} The same evaluation protocol is used as in Table~\ref{tab:zs_2y_all_baselines}. Bold denotes the best AUROC per condition.}
\label{tab:zs_5y_all_baselines}
\small
\setlength{\tabcolsep}{3.5pt}
\resizebox{\linewidth}{!}{%
\begin{tabular}{lcccccc}
\toprule
\textbf{Condition} & \textbf{RAVEN} & \textbf{Multiclass} & \textbf{SeqLoss} & \textbf{EGE} & \textbf{MedGemma} & \textbf{BERT (FT)} \\
\midrule
Dementia          & \textbf{0.773 / 0.108} & 0.669 / 0.089 & 0.693 / 0.102 & 0.670 / 0.088 & 0.609 / 0.041 & 0.721 / 0.146 \\
Knee OA           & 0.697 / 0.200          & 0.635 / 0.166 & 0.671 / 0.190 & 0.620 / 0.166 & 0.617 / 0.146 & \textbf{0.718 / 0.221} \\
COPD              & 0.676 / 0.164          & 0.531 / 0.124 & 0.640 / 0.169 & 0.615 / 0.149 & 0.586 / 0.115 & \textbf{0.699 / 0.189} \\
CHF               & 0.821 / 0.184          & 0.718 / 0.114 & 0.723 / 0.134 & 0.666 / 0.127 & 0.718 / 0.101 & \textbf{0.838 / 0.220} \\
Acute MI          & 0.754 / 0.080          & 0.613 / 0.060 & 0.569 / 0.047 & 0.623 / 0.059 & 0.699 / 0.067 & \textbf{0.782 / 0.106} \\
Pancreatic Cancer & 0.639 / 0.004          & 0.516 / 0.002 & 0.513 / 0.004 & 0.498 / 0.001 & 0.597 / 0.002 & \textbf{0.642 / 0.003} \\
Prostate Cancer   & \textbf{0.901 / 0.073} & 0.784 / 0.061 & 0.803 / 0.049 & 0.779 / 0.045 & 0.642 / 0.017 & 0.896 / 0.065 \\
\midrule
\textbf{Macro average} & 0.752 / 0.116 & 0.638 / 0.088 & 0.659 / 0.099 & 0.639 / 0.091 & 0.638 / 0.070 & \textbf{0.757 / 0.136} \\
\bottomrule
\end{tabular}%
}
\end{table}

Tables~\ref{tab:zs_2y_all_baselines} and~\ref{tab:zs_5y_all_baselines} report zero-shot performance at 2-year and 5-year horizons, respectively. All results in this section use the regularized RAVEN ($\lambda^\star = 0.5$) with 144 million parameters; the effect of regularization strength is examined later in the Recurrence-aware regularization section. 
We provide 95\% confidence intervals for all results in the Supplementary Tables \ref{tab:supp_zs_2y_auroc_ci} to \ref{tab:supp_zs_5y_auprc_ci}. Across conditions, RAVEN achieves the highest AUROC among the structured zero-shot approaches at both horizons. 
At 2 years (Table~\ref{tab:zs_2y_all_baselines}), RAVEN attains a macro-average AUROC of 0.780, substantially exceeding Multiclass (0.642), SeqLoss (0.662), EGE (0.626), and MedGemma (0.659). For example, RAVEN improves over the best next-token baseline for dementia (AUROC 0.789 vs.\ 0.704) and CHF (0.857 vs.\ 0.733).
A similar pattern holds at 5 years (Table~\ref{tab:zs_5y_all_baselines}), where RAVEN achieves a macro-average AUROC of 0.752 compared to 0.638, 0.659, 0.639, and 0.638 for Multiclass, SeqLoss, EGE, and MedGemma, respectively. \rev{The advantage over the rollout-based models does not stem from an insufficient sampling budget: varying the number of rollouts from 4 to 256 shows that their performance saturates near $R{=}64$, consistent with independent reports for simulation-based medical event models \cite{waxler2025generative}, and does not close the gap to RAVEN on any condition (Supplementary Fig.~\ref{fig:rollout_ablation}).} Although MedGemma shows non-trivial discrimination on several tasks, notably acute MI (AUROC 0.748 at 2 years) and CHF (0.736), its performance is generally weaker than RAVEN, suggesting that in-context inference over serialized records represented with text is not a sufficient substitute for structured longitudinal pretraining.

For dementia and prostate cancer, RAVEN achieves the highest AUROC across all methods at both horizons, matching or exceeding even the fine-tuned BERT baseline. For pancreatic cancer, RAVEN outperforms BERT at 2 years (0.693 vs.\ 0.607) and is comparable at 5 years (0.639 vs.\ 0.642), despite the extremely low condition prevalence reflected in AUPRC values below 0.005 in the 2-year window.  The fine-tuned BERT baseline achieves the highest AUROC on CHF (0.862 at 2 years, 0.838 at 5 years), knee OA, and COPD, where task-specific supervision provides an advantage. However, the gap between our zero-shot model and fully fine-tuned BERT is modest in most cases (e.g., COPD 2-year AUROC 0.691 vs.\ 0.704; knee OA 0.726 vs.\ 0.744), and RAVEN consistently outperforms all rollout-based autoregressive baselines and MedGemma across conditions and horizons. At the macro-average level, RAVEN (0.780 / 0.752 at 2 / 5 years) is competitive with BERT (0.767 / 0.757) while requiring no task-specific training.

\subsection*{\rev{Clinical utility and robustness}}

\rev{A potential concern in zero-shot onset prediction is that apparent discrimination may be driven primarily by demographic or cohort-selection effects rather than by longitudinal information learned from the clinical record. To address this, we added simple demographic and utilization baselines, including logistic regression (LR) and gradient boosting machine (GBM) classifiers using covariates such as age, sex, race and ethnicity, number of codes and visits, and other utilization measures, all computed strictly prior to the prediction time. We also evaluated endpoints on both the full evaluation population and a \nrev{demographically restricted} population under specific risk-set restrictions: prostate cancer was restricted to males, dementia to individuals aged $\geq65$ years, CHF and knee osteoarthritis to individuals aged $\geq45$ years, and COPD and acute myocardial infarction to individuals aged $\geq40$ years.}

\begin{figure}[h]
\centering
\includegraphics[width=\linewidth]{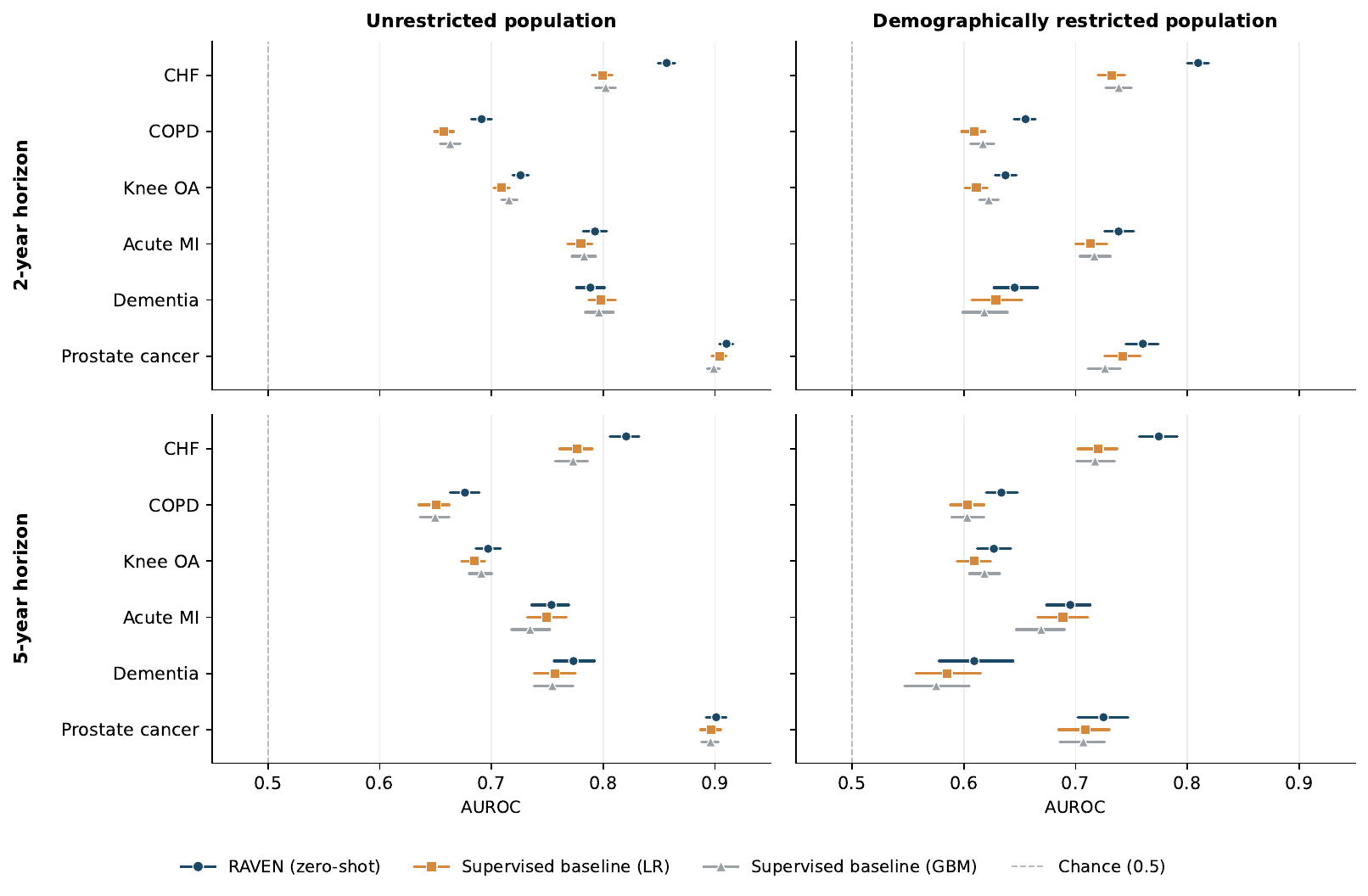}
\caption{\textbf{Comparison of RAVEN with supervised demographics baselines under unrestricted and demographically restricted populations (144M, $\lambda^\star=0.5$).} AUROC is shown across 2- and 5-year horizons. RAVEN is compared with LR and GBM baselines trained using simple covariates. Results are shown for the full evaluation population and for a demographically restricted population designed to reduce discrimination driven by simple factors. Points denote AUROC and horizontal bars denote 95\% patient-level bootstrap confidence intervals. The vertical dashed line indicates chance performance.}
\label{fig:baseline_restriction}
\end{figure}

\rev{RAVEN outperformed the strongest supervised baseline in most unrestricted comparisons, although the margin was smaller for tasks with strong demographic separability, such as prostate cancer. After \nrev{applying these demographic eligibility restrictions}, RAVEN achieved the highest AUROC in all 12 restricted settings, and the performance gap relative to the strongest baseline generally widened rather than narrowed (Fig.~\ref{fig:baseline_restriction}), indicating that RAVEN retains predictive signal beyond easily exploitable demographic separation. For example, for 2-year CHF prediction, RAVEN reached an AUROC of 0.810 compared with 0.739 for the best supervised baseline with simple covariates. In prostate cancer, where sex dominates prediction in the unrestricted population, RAVEN continued to outperform the supervised baselines after evaluation was restricted to males. The full age\,$\times$\,sex stratification showed the same pattern at finer resolution. RAVEN exceeded the best baseline in 73 of 88 evaluable strata, led by a uniform advantage on all 16 CHF strata (median $\Delta$AUROC $+0.113$) and consistent gains across the remaining endpoints; excluding the youngest age band ($<$40-year) gave the same qualitative conclusion, with RAVEN leading in 57 of 66 remaining strata. The complete stratified results are provided in Supplementary Table~\ref{tab:supp_stratified_age_sex_auroc_ci}.}

\rev{We further evaluated the model calibration, as well as how RAVEN performs at deployment-relevant operating points. 
The raw RAVEN scores showed meaningful risk-ranking ability but were not well calibrated as absolute probabilities. Raw Brier skill scores are negative for all settings, and reliability diagrams confirm miscalibration whose direction varies by condition (Supplementary Figs.~\ref{fig:supp_calib_2y}--\ref{fig:supp_calib_5y}; Table~\ref{tab:supp_calibration}). 
This is expected in our zero-shot setting: the model is not trained with endpoint-specific supervised labels, and the disease-specific scores are derived directly from pretrained event-prediction heads rather than from a task-specific calibration layer. Importantly, this miscalibration is largely correctable: we found that simple post-hoc recalibration substantially improves absolute-risk reliability while preserving discrimination.}

\rev{We fit Platt and isotonic recalibrations on the validation set and evaluated them on the held-out test set \cite{niculescu2005predicting}. Platt scaling reduced expected calibration error (ECE) by roughly an order of magnitude: the mean ECE moved from 0.0132 to 0.00085 at the 2-year horizon and from 0.0452 to 0.00318 at the 5-year horizon. The mean Brier skill score changed from negative to positive, increasing from -0.152 to 0.0147 at 2 years and from -0.0738 to 0.0379 at 5 years. The AUROC remains unchanged as the fitted Platt transformations were monotonic in the raw score. Isotonic recalibration produced similar reliability gains, with small AUROC changes in dementia due to its piecewise-constant mapping. In Supplementary Figs.~\ref{fig:supp_calib_2y} and \ref{fig:supp_calib_5y}, the effects from recalibrations are clearly visible as the recalibrated curves moved toward the diagonal. Full numerical results can be found in Supplementary Table~\ref{tab:supp_calibration}. Additionally, we queried the main RAVEN model at $\{1,2,3,4,5\}$-year horizons and evaluated risk monotonicity across horizons for the 5-year evaluation cohort. Mean scores generally increased with longer horizons for many tasks (Supplementary Fig.~\ref{fig:supp_mean_risk_by_horizon}). However, monotonicity was not guaranteed at the individual prediction-window level. The fraction of prediction windows with monotonically non-decreasing scores ranged from 19.9\% to 47.1\%, with 15.9\% to 33.2\% of adjacent-horizon steps decreasing (Supplementary Table~\ref{tab:supp_monotonicity}). These results indicate that the future-separator query provides a useful horizon-conditioned score for zero-shot ranking, but generally should not be interpreted as a calibrated cumulative-incidence function.}

\rev{Finally, we assessed utility at concrete operating points (Supplementary Table~\ref{tab:supp_utility}). Flagging the top 5\% of \nrev{windows} by predicted risk captured 37\%, 34\%, and 29\% of incident CHF, prostate cancer, and dementia at 2 years, roughly five- to sevenfold more than random review. Among the 100 highest-ranked \nrev{evaluation windows}, 27\% \nrev{were followed by incident} CHF against a 1.1\% base rate. Positive predictive value (PPV) at a 50\% sensitivity threshold was several-fold above prevalence across conditions. 
At the 5-year horizon, selecting the top 1\% of \nrev{prediction windows} by predicted risk identified 3.1\% to 9.3\% of future cases across diseases, with PPV ranging from 0.86\% for pancreatic cancer to 35.0\% for CHF. At a 5\% alert burden, sensitivity increased to 12.0\%--29.2\%, with PPV ranging from 0.42\% for pancreatic cancer to 27.1\% for knee osteoarthritis. These values depend strongly on disease prevalence. For example, pancreatic cancer had approximately 0.15\% 5-year prevalence; therefore, a PPV of 0.86\% at 1\% alert burden corresponds to several-fold enrichment over random review, even though the absolute PPV remains low.}

\subsection*{Generalization to EHRSHOT}

To assess out-of-distribution generalization, we take RAVEN, pretrained on internal NYU Langone data, to EHRSHOT, a public benchmark of longitudinal EHR tasks sourced from Stanford Medicine \cite{wornow2023ehrshot}, for zero-shot transfer evaluations.  There are six binary classification tasks predicting whether a patient will receive a first diagnosis of a condition within one year post-discharge from an inpatient visit: acute MI, lupus, hyperlipidemia, hypertension, celiac disease, and pancreatic cancer. 
Notably, EHRSHOT uses coding systems for many clinical concepts that differ from ours and includes features like vital that are not currently represented in RAVEN’s training vocabulary. 
Therefore, we perform lossy transfer of existing concepts and drop information that we do not cover in the vocabulary. We are only able to utilize 25\% of the data rows from EHRSHOT, meaning RAVEN accesses far fewer features in the input sequence than the baselines.

\begin{figure}[!ht]
    \centering
    \includegraphics[width=\linewidth]{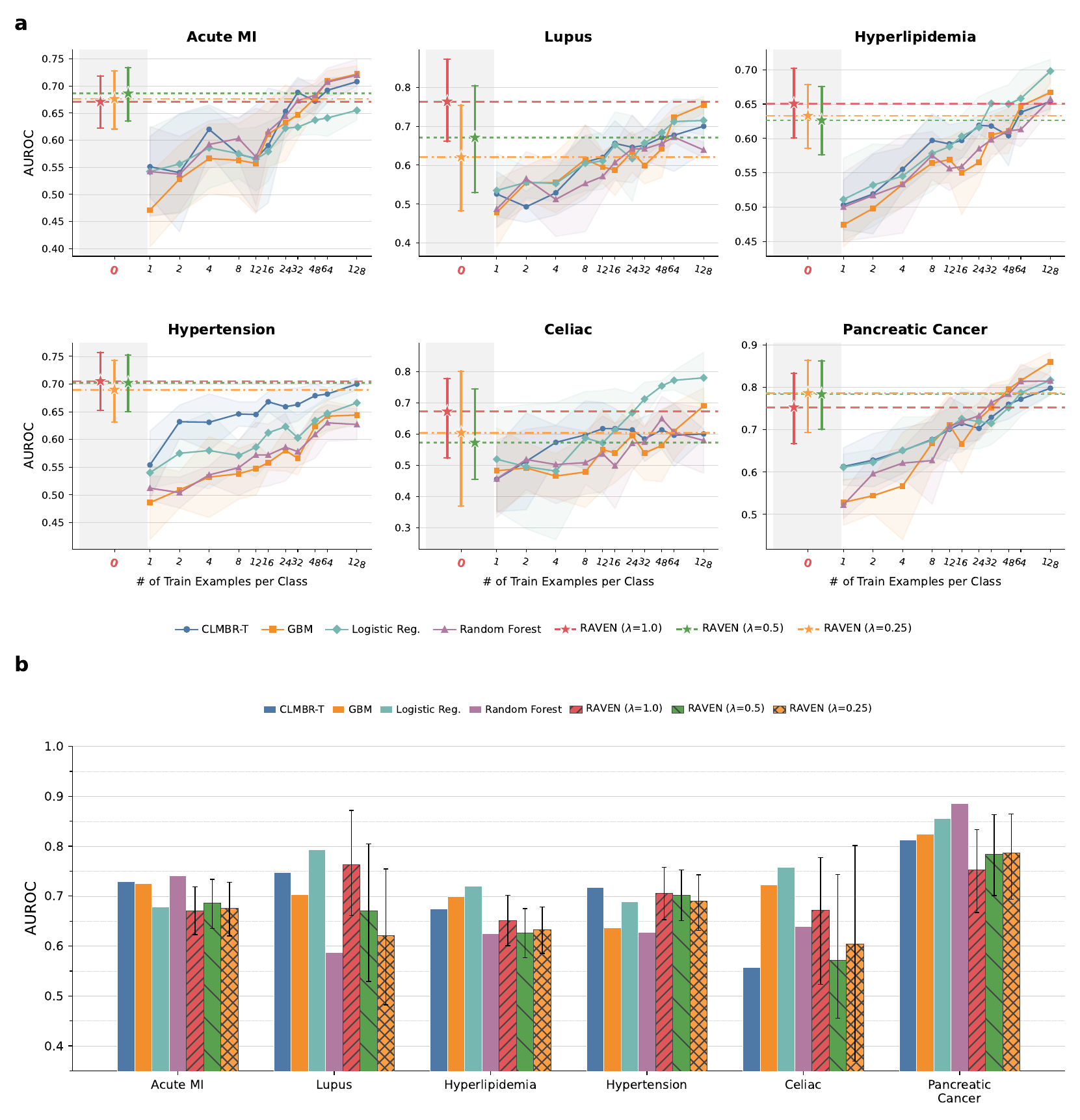}



    \caption{\textbf{Zero-shot generalization of RAVEN for new diagnosis predictions on EHRSHOT.}
    All baseline models (CLMBR-T, GBM, Logistic Regression, Random Forest) are trained on varying numbers of labeled examples per class ($K$) drawn from the training dataset, whereas RAVEN is evaluated in a zero-shot setting with no target-domain supervision.
    \textbf{a},~AUROC as a function of the number of training examples per class. Solid lines denote baseline models with s.d.\ shading across random seeds; dashed horizontal lines indicate RAVEN performance at $K=0$ for three regularization strengths ($\lambda = 1.0, 0.5, 0.25$), with 95\% confidence intervals shown at the zero-shot axis.
    RAVEN with different regularization strengths matches or exceeds baselines trained on different number of labeled examples for conditions including acute MI, hyperlipidemia, and hypertension.
    \textbf{b},~Comparison to baselines at $K = \text{all}$ (full EHRSHOT training set). RAVEN can be competitive on certain conditions with fully fine-tuned models despite having seen zero training examples. Error bars denote 95\% confidence intervals.}
    \label{fig:new-dx}
\end{figure}
\FloatBarrier

Importantly, RAVEN was pretrained on an external EHR corpus and received no labeled examples or any training from the target institution, making this a strict zero-shot transfer evaluation. 
We compared three RAVEN variants that differ in the strength of a history-dependent regularization used during pretraining to downweight repeated clinical events. Specifically, we considered $\lambda \in {1.0, 0.5, 0.25}$, where $\lambda = 1.0$ corresponds to no regularization; further details are provided later in the Recurrence-aware regularization section. We evaluated these models against four supervised baselines from EHRSHOT: CLMBR-T, gradient-boosted machines (GBM), logistic regression, and random forest \cite{steinberg2021language, wornow2023ehrshot}. Each baseline was trained with $K \in {1, 2, 4, \ldots, 128}$ labeled examples per class, as well as with the full training set (Fig.~\ref{fig:new-dx}).

\nrev{RAVEN achieved competitive discrimination across all six tasks without access to any target-domain labels (Fig.~\ref{fig:new-dx}a). Notably, for lupus and hypertension, RAVEN ($\lambda = 1.0$) achieved AUROCs of 0.763 and 0.705 (95\% CI: 0.661--0.872, 0.653--0.757), exceeding all four supervised baselines trained with up to $K = 128$ labeled examples per class. This also held for RAVEN ($\lambda = 0.5$) on hypertension, with an AUROC of 0.702 (95\% CI: 0.651--0.753). On pancreatic cancer, RAVEN ($\lambda = 0.25$) reached 0.786 (95\% CI: 0.694--0.865), outperforming all four baselines trained with up to $K = 32$ examples per class. These results suggest that RAVEN captures meaningful predictive signals that generalize across institutions without any task-specific supervision, despite substantial differences in coding systems and feature coverage.}

\nrev{The optimal regularization strength varied across conditions, and no single setting was uniformly preferred. The unregularized variant ($\lambda = 1.0$) achieved the highest AUROC on four of the six tasks: lupus, hyperlipidemia, hypertension, and celiac disease. For celiac disease, $\lambda = 1.0$ achieved an AUROC of 0.672 (95\% CI: 0.523--0.777), compared with 0.604 for $\lambda = 0.25$ and 0.572 for $\lambda = 0.5$. In contrast, intermediate regularization ($\lambda = 0.5$) yielded the highest AUROC of 0.686 (95\% CI: 0.635--0.734) on acute MI, while both regularized variants performed better on pancreatic cancer, where $\lambda = 0.25$ (AUROC = 0.786) and $\lambda = 0.5$ (AUROC = 0.784) improved over the unregularized variant (AUROC = 0.753). Overall, stronger regularization did not consistently improve zero-shot transfer under distribution shift.}

\nrev{When baselines were trained on the full labeled training set ($K = \text{all}$), the strongest supervised baseline achieved a higher AUROC than the best zero-shot RAVEN variant for each condition (Fig.~\ref{fig:new-dx}b). Nevertheless, the best RAVEN variant outperformed at least one fully supervised baseline on five of the six tasks and approached the strongest supervised baseline on hypertension (AUROC = 0.705 vs. 0.718) and lupus (0.763 vs. 0.793). The performance gap was most pronounced for pancreatic cancer (best baseline AUROC = 0.885 versus best RAVEN AUROC = 0.786) and celiac disease (0.758 vs. 0.672). These results indicate that while our pretrained model provides meaningful transfer capability, supervised training with sufficient in-distribution data remains advantageous, particularly when the model must handle missing information and incompatible clinical coding practices under distribution shift. Additionally, the wide confidence intervals observed for lupus and celiac disease reflect the limited number of positive evaluation examples ($N = 20$ and $N = 21$, respectively). Larger validation cohorts would therefore be needed to draw more definitive conclusions about zero-shot performance on low-prevalence diseases.}

\subsection*{Recurrence-aware regularization}

Certain clinical concepts such as hypertension or diabetes tend to reappear at nearly every visit once first documented. A model trained with a naive next-visit objective can therefore simply memorize and repeat existing events in the history, achieving strong overall performance without learning to anticipate new disease onsets. We introduce a history-dependent weighting scheme that uses decay parameter $\lambda$ to exponentially downweight the loss contribution of each positive target token according to how many times it has already appeared in the patient's prior visits.  When $\lambda = 1.0$, all tokens receive equal weight (no regularization), while smaller values of $\lambda$ progressively suppress the training signal from frequently repeated events. To assess the pretraining process and the effect of the repeat token regularization, we evaluate the overall next-visit prediction performance of RAVEN on several conditions.

\begin{figure}[h]
\centering
\includegraphics[width=\linewidth]{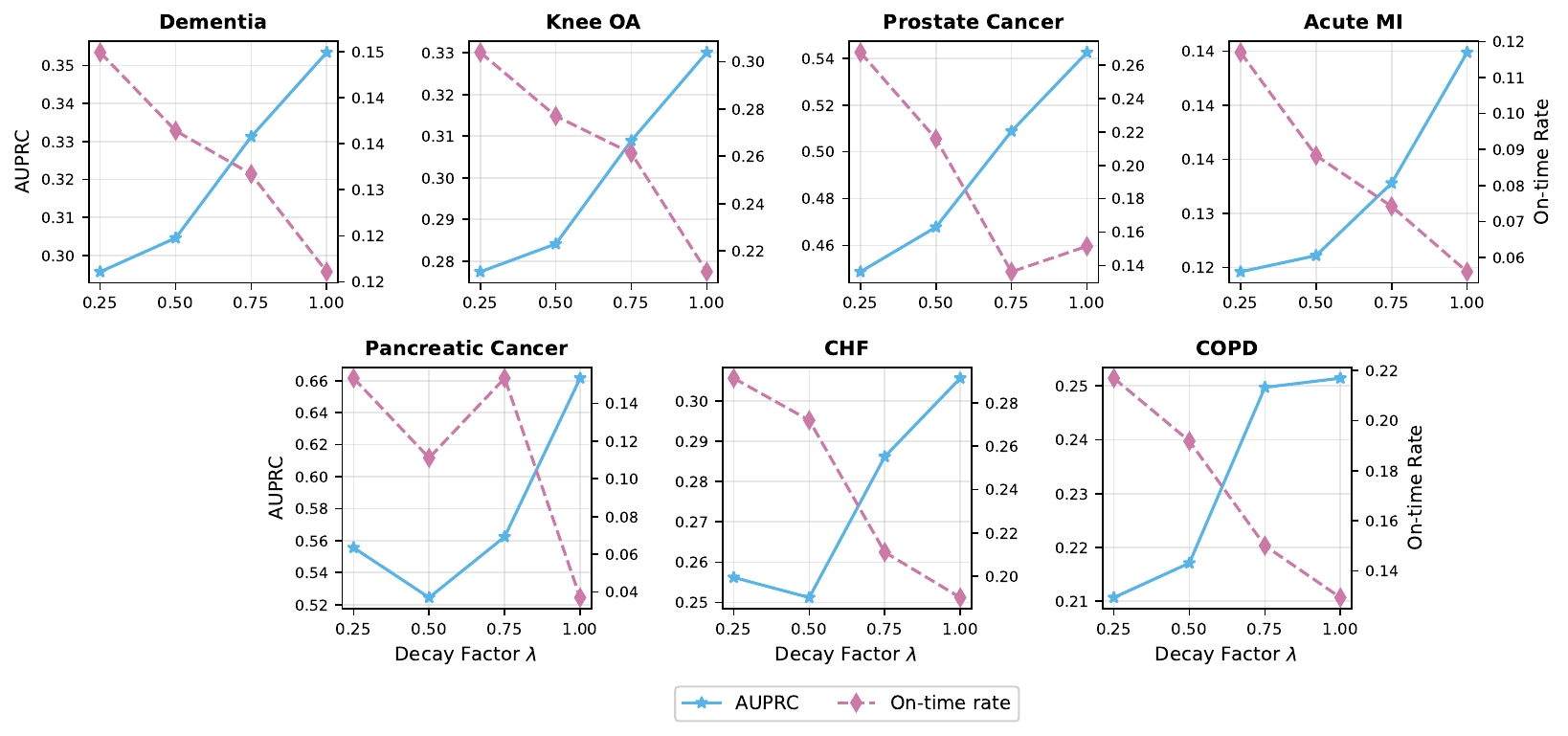}
\caption[Effect of history-dependent regularization strength]{\textbf{Effect of history-dependent regularization strength.} We sweep the decay parameter $\lambda$ that downweights repeated target events during training and report on-time rate and AUPRC across conditions. All main downstream results use a single global setting with intermediate level $\lambda^\star = 0.5$.}
\label{fig:decay_tradeoff}
\end{figure}
\newpage

Standard metrics can be inflated in longitudinal EHR data because many endpoints are chronic conditions that recur after onset, allowing models to appear accurate by simply echoing prior diagnoses.
To explicitly evaluate onset timing, we define a new metric called the on-time ratio for each condition, which measures the proportion of true positive predictions that occur at or before the first recorded onset of the condition in the patient trajectory. Here, true positives are defined at the trajectory level: a patient is considered a true positive if they eventually develop the condition and the model predicts it at any point across the rolling evaluation windows. This metric is further defined in the Methods section. 
The on-time ratio helps us distinguish performance in forecasting new onsets versus merely repeating known information. 

We investigated the effects of our decay regularization approach on repeated clinical events and varied the decay factor parameter $\lambda$ from $0.25$ to $1.0$ where smaller $\lambda$ leads to stronger penalization. Figure~\ref{fig:decay_tradeoff} shows the effect of sweeping the decay parameter $\lambda$ that downweights repeated tokens during training on both the on-time ratio and aggregate AUPRC on all rolling windows including those with repeated events. Increasing the penalization (smaller $\lambda$) significantly improved the on-time rate at the expense of the aggregate precision and recall. For example, in acute MI, the on-time ratio nearly doubles when moving from $\lambda = 1.0$ to $\lambda = 0.25$, and similar gains are observed for knee OA, CHF, and COPD. This improvement does come at a cost: AUPRC at a broad metric over all (recurring and non-recurring) events tends to decline under stronger regularization, since the model becomes less optimized on repeating the tokens it has seen before, and more encouraged to produce new onsets. 

As a more clinically relevant task, to evaluate whether the recurrence-aware regularization improves downstream disease onset prediction, we compared zero-shot forecasting performance on seven conditions at 2-year and 5-year horizons with different $\lambda$ (Fig.~\ref{fig:reg_zeroshot}). Regularization consistently improved AUROC for many conditions: at the 2-year horizon, $\lambda = 0.5$ improved AUROC over the unregularized baseline for dementia ($+0.017$), knee OA ($+0.020$), and acute MI ($+0.017$), with similar gains at the 5-year horizon (Fig.~\ref{fig:reg_zeroshot}a, b). Prostate cancer, which already exhibited strong discrimination (AUROC $> 0.90$), showed modest but consistent improvement across all regularization strengths. Full per-condition results for each $\lambda$ are reported in Supplementary Tables \ref{tab:zs_reg025} to \ref{tab:zs_reg100}.
{\setlength{\textfloatsep}{8pt plus 2pt minus 2pt}%
\begin{figure}[!t]
\centering
\includegraphics[width=\linewidth]{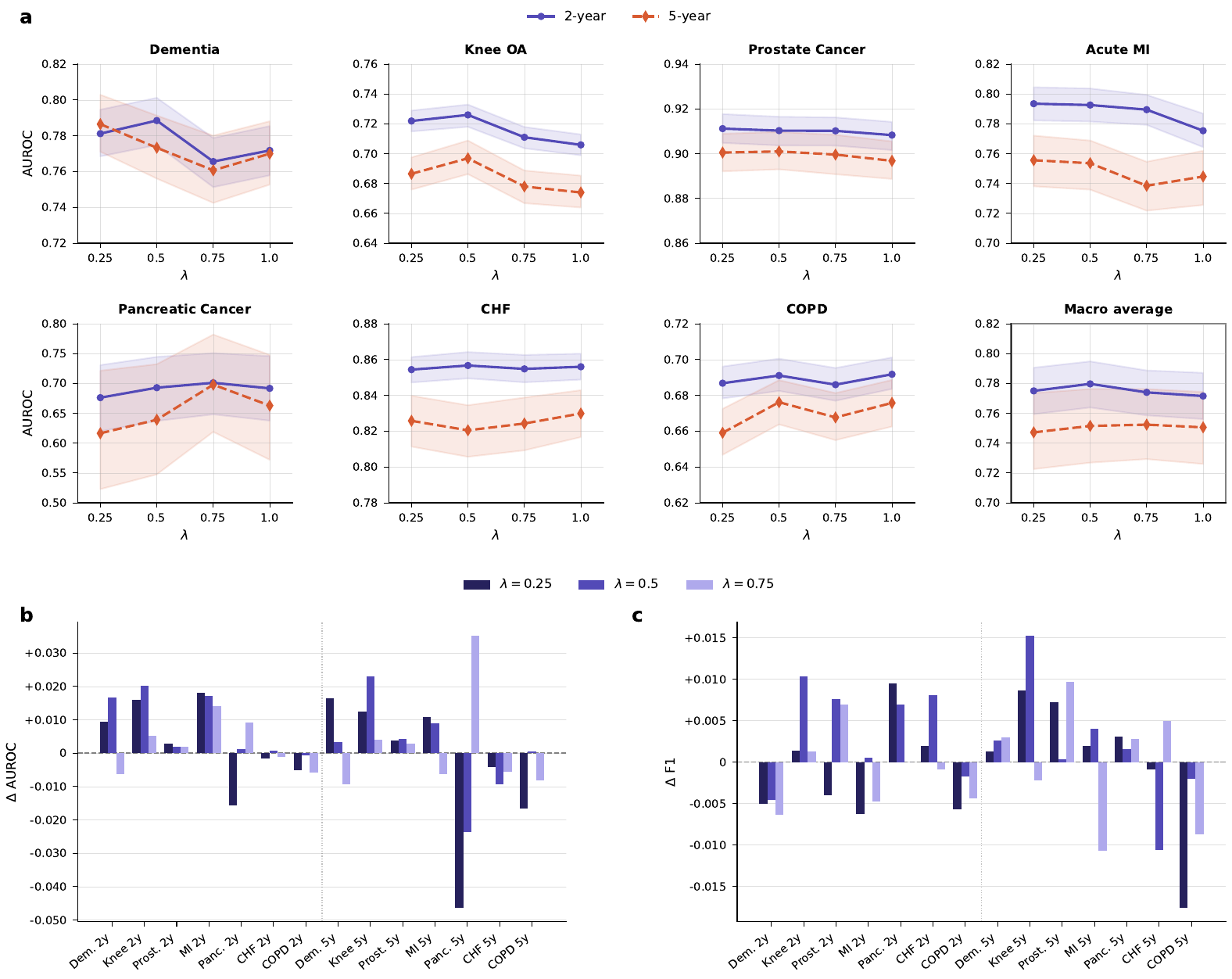}
\caption[Effect of recurrence-aware regularization on zero-shot disease onset prediction]{\textbf{Effect of recurrence-aware regularization on zero-shot disease onset prediction.} \textbf{a,} Per-condition AUROC as a function of the decay parameter $\lambda$ for the 2-year and 5-year horizons, with shaded 95\% confidence intervals. The macro-average panel (bottom right) summarizes the overall trend. All results use 144M model trained on the full dataset; lower $\lambda$ corresponds to stronger penalization of repeated clinical events. \textbf{b,} Change in AUROC from no regularization ($\lambda$ = 1.0) across seven conditions at 2-year and 5-year horizons. Bars above zero indicate improved discrimination under regularization. \textbf{c,} Corresponding change in F1 score. Dashed vertical lines separate the 2-year and 5-year evaluations. }
\label{fig:reg_zeroshot}
\end{figure}
\FloatBarrier}
In addition to change in AUROC, the threshold-dependent F1 score remained stable or improved for most conditions under regularization (Fig.~\ref{fig:reg_zeroshot}c). The benefit to downstream performance metrics, however, was not universal. CHF and COPD showed minimal sensitivity to $\lambda$ at both horizons, with AUROC varying by less than 0.01 across settings. Pancreatic cancer presented an exception: at the 5-year horizon, strong regularization ($\lambda = 0.25$) reduced AUROC by 0.046 relative to the unregularized model, likely reflecting the extremely low event prevalence where the model benefits from retaining all available signal, including recurrence patterns. Per-condition AUROC curves (Fig.~\ref{fig:reg_zeroshot}a) confirmed these trends. At the macro-average level, regularized models ($\lambda = 0.5$) achieved higher AUROC than the unregularized baseline at both horizons ($0.780$ vs.\ $0.772$ at 2 years; $0.752$ vs.\ $0.751$ at 5 years), supporting $\lambda^{\star} = 0.5$ as a balanced operating point which we use for downstream evaluations.

\begin{figure}[ht]
\centering
\includegraphics[width=\linewidth]{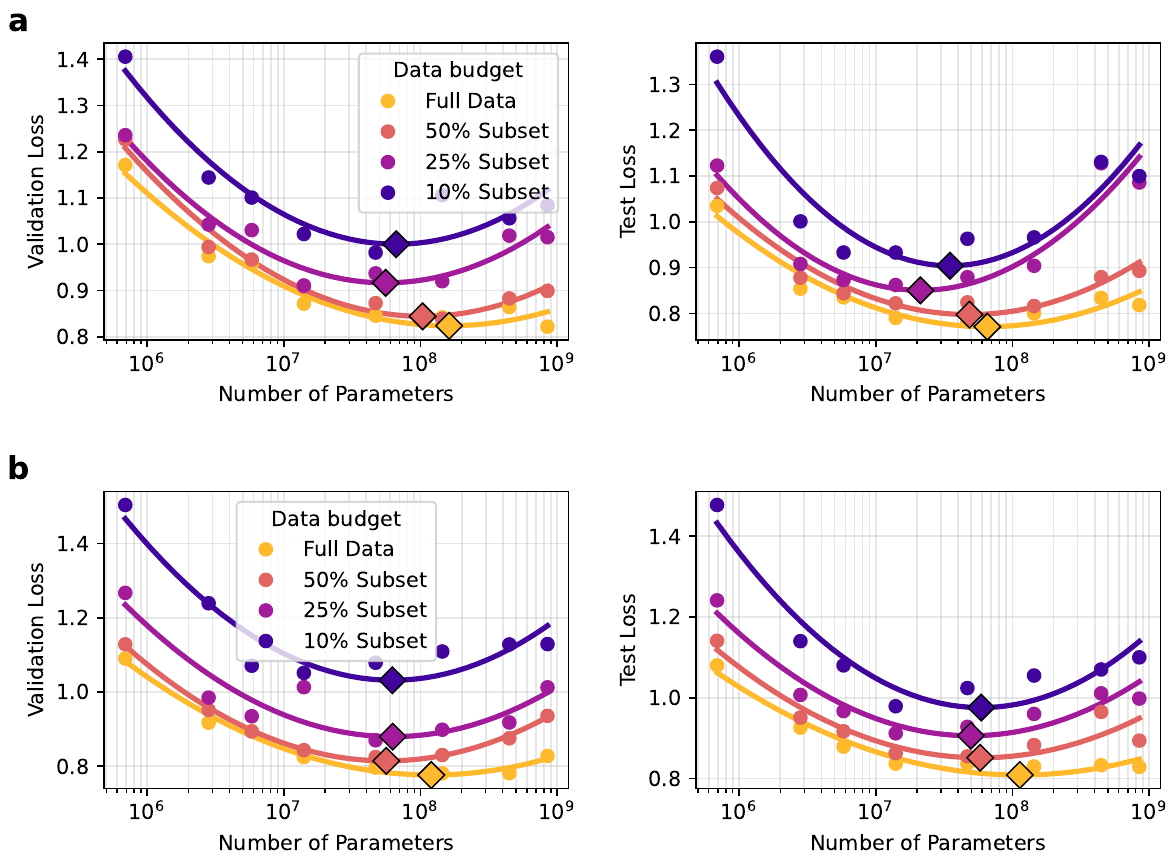}


\caption[Scaling under standard and regularized objectives]{\textbf{Compute-saturated scaling of standard and RAVEN pretraining.}  \textbf{a,} Validation loss and test loss under the standard next-visit objective as a function of model size across multiple dataset budgets. The fitted minima shift toward larger models as the data budget increases, indicating that the optimal capacity depends strongly on the amount of available training data. \textbf{b,} The same scaling evidence with full RAVEN history-dependent regularization enabled during pretraining, showing consistent behaviors.}
\label{fig:loss_scaling}
\end{figure}
\FloatBarrier

\subsection*{Compute-saturated scaling}

Unlike traditional scaling studies, we consider a regime of unconstrained compute, in which we train the largest feasible models for as long as needed. The goal is to build the best possible model with the sole constraint of data. To study how RAVEN scales under this regime, we evaluate eight model configurations spanning 0.69 million to 848.80 million parameters, trained on four data budgets. The configurations also vary in width-to-depth ratio to probe whether this affects performance independently of model size; full configuration details are in the Supplementary Table \ref{tab:model_configs}. The four data budgets consist of three randomly drawn subsets of the training data (10\%, 25\%, and 50\%) plus the full dataset, enabling us to characterize how data scale interacts with model size. Rather than fixing the number of epochs or the total compute budget, we adopt a convergence-oriented training protocol with early stopping on validation loss. The actual number of epochs trained ranges from 10 to 20 depending on the data scale where smaller datasets benefited more from further repeating. In total, we trained more than 80 models to convergence across all settings. This design allows us both to identify the model sizes that perform best on downstream tasks and to characterize the relationship between model size and data scale in this scaling regime.

Figure~\ref{fig:loss_scaling}a reports validation and test pretraining loss as a function of model size across dataset budgets. Loss exhibits a U-shaped dependence on parameter count: larger models are more sample-efficient up to a point, beyond which performance degrades. To summarize these trends, we fit a smooth curve to the loss-versus-model-size relationship for each budget and identify the fitted minimum (diamond markers). In the repeated-epoch regime, we observe that increasing parameter count will eventually lead to overfitting, consistent with observations in LLMs \cite{kim2026pre}. The estimated optimal model size shifts with data budget: larger datasets support larger optimal models, while smaller budgets favor more compact architectures. This highlights that over-parameterization can be harmful when training repeatedly over a finite cohort.

Additionally, Figure~\ref{fig:loss_scaling}b repeats the same scaling analysis with history-dependent downweighting of repeated events, and the results are broadly consistent with the standard pretraining findings. We also note that the pretraining loss performance correlates largely with the downstream tasks' performance. Under full RAVEN pretraining, we see the right tail of the U-shaped curve continues to decrease with model size, suggesting headroom for further scaling, a pattern consistent with LLM scaling laws \cite{kaplan2020scaling}. Unfortunately, we do not have access to additional data needed to trace a strict power law across more budget levels. Unless otherwise noted, we use the 144M-parameter model as our main model, selected by validation loss under the full-data budget in the sweep.

\vspace{-0.25em}

\begin{figure}[h]
\centering
\includegraphics[width=\linewidth]{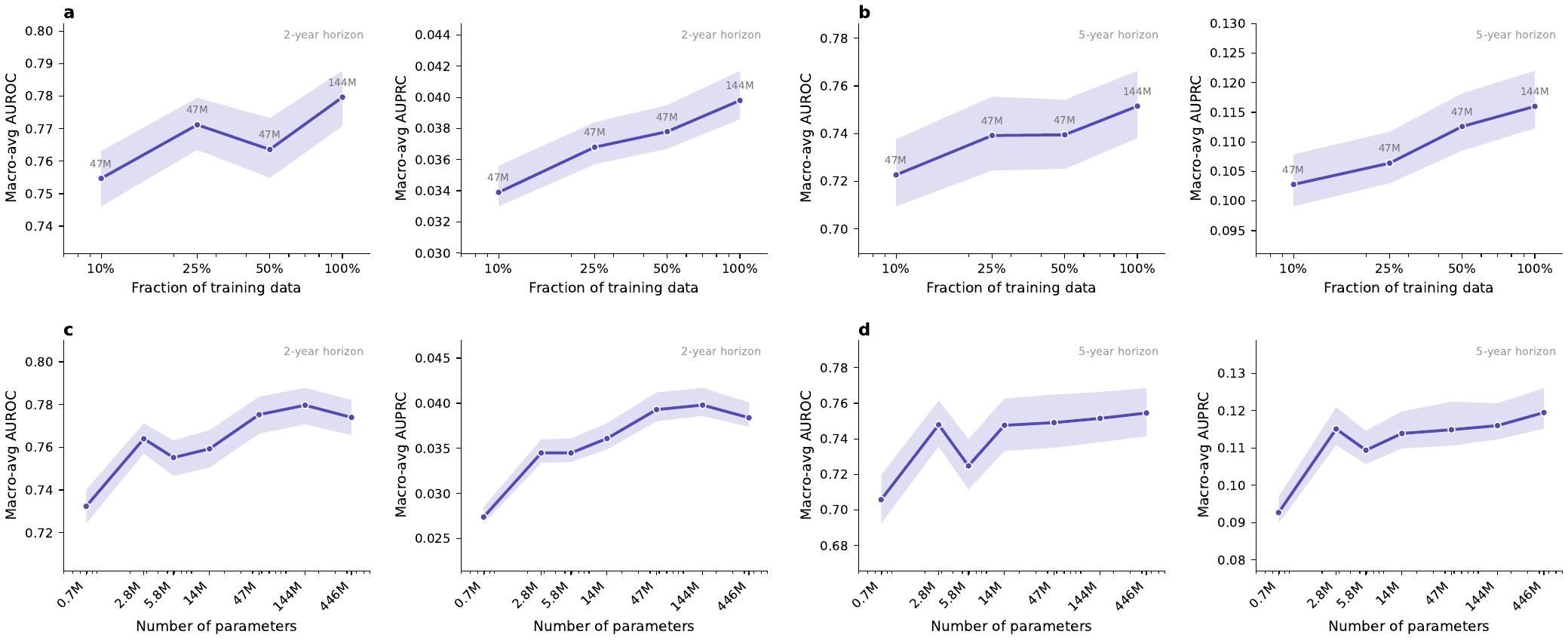}


\caption[Scaling downstream zero-shot disease forecasting performance]{\textbf{Scaling downstream zero-shot disease forecasting performance.} \textbf{a., b.} Macro-average AUROC and AUPRC for 2-year (\textbf{a}) and 5-year (\textbf{b}) zero-shot disease onset prediction with shaded 95\% confidence intervals, averaged across seven conditions. At each data budget, the optimal model size from the compute-saturated scaling analysis (Fig. \ref{fig:loss_scaling}) is used: 47M parameters for the 10\%, 25\%, and 50\% subsets, and 144M parameters for the full dataset. All models are trained with recurrence-aware regularization ($\lambda = 0.5$). \textbf{c., d.} Same zero-shot task metrics 2-year (\textbf{c}) and 5-year (\textbf{d}) plotted with respect to model sizes. All models are trained with full data and regularization ($\lambda = 0.5$).}
\label{fig:scaling_zero_shot}
\end{figure}
 
\vspace{-0.25em}

We further explored whether the scaling behaviors observed in pretraining loss propagate to downstream zero-shot forecasting metrics by evaluating the optimal model at each data budget on disease onset prediction at both 2-year and 5-year horizons (Fig. \ref{fig:scaling_zero_shot}a, b). Across all four settings, each data budget uses the model size selected by the compute-saturated scaling sweep: 47M parameters for the 10–50\% subsets and 144M for the full dataset. At the 5-year horizon, both metrics show consistent improvement with increasing data: macro-average AUROC rises from 0.723 at 10\% of the training data to 0.752 at full data, while macro-average AUPRC increases from 0.103 to 0.116. At the 2-year horizon, AUROC improves from 0.755 to 0.780 and AUPRC from 0.034 to 0.040. Full macro-averaged numbers across data budgets are reported in Supplementary Table \ref{tab:zs_data_budget_macro_ci}.

Also, we examined how model size affects downstream zero-shot forecasting when all models are trained on the full cohort. Figure \ref{fig:scaling_zero_shot}c, d shows macro-average AUROC and AUPRC across seven conditions at both horizons for models ranging from 0.69M to 446M parameters (full results in Supplementary Table \ref{tab:zs_model_size_macro_ci}). The most pronounced gains occur in the smallest model range: scaling from 0.69M to 2.81M parameters improves 5-year AUROC from 0.706 to 0.748 and 5-year AUPRC from 0.093 to 0.115. Performance plateaus and even diminishes once we move to models larger than the optimal 144 million parameters from Fig. \ref{fig:loss_scaling}. These findings suggest that increasing model capacity beyond a modest threshold yields diminishing returns on downstream task performance, whereas expanding the pretraining cohort continues to produce meaningful gains. 
This asymmetry has practical implications for resource allocation: acquiring additional patient data is likely to be more impactful than scaling model parameters alone.

\section*{Discussion}

Zero-shot disease-onset forecasting with RAVEN can match or approach per-endpoint supervised training across diverse conditions and multi-year horizons. RAVEN respects the unordered nature of events within a visit, supports zero-shot querying at future horizons, and avoids retraining separate supervised models for every endpoint and prediction window. We found that predictions via a single forward pass produced competitive performance compared to commonly used autoregressive rollout baselines, which require many simulated trajectories per query, as well as prompted medical LLMs, and in several settings approached a fully fine-tuned representation baseline (Tables~\ref{tab:zs_2y_all_baselines}--\ref{tab:zs_5y_all_baselines}). The same pretrained model also transferred competitively to new diagnosis tasks on an external benchmark under challenging coding and feature-coverage mismatches. Our findings indicate that zero-shot forecasting is not only feasible but also operationally attractive and demonstrate that competitive performance does not require per-endpoint supervised training. For health systems that need to monitor many conditions simultaneously, this shifts the bottleneck from building and maintaining a large collection of task-specific models to validating and calibrating a single foundation-model interface. In practice, this can enable rapid endpoint expansion to new code sets, substantially lower inference cost than rollout-based scoring, and simpler deployment surfaces, with the caveat that calibration and threshold selection remain application-specific.

\rev{These zero-shot results are not an artifact of demographic or cohort-selection shortcuts. RAVEN maintained strong performance when each task was restricted \nrev{by demographic eligibility criteria}, and it matched or exceeded supervised demographic and utilization baselines, despite using no endpoint-specific labels. Our calibration results indicate that, in its raw uncalibrated form, RAVEN is best interpreted as a zero-shot risk stratification model rather than as a source of deployment-ready absolute risk probabilities. However, this limitation is not inherent to the approach: a single recalibration fit on a validation cohort substantially improves absolute-risk reliability without altering the underlying risk ranking. We therefore frame the present zero-shot use as risk stratification across a specified horizon, with calibrated absolute-risk deployment requiring target-cohort recalibration and, ideally, a survival-specific output parameterization or cumulative-incidence modeling layer.}

A key contribution of this work is highlighting an important lesson for evaluating foundation models trained on longitudinal EHRs. Our pretraining evaluations revealed an inherent trade-off when emphasizing the prediction of new clinical events: stronger regularization improved the on-time rate for new onsets but could reduce overall precision and recall. This underscores that if models are blindly evaluated and optimized solely on aggregate metrics without careful design, there is a risk of developing clinically ineffective systems that primarily repeat patient history rather than predict new disease onsets. For instance, a model might achieve high recall by simply echoing chronic diagnoses seen earlier in the trajectory, without demonstrating a genuine understanding of when a condition first arose or changed in severity. Our regularization scheme directly mitigates this issue by penalizing predictions of frequently repeated clinical events, guiding the model to prioritize meaningful predictions of new disease occurrences. By explicitly separating and evaluating these two phenomena—predicting new onsets versus repeating known information—evaluations can better reflect real-world clinical needs, such as providing early warnings for new diagnoses or enabling timely intervention, leading to more clinically relevant and accurate assessments of model performance.

In addition, because patient data are inherently scarce, EHR foundation models should not adopt the same scaling paradigm as LLMs trained on web-scale corpora. In our data-constrained, compute-saturated regime, we observed a U-shaped dependence of validation and test loss on model size across multiple data budgets, implying that practical EHR foundation modeling behaves more like a capacity–data matching problem. Naively increasing parameter count without increasing effective data diversity can be neutral or even harmful, and capacity should instead be selected jointly with data budget and training protocol. That said, there remains substantial opportunity to scale models like RAVEN further in healthcare by aggregating health-system level data from multiple sources.

Our comparison with a prompted medical LLM further shows where general-purpose models currently fit in this landscape. The comparatively weaker performance of MedGemma-27B in our experiments
suggests that general-purpose medical LLMs, when used purely through
prompting on serialized records, do not yet replace structured foundation
models for long-horizon EHR forecasting. Nevertheless, MedGemma did exhibit considerable discrimination on several disease endpoints, indicating that general medical knowledge encoded in LLMs can extract meaningful longitudinal risk signals from raw records even without any task-specific training or exposure to specific structured EHR pretraining data. These observations position medical LLMs and structured EHR foundation models as complementary rather than competing: the former offer broad clinical priors and natural-language interfaces, while the latter provide efficient, horizon-conditioned risk estimates grounded in the structural regularities of longitudinal records.

Several limitations remain. First, \rev{pretraining and the primary evaluations} were conducted on a single large de-identified health-system cohort, so the precise trade-offs may differ under other \rev{reference conditions, including different} coding practices, population mixes, and follow-up patterns. In particular, our current vocabulary does not operate under a hierarchical coding system such as ICD-10; consequently, codes from external systems that cannot be mapped to our granular token concepts may fail to provide even generic high-level signals to the model. \rev{We do provide out-of-distribution evidence through zero-shot transfer to EHRSHOT, a benchmark drawn from a separate health system with different coding conventions and feature coverage; however, this evaluates a model trained on our cohort rather than replicating the framework itself across health systems. A more complete validation, which we leave to future work, would retrain RAVEN end-to-end on a distinct source cohort and coding ontology to confirm that our central findings generalize rather than reflect properties of our particular reference data. This is a foundational step that precedes the broader extensions discussed below.} 

Second, documented onset in the EHR is only a proxy for true clinical onset and may lag the underlying disease process. \rev{Our labels mitigate this in part by relying on curated code sets of diagnosis and medications rather than single billing codes, and by excluding windows whose input history already contains a label code as well as onsets within one year of the prediction point. The residual mismatch could be narrowed further with more elaborate, disease-specific phenotype definitions. For example, we could require multiple qualifying occurrences, use computable phenotype by combining confirmatory laboratory or medication evidence, and even incorporate data-driven methods that learn from noisy labels \cite{agarwal2016learning, banda2017electronic}. These choices may improve specificity for clinically meaningful onset but can also limit portability across cohorts and health systems, as they may become inherently more coding-system-specific.} It is also unclear whether imitating existing clinical practices is optimal for every patient in the healthcare system \cite{jayaraman2024primer, navar2019electronic}. Finally, we have not addressed interventional or counterfactual queries that would be needed to estimate the effects of potential new treatment policies, an important capability for moving from risk estimation toward decision support. \rev{These effect estimation queries need not be limited to clinical treatments. Lifestyle modifications and social determinants of health are increasingly captured in EHR ecosystems and are clinically meaningful intervention targets. Extending the framework toward decision support would require explicit causal assumptions, validation designs, and modeling of these intervention variables with appropriate structures in the architecture.}

\rev{Beyond strengthening the generalization of our findings,} a subsequent step is to expand beyond binary onset toward progression, remission, and treatment response, where repetition and timing are clinically meaningful in more nuanced ways. Our findings in model scaling further motivate the development of more data-efficient algorithms for learning longitudinal EHRs, as well as methods for generating synthetic data to augment constrained data sources. Moreover, we have not yet fully leveraged the generative nature of this pretrained model, as we have only focused on one-step predictions so far, and it is not entirely clear what the best way is to sequentially generate future trajectories with a next-visit prediction framework. In contrast, for LLMs, there is a rich literature on inference-time algorithms that improves reasoning and generates long-form responses through reinforcement learning, and we believe these methods could potentially continue to improve foundation models on sequential tabular data. Our empirical results with prompted medical LLMs in the zero-shot comparisons also point to a promising direction: combining structured EHR foundation models with the broad clinical priors encoded in large language models. For example, using LLMs to inject prior knowledge about rare conditions or to provide retrieval-augmented context for structured generative models operating on tokenized records.

\section*{Methods}
\label{sec:methods}

\subsection*{Dataset and cohort}

We develop and evaluate our pretraining framework using a large-scale, de-identified longitudinal EHR dataset derived from NYU Langone Health, a major academic health system in New York City. This study was approved \rev{under i20-01095} by the Institutional Review Board (IRB) of NYU Langone, \rev{and all procedures were performed in accordance with the Declaration of Helsinki}. \rev{Given the retrospective nature of the study and the use of de-identified data, the requirement for informed consent was waived by the IRB.} The dataset spans both inpatient and outpatient encounters over a ten-year period (January 2013 to January 2023) and comprises approximately 1.29 million unique patients (N=1,288,242), with a median of 21 visits per patient (mean: 37.76, range: 2--2123). All data splits are performed at the patient level to prevent information leakage across training (70\%), validation (15\%), and test (15\%) sets.

For external evaluation, we use EHRSHOT, a public benchmark of longitudinal structured EHR data from 6,739 patients at Stanford Medicine \cite{wornow2023ehrshot}. It contains adult patients only and excludes patients with a history of fewer than 10 total clinical events. The training, validation, and test sets are roughly the same size; we only leverage the test set for evaluating RAVEN ($N = 2212$). We include statistics of demographic information for both our internal data at NYU Langone and EHRSHOT in Table~\ref{tab:demographics}. \rev{As stated in the benchmark report, all EHRSHOT data are de-identified and the study therefore did not require Institutional Review Board approval \cite{wornow2023ehrshot}.}

\begin{table}[ht]
\centering
\caption[Patient demographics for NYU Langone and EHRSHOT cohorts]{\textbf{Patient demographics for NYU Langone and EHRSHOT cohorts.} Patient demographics in the train, validation, and test splits for the two cohorts. For EHRSHOT, we only used the test set.}
\label{tab:demographics}
\small
\setlength{\tabcolsep}{4pt}
\resizebox{\linewidth}{!}{%
\begin{tabular}{ll rrrr rrrr}
\toprule
& & \multicolumn{4}{c}{\textbf{NYU Langone (N=1{,}288{,}242)}} & \multicolumn{4}{c}{\textbf{EHRSHOT (N=6{,}739)}} \\
\cmidrule(lr){3-6} \cmidrule(lr){7-10}
\textbf{Attribute} & & \textbf{Train} & \textbf{Val} & \textbf{Test} & \textbf{All} & \textbf{Train} & \textbf{Val} & \textbf{Test} & \textbf{All} \\
\midrule
\textbf{Gender}
 & Male        & 347{,}497 & 74{,}128  & 74{,}291  & 495{,}916   & 1{,}122 & 1{,}090 & 1{,}086 & 3{,}298 \\
 & Female      & 531{,}280 & 114{,}160 & 113{,}995 & 759{,}435   & 1{,}173 & 1{,}142 & 1{,}126 & 3{,}441 \\
\midrule
\textbf{Age}
 & $\leq$ 20      & 73{,}478  & 15{,}743  & 15{,}597  & 104{,}818   & 8       & 3       & 2       & 13      \\
 & 21--40      & 228{,}578 & 49{,}002  & 48{,}939  & 326{,}519   & 412     & 457     & 431     & 1{,}300 \\
 & 41--60      & 291{,}009 & 62{,}040  & 62{,}348  & 415{,}397   & 648     & 597     & 576     & 1{,}821 \\
 & 61--80      & 179{,}244 & 38{,}343  & 38{,}547  & 256{,}134   & 916     & 892     & 905     & 2{,}713 \\
 & $\geq$ 81      & 17{,}866  & 3{,}939   & 3{,}842   & 25{,}647    & 311     & 283     & 298     & 892     \\
\midrule
\textbf{Race}
 & American Indian    &          &          &          &            & 14      & 7       & 4       & 25      \\
 & Asian              & 40{,}859  & 8{,}829   & 8{,}718   & 58{,}406    & 356     & 347     & 340     & 1{,}043 \\
 & Black              & 85{,}477  & 18{,}384  & 18{,}442  & 122{,}303   & 98      & 105     & 95      & 298     \\
 & Pacific Islander   & 3{,}573   & 804       & 755       & 5{,}132     & 23      & 21      & 30      & 74      \\
 & White              & 498{,}249 & 106{,}429 & 106{,}583 & 711{,}261   & 1{,}286 & 1{,}222 & 1{,}228 & 3{,}736 \\
 & Unknown            & 195{,}931 & 41{,}999  & 42{,}041  & 279{,}971   & 518     & 530     & 515     & 1{,}563 \\
\midrule
\textbf{Ethnicity}
 & Hispanic       & 12{,}053  & 2{,}639   & 2{,}575   & 17{,}267    & 374     & 342     & 322     & 1{,}038 \\
 & Non-Hispanic   & 143{,}565 & 30{,}697  & 30{,}955  & 205{,}217   & 1{,}921 & 1{,}890 & 1{,}890 & 5{,}701 \\
\midrule
\textbf{Total}
 &                & 901{,}769 & 193{,}236 & 193{,}237 & 1{,}288{,}242 & 2{,}295 & 2{,}232 & 2{,}212 & 6{,}739 \\
\bottomrule
\end{tabular}%
}
\end{table}

\subsection*{Tokenization and input representation}

Each patient record is represented as a sequence of visits, where each visit consists of a set of heterogeneous clinical events. We include patient demographics (self-reported ethnicity, race, sex), age at visit, diagnoses (ICD-10 codes), prescribed medications, and laboratory results (LOINC codes and values). Continuous variables, namely age and laboratory test results, are discretized into quantile-based bins (e.g., 10 bins for lab results). Each unique demographic category, discretized age bin, medication concept, ICD-10 code, and discretized lab result bin is treated as a distinct token. This results in a total vocabulary size $|V|$ of 42,337 unique tokens. On average, each visit contains 11.16 tokens, and each patient trajectory comprises 474.21 tokens (median: 191).

Within each patient trajectory, visits are ordered chronologically and separated by a special \texttt{<sep>} token that explicitly marks visit boundaries. Importantly, no artificial ordering is imposed on tokens within a visit, reflecting the fact that many clinical events recorded at the same encounter are unordered.

Notably, EHRSHOT uses different coding systems for many clinical concepts including diagnoses and medications, and it contains features such as vitals and care sites that are not in RAVEN's training vocabulary. We incorporated mapping mechanisms (SNOMED to ICD-10 for diagnoses; RxNorm to internal codes for medications) that do not precisely translate to our granular vocabulary given the lossy nature of the mapping, and we dropped unmapped or uncovered information. Certain mapped codes are dropped because they do not appear in our vocabulary. For instance, we were able to translate and use 6,211 out of 11,598 unique diagnosis codes and 3,743 out of 5,433 medication codes from EHRSHOT to our own vocabulary. While we mapped 92\% of medication events and 29\% diagnosis events, we only utilized approximately 25\% of total rows of data from EHRSHOT with loss primarily coming from missing vitals, meaning RAVEN accesses fewer features in the input sequence than other baselines.

\subsection*{Model architecture}

We employ a decoder-only Transformer \cite{vaswani2017attention} based on the GPT-2 architecture \cite{radford2019language} for RAVEN, adapted to model longitudinal EHR data. Let $\{x_1, \dots, x_T\}$ denote the tokenized multi-hot representation of a patient trajectory. The model parameterizes the conditional distribution $p(x_t \mid x_{<t})$ using causal self-attention. We use 8 attention heads, 8 layers, and hidden dimension of 1024 for our main 144 million parameter model with a maximum sequence length of 512. For our scaling studies, we used a variety of model architecture configurations with each model size. 
Full architecture and training hyperparameter details are provided in Supplementary Tables \ref{tab:hyper} and \ref{tab:model_configs}. 

To better reflect the structure of EHR data, we modify the standard causal attention mechanism to operate at the visit level. Specifically, tokens belonging to the same visit are allowed to attend to one another without restriction, while causal ordering is enforced only across visits. That is, a token from visit $v$ may attend to all tokens from visits $v' \leq v$, but not to tokens from future visits. This design avoids imposing spurious order within visits while preserving temporal causality across encounters. We also incorporate rotary positional embeddings (RoPE) \cite{su2024roformer} to encode temporal information. All tokens within a visit share the same positional embedding, corresponding to the elapsed time since the patient's first recorded visit. The \texttt{<sep>} token concluding visit $v$ is assigned the positional embedding of the subsequent visit $v+1$, explicitly encoding the inter-visit time gap. This allows the model to condition the prediction on both patient history and the timing of the next encounter. 

Moreover, for training efficiency, we use sequence packing to have multiple patients in one training sequence if their tokens can fit the context length \cite{raffel2020exploring, krell2021efficient}. To prevent information leakage across patients, we explicitly condition the model on patient rank in the batch. Specifically, each token embedding is augmented with a patient-specific embedding derived from a sinusoidal encoding of the window’s local identifier within the packed sequence. For example, if a packed sequence contains windows from three patients, tokens from those windows are tagged with identifiers such as 0, 1, and 2 to indicate which packed patient window they belong to. Our implementation allows attention across concatenated patient sequences, potentially offering broader context alongside improved GPU throughput. All downstream evaluations and zero-shot inference are performed on individual patient trajectories without packing.



\subsection*{RAVEN Pretraining}

RAVEN is pretrained using a next-visit multi-label prediction objective. Given a patient history up to visit $v$, the model predicts the set of clinical events that will occur at visit $v+1$. Concretely, the hidden representation of the \texttt{<sep>} token following visit $v$ is passed through a linear projection and sigmoid activation to produce a probability $\hat{P}_{v+1, k} = P(k | H_{t_v}, t_{v+1})$ for each token $k$ in the vocabulary.

Let $V_{v+1} \in \{0,1\}^{|V|}$ denote the multi-hot vector indicating which tokens occur at visit $v+1$, and let $\hat{P}_{v+1}$ denote the model's predicted probabilities. The base loss is a binary cross-entropy objective summed over all tokens:
\[
\mathcal{L}_{\mathrm{BCE}}
=
- \sum_{k=1}^{|\mathcal{V}|}
\left[
V_{v+1,k}\log(\hat{P}_{v+1,k}) +
(1 - V_{v+1,k})\log(1-\hat{P}_{v+1,k})
\right].
\]


To prevent the model from memorizing trivial repetition, we introduce a history-dependent regularization scheme that downweights the contribution of frequently observed tokens. For each positive token $k$ in the target visit $v+1$, we compute its count $c(k, H_v)$ in the patient's prior history $H_v$. The weight is:
\[
w_{v+1,k} = \max(\lambda^{c(k,H_v)}, w_{\min}),
\]
where $\lambda \in (0,1]$ is a decay factor hyperparameter, and $w_{\min}$ is a minimum weight hyperparameter that prevents the weight from vanishing entirely. For token predictions where the ground truth is negative ($V_{v+1, k}=0$), the weight is 1. The final loss is the weighted binary cross-entropy:
\[
\mathcal{L}_{\mathrm{reg}}
=
- \sum_{k=1}^{|\mathcal{V}|}
w'_{v+1,k}
\left[
V_{v+1,k}\log(\hat{P}_{v+1,k}) +
(1 - V_{v+1,k})\log(1-\hat{P}_{v+1,k})
\right],
\]
where $w'_{v+1, k} = w_{v+1, k}$ if $V_{v+1, k}=1$, and $w'_{v+1, k}=1$ otherwise. This allows the model to focus on potentially novel events rather than highly predictable recurrent tokens. In the main experiments, we regularize with intermediate strengths and select a single global setting $\lambda^\star = 0.5$ through extensive evaluations.

\subsection*{Comparative baselines}

We benchmark our next-visit objective against several representative baselines, including autoregressive next-token simulation methods, a supervised BERT-based model, and a prompted large language model baseline. All EHR models use the same tokenizer and vocabulary ($|V| = 42337$). For simulation-based models, long-horizon risk is estimated via inference with $R{=}\rev{64}$ rollouts per patient window. First, we include a standard next-token model with cross-entropy loss (Multiclass). This baseline flattens each patient trajectory into a single token sequence with pre-defined structural order for different types of clinical concepts within the same visit. For instance, the demographics and diagnosis codes are inserted at the front of the visit, followed by labs and medications, similar to prior work in this space \cite{waxler2025generative}. It also trains a decoder-only Transformer with standard cross-entropy loss for predicting the next token. Inter-visit time gaps are provided as observed input tokens during both training and evaluation.

Many clinical events within a visit are unordered. Similar to RAVEN, to reduce sensitivity to arbitrary within-visit token order, we also include a next-token set-based loss model (SeqLoss) that uses a set-prediction loss \cite{welleck2018loss}. At each decoding step within a visit, the target distribution is uniform over the remaining unseen tokens:
\[
q_t(x) =
\begin{cases}
\frac{1}{|U_t|}, & x \in U_t,\\
0, & \text{otherwise,}
\end{cases}
\]
where $U_t$ is the set of tokens not yet predicted at step $t$. Let $p_\theta(\cdot \mid \cdot)$ be the model's next-token distribution. The model is rewarded for selecting any remaining correct token, rather than a specific permutation. The last next-token baseline we benchmark is a joint gap-and-event generation model (EGE). This baseline jointly models inter-visit time gaps and clinical events in a fully autoregressive manner, predicting the next gap token and then the next clinical event token(s), repeating until a visit-separator token is produced.

Additionally, we prompt a medical large language model baseline
using MedGemma-27B \cite{sellergren2026medgemma}, evaluated in a zero-shot setting with no task-specific fine-tuning. Unlike the structured EHR baselines above, MedGemma operates directly on serialized patient histories rendered as text with JSON formatting and produces predictions through in-context reasoning alone. The prompt template can be found in the Supplementary. This baseline is intended to measure how well a general-purpose medical LLM, leveraging broad clinical knowledge encoded during its own pretraining, can perform long-horizon disease forecasting without any exposure to structured longitudinal EHR pretraining or task-specific adaptation.

To form a strong reference, we compare against a supervised baseline using a BERT-based foundation model \cite{zhu2024predicting} that leverages masked pretraining followed by task-specific fine-tuning for each condition and horizon. This bidirectional Transformer encoder is trained to reconstruct 20\% randomly masked tokens and uses a prediction head on top of final representation token to obtain supervised signals during fine-tuning. \rev{Lastly, to assess whether model performance could be explained by simple demographics or healthcare utilization signals, we include two simple supervised baselines trained for each specific disease and prediction horizon: logistic regression (LR) and gradient boosting machine (GBM) classifiers. Both use a fixed feature vector computed strictly from data available before the prediction time, comprising age, sex, race and ethnicity, code-type counts, and cumulative counts of prior visits and codes together with similar counts of recent healthcare utilization in the last 30, 90, and 180 days.}


\subsection*{Baseline inference procedure}

Our next-visit model produces a horizon-conditioned risk score in a single forward pass. In contrast, next-token baselines do not natively output a horizon-conditioned probability; therefore, we estimate disease-onset risk via Monte Carlo simulations of future trajectories \cite{renc2024zero, waxler2025generative}.

Given a context history window ending at time $t$ and a horizon $H \in \{2,5\}$ years, we estimate risk using $R{=}\rev{64}$ independent rollouts. For the Multiclass and SeqLoss baselines, we use a discrete-time rollout: we query the model at fixed 3-month increments $\Delta$ and sample the predicted event tokens for each step, yielding simulated timestamps $\{t+\Delta, t+2\Delta, \dots, t+H\}$ with $\Delta{=}3$ months. For EGE, we generate a fully autoregressive token stream and stop when either (i) the simulated time exceeds $t+H$ (via generated gap tokens) or (ii) a maximum of 512 tokens is generated, whichever occurs first. Let $\mathcal{X}_r(t,H)$ be the set of tokens generated in rollout $r$ whose simulated timestamps lie in the prediction interval $(t{+}365,\, t{+}H]$. For condition $c$ with code-set tokens $\mathcal{S}_c$, the rollout-derived label is
\begin{equation}
y_{r,c}(t,H)=\mathbbm{1}\!\left[\mathcal{X}_r(t,H)\cap\mathcal{S}_c\neq\emptyset\right].
\end{equation}
The Monte Carlo risk estimate is:
\begin{equation}
\hat{p}^{\text{roll}}_c(t,H)=\frac{1}{R}\sum_{r=1}^R y_{r,c}(t,H).
\end{equation}


\rev{To evaluate our medical language model baseline}, for each evaluation example, we construct a prompt containing (i) an instruction specifying the target condition and prediction horizon, (ii) the serialized patient history rendered as JSON, and (iii) a required JSON output schema asking the model to return a binary prediction together with a confidence score representing the estimated probability of disease onset within the prediction window. To reduce variance from stochastic decoding, we draw $K=50$ samples per patient window with temperature sampling ($T=0.7$, top-$p=0.95$) and aggregate to a final risk score by averaging the parsed confidence values across valid generations. Free-form generations are processed by a structured parser with rule-based fallbacks to extract the predicted label and confidence; generations that cannot be parsed are discarded. For patient histories that exceed the model context window, we truncate the serialized record by retaining the most recent portion of the sequence.

\subsection*{RAVEN inference}

Given an input window ending at time $t$, we perform a single forward pass to obtain condition risk at horizon $H$ by appending a separator token that encodes the desired future time. Specifically, we append a \texttt{<sep>} token whose positional embedding corresponds to $t+H$ (the last observed visit day plus the prediction horizon). The model then produces logits over the full token vocabulary at the \texttt{<sep>} position, representing the predicted set of clinical events at the next encounter occurring at time $t+H$ conditioned on the history up to $t$.

To obtain a scalar risk score for condition $c$ comprised of several tokens, we aggregate the model outputs over the condition's code set. Let $\mathcal{S}_c$ denote the set of token indices corresponding to diagnosis and/or medication codes used to define condition $c$. 
We compute the condition score by pooling logits over $\mathcal{S}_c$, yielding \rev{a horizon-conditioned risk estimate $\hat{p}_{c}(t, H) \in (0,1)$. \nrev{In our experiments, we sum the logits corresponding to the condition code set and apply a sigmoid.} We emphasize that the raw $\hat{p}_{c}(t, H)$ is a discriminative risk score rather than a formally calibrated cumulative-incidence estimate without additional modeling layers.}

\subsection*{Zero-shot disease forecasting evaluation}

We evaluate whether pretrained RAVEN can be used for zero-shot long-horizon disease-onset forecasting without any task-specific fine-tuning. The key idea is to convert long-horizon forecasting into a single next-visit prediction query by explicitly conditioning the model on a future time horizon via the visit-separator token. For each target condition $c$, we construct evaluation examples from longitudinal patient trajectories using rolling prediction windows. Each example consists of: (i) an input window containing a fixed-length history of past visits (365 days), (ii) a prediction horizon (e.g., 2 years or 5 years), and (iii) a binary label indicating whether the first onset of condition $c$ occurs within the specified horizon.

To further ensure that performance reflects true onset forecasting rather than repetition, we filter out any window where the patient already exhibits the target condition in the input history. We also exclude windows where the onset occurs within the first year after the anchor time (e.g. a gap of 1 year), which mitigates leakage from preclinical signals that may correspond to a diagnosis already underway.

We evaluate on seven disease endpoints: dementia, knee OA, pancreatic cancer, prostate cancer, acute MI, CHF, and COPD. Each endpoint is defined by a set of fine-grained diagnosis and/or medication codes; full code lists in Supplementary Table \ref{tab:disease_code}. \rev{For the primary zero-shot evaluations}, we do not perform additional cohort filtering or matching based on age, sex, or other demographic variables beyond the task-specific endpoint definitions. Certain conditions have low prevalence; for example, there are only 143 positive cases out of 283,814 examples for 2-year pancreatic cancer prediction. We benchmark our next-visit objective against three representative autoregressive paradigms: Multiclass, SeqLoss, and EGE. For these baselines, long-horizon risk is estimated via inference autoregressively with $R{=}\rev{64}$ rollouts per patient window, whereas RAVEN uses a single forward pass per example. We also compare against a fully fine-tuned BERT baseline for each task that was pretrained using the same dataset as RAVEN. 

\rev{To evaluate whether performance was driven primarily by broad demographic separation, we additionally performed a restricted risk-set analysis and full stratified evaluations for tasks with strong age or sex associations. Specifically, prostate cancer prediction was restricted to males, dementia to individuals aged $\geq65$ years, CHF and knee OA to individuals aged $\geq45$ years, and COPD and acute MI to individuals aged $\geq40$ years. These restrictions were applied identically to RAVEN and all simple supervised baselines. Pancreatic cancer was not included in the restricted comparison because additional stratification left few incident cases and unstable estimates.}

For the external EHRSHOT evaluation, we applied the same pretrained RAVEN model without any parameter updates or target-domain supervision. EHRSHOT defines six binary classification tasks predicting whether a patient will receive a first diagnosis of a condition within one year post-discharge from an inpatient visit: acute MI, lupus, hyperlipidemia, hypertension, celiac disease, and pancreatic cancer. Prevalence is low for several tasks, including celiac and lupus, which have very few positive test labels ($21$ and $20$, respectively). Prediction time is 11:59 PM on the day of discharge, and patients who already carry the diagnosis at prediction time are excluded. EHRSHOT does not impose any gap between prediction time and condition onset. 

We compared three RAVEN variants corresponding to different regularization strengths ($\lambda \in \{1.0, 0.5, 0.25\}$, where $\lambda = 1.0$ denotes no regularization) against four supervised baselines from EHRSHOT: CLMBR-T \cite{steinberg2021language}, gradient-boosted machines (GBM) \cite{ke2017lightgbm}, logistic regression, and random forest \cite{breiman2001random}, each trained with $K \in \{1, 2, 4, \ldots, 128\}$ labeled examples per class as well as the full training set.

\subsection*{Evaluation metrics}

Throughout the experiments, we report AUROC and AUPRC for each condition and horizon. To compute thresholded predictions, we select a condition-specific decision threshold $\tau_c$ on the validation set by maximizing the validation F1 score for condition $c$ (alternatively, we consider a prevalence-matched quantile thresholding scheme). Also, we obtain 95\% confidence intervals via a \rev{patient-level clustered bootstrap with 1000 resamples: we resample patients with replacement and include all evaluation windows from each sampled patient.} Given the low prevalence of certain conditions, we stratify the resampling \rev{by ever-positive versus never-positive patients}, which holds the number of positive \rev{patients} fixed across resamples.

Additionally, we report the on-time ratio to evaluate whether model predictions arrive before or at the time of first documented onset. For each patient and condition $c$, let $t^{\text{gt}}_c$ denote the time of the first ground-truth occurrence of any code in the condition's code set, and let $t^{\text{pred}}_c$ denote the first time the model predicts $c$ above the selected threshold $\tau_c$. Among patients counted as true positives, who developed the condition and predicted to have the condition along any evaluation windows, we define the on-time indicator as $\mathbbm{1}[t^{\text{pred}}_c \le t^{\text{gt}}_c]$. The on-time ratio is the fraction between true positive patients whose first prediction occurs at or before the first documented onset and the total patients who had the condition:
\[
\text{OnTime}(c) \;=\; \frac{\#\{ \text{TP patients with } t^{\text{pred}}_c \le t^{\text{gt}}_c\}}{\#\{\text{Positive patients}^*\}}.
\]
We define positive patients as those who eventually developed the condition, excluding patients who already had the condition at the first visit, since the model requires at least one prior visit as context. This metric directly quantifies whether the model provides clinically meaningful early warning rather than simply repeating an already-documented condition. 

\rev{Moreover, we assess calibration and clinical utility for the main RAVEN model. For a target condition $c$ and horizon, let $\hat{p}_i \in [0,1]$ be the predicted risk for evaluation window $i$, $y_i \in \{0,1\}$ the binary onset label, $N$ the number of windows, and $\pi_c = \tfrac{1}{N}\sum_i y_i$ the empirical prevalence. We quantify calibration with the Brier score $\mathrm{BS}(c) = \tfrac{1}{N}\sum_i (\hat{p}_i - y_i)^2$ and, because the Brier score is dominated by true negatives at low prevalence, the Brier skill score (BSS)}
\[
\mathrm{BSS}(c) \;=\; 1 - \frac{\mathrm{BS}(c)}{\mathrm{BS}_{\mathrm{ref}}(c)}, \qquad
\mathrm{BS}_{\mathrm{ref}}(c) = \frac{1}{N}\sum_i (\pi_c - y_i)^2,
\]
\rev{where the reference predicts the constant prevalence $\pi_c$ for every window; $\mathrm{BSS}(c) > 0$ indicates better probabilistic accuracy than this prevalence baseline. We also report the expected calibration error (ECE) and the calibration curves with the diagonal denoting perfect calibration. We also evaluate two standard post-hoc recalibration maps: Platt scaling and isotonic regression. Both maps are fit on the validation split and applied unchanged to the test split, so that no test labels inform recalibration. Platt scaling is strictly monotonic and therefore leaves AUROC and all rank-based metrics unchanged.}

\rev{To characterize utility at concrete operating points, we report extra performance summaries, all computed on the raw scores. Positive predictive value (PPV) at fixed sensitivity of $s = 0.5$ selects the highest threshold achieving sensitivity $\ge s$ and reports the precision at that threshold. Sensitivity at fixed alert burden $b \in \{0.01, 0.05\}$ flags the top-$b$ fraction of windows by predicted risk and reports the fraction of all positive windows captured. PPV at fixed alert burden uses the same top-$b$ flagged set and reports the fraction of flagged windows that are true positives. Top-$k$ precision reports the fraction of true onsets among the $k$ highest-ranked windows, with $k =100$. The no-skill value of every precision-type metric equals the prevalence $\pi_c$, which we report alongside each endpoint so that values can be read within context. As these operating-point metrics are rank-based, they are unchanged by the monotonic recalibration above.}

\newpage

\section*{Data availability}

The raw pretraining and internal evaluation datasets sourced from NYU Langone Health are not publicly available due to patient privacy reasons and institutional policy. Additional non-protected materials may be available from the corresponding author upon reasonable request, subject to institutional approval. The EHRSHOT dataset and benchmark are available via a research data use agreement from \url{https://som-shahlab.github.io/ehrshot-website/}.

\section*{Code availability}

The source code for RAVEN model training and evaluation, along with various baselines, is available at \url{https://github.com/xgao0o/RAVEN}. \nrev{The pretrained RAVEN model weights are available at \url{https://huggingface.co/NYUMedML/RAVEN}.} The pretrained BERT model is available at \url{https://huggingface.co/NYUMedML/EHRTransformer}.

\section*{Acknowledgments}



X.G., L.C., and N.R. are supported by the National Institutes of Health, National Institute on Aging award R01AG085617. H.R.R., C.M.D., and K.C. are supported by the National Institutes of Health under grant R01AR074453. N.R. is also supported by the National Institutes of Health, National Institute on Aging award R01AG079175.

\section*{Author contributions}

X.G. and H.R.R. contributed equally to this work. X.G. and H.R.R. conceived the study, conducted all experiments, analyzed the results, and wrote the manuscript. W.Z.,  S.H., G.S., and H.R.R. contributed to data preprocessing and initial buildout of the implementation. S.H. and H.R.R. contributed to LLM evaluations.  L.C. and X.G. contributed to data mapping and external evaluations. H.J., S.L., Y.W., H.Y., and X.G. contributed to scaling experiments and downstream evaluations. K.C. and C.M.D. provided guidance on study design and reviewed the paper. N.R. conceived the study, supervised the project, and revised the paper. All authors read and approved the final paper.

\section*{Competing interests}

The authors declare no competing interests.

\clearpage
\bibliography{ref}

\clearpage
\setcounter{figure}{0}
\setcounter{table}{0}
\setcounter{equation}{0}
\renewcommand{\thefigure}{S\arabic{figure}}
\renewcommand{\thetable}{S\arabic{table}}
\renewcommand{\theequation}{S\arabic{equation}}

\section*{Supplementary Information}

\subsection*{Disease code-set details for labeling}

For zero-shot evaluation, disease onset (label) was defined based on the first occurrence of any code from a predefined set of relevant diagnosis codes (ICD-10) and potentially medication codes associated with that condition within the specified 2- or 5-year prediction window. For instance, the dementia label relied on a group of specific ICD codes (e.g., G30.x, F01.x, F03.x) and dementia-related medication codes (e.g., RxNorm codes for donepezil, memantine). Similar specific code sets were defined for other conditions. The model's output \nrev{logits} for tokens in the target condition's code set were \nrev{summed and a sigmoid was applied to obtain} a single \rev{risk estimate for the condition}.

{\footnotesize
\begin{longtable}{ll p{11cm} l}
\caption[Supplementary disease code sets]{Diagnosis and medication codes for disease definition and labeling.} \\
\label{tab:disease_code}\\
\toprule
\textbf{Disease} & \textbf{Type} & \textbf{Description} & \textbf{Code} \\
\midrule
\endfirsthead
\multicolumn{4}{l}{\textit{(Continued from previous page)}} \\
\toprule
\textbf{Disease} & \textbf{Type} & \textbf{Description} & \textbf{Code} \\
\midrule
\endhead
\midrule
\multicolumn{4}{r}{\textit{(Continued on next page)}} \\
\endfoot
\bottomrule
\endlastfoot
Dementia & Diagnosis & Vascular dementia without behavioral disturbance & F01.50 \\
Dementia & Diagnosis & Vascular dementia with behavioral disturbance & F01.51 \\
Dementia & Diagnosis & Dementia in other diseases w/o behavioral disturbance & F02.80 \\
Dementia & Diagnosis & Dementia in other diseases w/ behavioral disturbance & F02.81 \\
Dementia & Diagnosis & Unspecified dementia without behavioral disturbance & F03.90 \\
Dementia & Diagnosis & Unspecified dementia with behavioral disturbance & F03.91 \\
Dementia & Diagnosis & Amnestic disorder due to physiological condition & F04 \\
Dementia & Diagnosis & Progressive supranuclear ophthalmoplegia & G23.1 \\
Dementia & Diagnosis & Alzheimer's disease with early onset & G30.0 \\
Dementia & Diagnosis & Alzheimer's disease with late onset & G30.1 \\
Dementia & Diagnosis & Other Alzheimer's disease & G30.8 \\
Dementia & Diagnosis & Alzheimer's disease, unspecified & G30.9 \\
Dementia & Diagnosis & Pick's disease & G31.01 \\
Dementia & Diagnosis & Other frontotemporal dementia & G31.09 \\
Dementia & Diagnosis & Senile degeneration of brain & G31.1 \\
Dementia & Diagnosis & Dementia with Lewy bodies & G31.83 \\
Dementia & Diagnosis & Mild cognitive impairment & G31.84 \\
Dementia & Diagnosis & Corticobasal degeneration & G31.85 \\
Dementia & Diagnosis & Degenerative disease of nervous system, unspecified & G31.9 \\
Dementia & Medication & RIVASTIGMINE TARTRATE 1.5 MG ORAL CAP & 57619 \\
Dementia & Medication & MEMANTINE 10 MG ORAL TAB & 30323 \\
Dementia & Medication & GALANTAMINE 4 MG ORAL TAB & 1232 \\
Dementia & Medication & DONEPEZIL 10 MG ORAL TAB & 31624 \\
Dementia & Medication & DONEPEZIL 5 MG ORAL TAB & 31811 \\
Dementia & Medication & GALANTAMINE 24 MG ORAL C24P & 31774 \\
Dementia & Medication & DONEPEZIL 5 MG ORAL TBDL & 31866 \\
Dementia & Medication & RIVASTIGMINE TARTRATE 3 MG ORAL CAP & 57575 \\
Dementia & Medication & RIVASTIGMINE TARTRATE 6 MG ORAL CAP & 58034 \\
Dementia & Medication & DONEPEZIL ORAL & 60609 \\
Dementia & Medication & GALANTAMINE 8 MG ORAL TAB & 6685 \\
Dementia & Medication & MEMANTINE ORAL & 73925 \\
Dementia & Medication & DONEPEZIL 10 MG ORAL TBDL & 31609 \\
Dementia & Medication & RIVASTIGMINE 4.6 MG/24 HR TRANSDERMAL PT24 & 32859 \\
Dementia & Medication & MEMANTINE 5 MG ORAL TAB & 20149 \\
Dementia & Medication & GALANTAMINE 8 MG ORAL C24P & 31828 \\
Dementia & Medication & MEMANTINE 5-10 MG ORAL DSPK & 20151 \\
Dementia & Medication & RIVASTIGMINE TARTRATE 4.5 MG ORAL CAP & 57679 \\
Dementia & Medication & GALANTAMINE 16 MG ORAL C24P & 31604 \\
Dementia & Medication & RIVASTIGMINE 9.5 MG/24 HR TRANSDERMAL PT24 & 33175 \\
Dementia & Medication & MEMANTINE-DONEPEZIL 28-10 MG ORAL CSPX & 143836 \\
Dementia & Medication & MEMANTINE-DONEPEZIL 7-10 MG ORAL CSPX & 150473 \\
Dementia & Medication & MEMANTINE-DONEPEZIL 14-10 MG ORAL CSPX & 143871 \\
Dementia & Medication & MEMANTINE 28 MG ORAL CSPX & 101254 \\
Dementia & Medication & MEMANTINE-DONEPEZIL 21-10 MG ORAL CSPX & 150139 \\
Dementia & Medication & MEMANTINE 7 MG ORAL CSPX & 112860 \\
Dementia & Medication & MEMANTINE 7-14-21-28 MG ORAL C24K & 101328 \\
Dementia & Medication & DONEPEZIL 23 MG ORAL TAB & 111311 \\
Dementia & Medication & MEMANTINE 14 MG ORAL CSPX & 101112 \\
Dementia & Medication & MEMANTINE 21 MG ORAL CSPX & 101141 \\
Dementia & Medication & RIVASTIGMINE 13.3 MG/24 HR TRANSDERMAL PT24 & 98524 \\
Knee OA & Diagnosis & Bilateral primary osteoarthritis of knee & M17.0 \\
Knee OA & Diagnosis & Unilateral primary osteoarthritis, unspecified knee & M17.10 \\
Knee OA & Diagnosis & Unilateral primary osteoarthritis, right knee & M17.11 \\
Knee OA & Diagnosis & Unilateral primary osteoarthritis, left knee & M17.12 \\
Knee OA & Diagnosis & Bilateral post-traumatic osteoarthritis of knee & M17.2 \\
Knee OA & Diagnosis & Unilateral post-traumatic osteoarthritis, unspecified & M17.30 \\
Knee OA & Diagnosis & Unilateral post-traumatic osteoarthritis, right knee & M17.31 \\
Knee OA & Diagnosis & Unilateral post-traumatic osteoarthritis, left knee & M17.32 \\
Knee OA & Diagnosis & Other bilateral secondary osteoarthritis of knee & M17.4 \\
Knee OA & Diagnosis & Other unilateral secondary osteoarthritis of knee & M17.5 \\
Knee OA & Diagnosis & Osteoarthritis of knee, unspecified & M17.9 \\
Prostate Cancer & Diagnosis & Malignant neoplasm of prostate & C61 \\
Prostate Cancer & Diagnosis & Carcinoma in situ of prostate & D07.5 \\
Prostate Cancer & Diagnosis & Personal history of malignant neoplasm of prostate & Z85.46 \\
Acute MI & Diagnosis & Acute myocardial infarction, unspecified & I21.9 \\
Acute MI & Diagnosis & Myocardial infarction type 2 & I21.A1 \\
Acute MI & Diagnosis & Other myocardial infarction type & I21.A9 \\
Acute MI & Diagnosis & ST elevation (STEMI) myocardial infarction involving left main coronary artery & I21.01 \\
Acute MI & Diagnosis & ST elevation (STEMI) myocardial infarction involving left anterior descending coronary artery & I21.02 \\
Acute MI & Diagnosis & ST elevation (STEMI) myocardial infarction involving other coronary artery of anterior wall & I21.09 \\
Acute MI & Diagnosis & ST elevation (STEMI) myocardial infarction involving right coronary artery & I21.11 \\
Acute MI & Diagnosis & ST elevation (STEMI) myocardial infarction involving other coronary artery of inferior wall & I21.19 \\
Acute MI & Diagnosis & ST elevation (STEMI) myocardial infarction involving left circumflex coronary artery & I21.21 \\
Acute MI & Diagnosis & ST elevation (STEMI) myocardial infarction involving other sites & I21.29 \\
Acute MI & Diagnosis & ST elevation (STEMI) myocardial infarction of unspecified site & I21.3 \\
Acute MI & Diagnosis & Non-ST elevation (NSTEMI) myocardial infarction & I21.4 \\
Acute MI & Diagnosis & Subsequent ST elevation (STEMI) myocardial infarction of anterior wall & I22.0 \\
Acute MI & Diagnosis & Subsequent ST elevation (STEMI) myocardial infarction of inferior wall & I22.1 \\
Acute MI & Diagnosis & Subsequent non-ST elevation (NSTEMI) myocardial infarction & I22.2 \\
Acute MI & Diagnosis & Subsequent ST elevation (STEMI) myocardial infarction of other sites & I22.8 \\
Acute MI & Diagnosis & Subsequent ST elevation (STEMI) myocardial infarction of unspecified site & I22.9 \\
Acute MI & Diagnosis & Hemopericardium as current complication following acute myocardial infarction & I23.0 \\
Acute MI & Diagnosis & Atrial septal defect as current complication following acute myocardial infarction & I23.1 \\
Acute MI & Diagnosis & Rupture of cardiac wall without hemopericardium as current complication following acute myocardial infarction & I23.3 \\
Acute MI & Diagnosis & Rupture of chordae tendineae as current complication following acute myocardial infarction & I23.4 \\
Acute MI & Diagnosis & Thrombosis of atrium, auricular appendage, and ventricle as current complications following acute myocardial infarction & I23.6 \\
Acute MI & Diagnosis & Postinfarction angina & I23.7 \\
Acute MI & Diagnosis & Other current complications following acute myocardial infarction & I23.8 \\
Pancreatic Cancer & Diagnosis & Malignant neoplasm of head of pancreas & C25.0 \\
Pancreatic Cancer & Diagnosis & Malignant neoplasm of body of pancreas & C25.1 \\
Pancreatic Cancer & Diagnosis & Malignant neoplasm of tail of pancreas & C25.2 \\
Pancreatic Cancer & Diagnosis & Malignant neoplasm of pancreatic duct & C25.3 \\
Pancreatic Cancer & Diagnosis & Malignant neoplasm of other parts of pancreas & C25.7 \\
Pancreatic Cancer & Diagnosis & Malignant neoplasm of overlapping sites of pancreas & C25.8 \\
Pancreatic Cancer & Diagnosis & Malignant neoplasm of pancreas, unspecified & C25.9 \\
CHF & Diagnosis & Right heart failure, unspecified & I50.810 \\
CHF & Diagnosis & Acute right heart failure & I50.811 \\
CHF & Diagnosis & Chronic right heart failure & I50.812 \\
CHF & Diagnosis & Acute on chronic right heart failure & I50.813 \\
CHF & Diagnosis & Right heart failure due to left heart failure & I50.814 \\
CHF & Diagnosis & Biventricular heart failure & I50.82 \\
CHF & Diagnosis & End stage heart failure & I50.84 \\
CHF & Diagnosis & Other heart failure & I50.89 \\
CHF & Diagnosis & Rheumatic heart failure & I09.81 \\
CHF & Diagnosis & Hypertensive heart disease with heart failure & I11.0 \\
CHF & Diagnosis & Hypertensive heart and chronic kidney disease with heart failure and stage 1 through stage 4 chronic kidney disease, or unspecified chronic kidney disease & I13.0 \\
CHF & Diagnosis & Hypertensive heart and chronic kidney disease with heart failure and with stage 5 chronic kidney disease, or end stage renal disease & I13.2 \\
CHF & Diagnosis & Left ventricular failure & I50.1 \\
CHF & Diagnosis & Unspecified systolic (congestive) heart failure & I50.20 \\
CHF & Diagnosis & Acute systolic (congestive) heart failure & I50.21 \\
CHF & Diagnosis & Chronic systolic (congestive) heart failure & I50.22 \\
CHF & Diagnosis & Acute on chronic systolic (congestive) heart failure & I50.23 \\
CHF & Diagnosis & Unspecified diastolic (congestive) heart failure & I50.30 \\
CHF & Diagnosis & Acute diastolic (congestive) heart failure & I50.31 \\
CHF & Diagnosis & Chronic diastolic (congestive) heart failure & I50.32 \\
CHF & Diagnosis & Acute on chronic diastolic (congestive) heart failure & I50.33 \\
CHF & Diagnosis & Unspecified combined systolic (congestive) and diastolic (congestive) heart failure & I50.40 \\
CHF & Diagnosis & Acute combined systolic (congestive) and diastolic (congestive) heart failure & I50.41 \\
CHF & Diagnosis & Chronic combined systolic (congestive) and diastolic (congestive) heart failure & I50.42 \\
CHF & Diagnosis & Acute on chronic combined systolic (congestive) and diastolic (congestive) heart failure & I50.43 \\
CHF & Diagnosis & Heart failure, unspecified & I50.9 \\
COPD & Diagnosis & Bronchitis, not specified as acute or chronic & J40 \\
COPD & Diagnosis & Simple chronic bronchitis & J41.0 \\
COPD & Diagnosis & Mucopurulent chronic bronchitis & J41.1 \\
COPD & Diagnosis & Mixed simple and mucopurulent chronic bronchitis & J41.8 \\
COPD & Diagnosis & Unspecified chronic bronchitis & J42 \\
COPD & Diagnosis & Unilateral pulmonary emphysema [MacLeod’s syndrome] & J43.0 \\
COPD & Diagnosis & Panlobular emphysema & J43.1 \\
COPD & Diagnosis & Centrilobular emphysema & J43.2 \\
COPD & Diagnosis & Emphysema, unspecified & J43.9 \\
COPD & Diagnosis & Chronic obstructive pulmonary disease with acute lower respiratory infection & J44.0 \\
COPD & Diagnosis & Chronic obstructive pulmonary disease with (acute) exacerbation & J44.1 \\
COPD & Diagnosis & Chronic obstructive pulmonary disease, unspecified & J44.9 \\
COPD & Diagnosis & Bronchiectasis with acute lower respiratory infection & J47.0 \\
COPD & Diagnosis & Bronchiectasis with (acute) exacerbation & J47.1 \\
COPD & Diagnosis & Bronchiectasis, uncomplicated & J47.9 \\
COPD & Diagnosis & Interstitial emphysema & J98.2 \\
\end{longtable}
}

\newpage

\subsection*{Hyperparameters and configurations}

We present the training hyperparameters for the main 144M model on the full data in Table \ref{tab:hyper}. The hyperparameters are largely the same among models trained on different data budgets except for the number of iterations for training and learning rate decay. Table \ref{tab:model_configs} contains architecture configurations for different model sizes used in the scaling analysis.

\begin{longtable}{ll}
\caption[Supplementary training hyperparameters]{Training hyperparameters and configuration settings for the main 144M model.} \\
\label{tab:hyper}\\
\toprule
\textbf{Hyperparameter} & \textbf{Value} \\
\midrule
\endfirsthead
\multicolumn{2}{l}{\textit{(Continued from previous page)}} \\
\toprule
\textbf{Hyperparameter} & \textbf{Value} \\
\midrule
\endhead
\midrule
\multicolumn{2}{r}{\textit{(Continued on next page)}} \\
\endfoot
\bottomrule
\endlastfoot
\texttt{n\_embd} & 1024 \\
\texttt{n\_head} & 8 \\
\texttt{n\_layer} & 8 \\
\texttt{n\_tokens} & 42,337 \\
\texttt{bias} & false \\
\texttt{dropout} & 0 \\
\texttt{block\_size} & 512 \\
\texttt{batch\_size} & 16 \\
\texttt{optimizer} & AdamW \\
\texttt{beta1} & 0.9 \\
\texttt{beta2} & 0.95 \\
\texttt{decay\_lr} & true \\
\texttt{grad\_clip} & 1 \\
\texttt{gradient\_accumulation\_steps} & 8 \\
\texttt{learning\_rate} & 0.00022 \\
\texttt{lr\_decay\_iters} & 800,000 \\
\texttt{max\_iters} & 810,000 \\
\texttt{min\_lr} & 0.000022 \\
\texttt{rotary} & true \\
\texttt{temporal\_decay} & 0.5 \\
\texttt{warmup\_iters} & 20,000 \\
\texttt{weight\_decay} & 0.01 \\
\end{longtable}

\begin{longtable}{rrrr}
\caption{Model configurations used in the scaling experiments.} \label{tab:model_configs} \\
\toprule
\textbf{Layers} & \textbf{Heads} & \textbf{Embedding Dim} & \textbf{Parameters} \\
\midrule
\endfirsthead
\toprule
\textbf{Layers} & \textbf{Heads} & \textbf{Embedding Dim} & \textbf{Parameters} \\
\midrule
\endhead
\midrule
\multicolumn{4}{r}{\textit{Continued on next page}} \\
\endfoot
\bottomrule
\endlastfoot
4  & 4  & 16   & 0.69M  \\
2  & 2  & 64   & 2.81M  \\
2  & 2  & 128  & 5.81M  \\
4  & 4  & 256  & 13.99M \\
\midrule
8  & 8  & 512  & 46.85M  \\
8  & 8  & 1024 & 144.04M \\
32 & 32 & 1024 & 446.08M \\
64 & 64 & 1024 & 848.80M \\
\end{longtable}

\subsection*{Zero-shot tasks statistics}

In Table \ref{tab:eval_cohort_stats}, we report task-specific test set information across horizons and conditions \rev{at both the window (example) and patient level}. \rev{Each evaluation example is a single rolling window extracted from a patient and prediction time. Since a patient may contribute several windows, the two levels give complementary measures of cohort size.} \rev{Window-level} prevalence varies dramatically across conditions and horizons: pancreatic cancer at the 2-year window is just 0.05\%, while knee OA at 5 years reaches 10.49\%.

{
\scriptsize
\begin{table}[ht]
\centering
\caption{Evaluation cohort statistics for zero-shot disease-onset forecasting tasks on the held-out test set. \rev{Because a patient may contribute multiple windows, we report counts and prevalence at both the \emph{window} and \emph{patient} level. A patient is counted as positive if at least one of their windows is positive, so patient-level prevalence is generally higher than window-level prevalence.} Prevalence denotes the fraction of positive windows / patients in each task. We note that, for zero shot evaluations, we exclude those who already have the condition by prediction time and by prediction time + 1 year gap window, and therefore these rates represent fractions of "new onset" within the prediction window.}
\label{tab:eval_cohort_stats}
\small
\setlength{\tabcolsep}{4pt}
\resizebox{\linewidth}{!}{%
\begin{tabular}{llrrc|rrc}
\toprule
& & \multicolumn{3}{c|}{\textbf{Windows}} & \multicolumn{3}{c}{\textbf{Patients}} \\
\cmidrule(lr){3-5} \cmidrule(lr){6-8}
\textbf{Condition} & \textbf{Horizon} & \textbf{Total ($n$)} & \textbf{Positive} & \textbf{Prevalence (\%)} & \textbf{Total ($n$)} & \textbf{Positive} & \textbf{Prevalence (\%)} \\
\midrule
COPD              & 2-year & 239{,}463 & 5{,}255 & 2.19  & 77{,}315 & 3{,}666 & 4.74  \\
COPD              & 5-year & 47{,}627  & 4{,}251 & 8.93  & 21{,}870 & 2{,}109 & 9.64  \\
\midrule
CHF               & 2-year & 263{,}220 & 2{,}874 & 1.09  & 81{,}259 & 1{,}989 & 2.45  \\
CHF               & 5-year & 52{,}072  & 2{,}168 & 4.16  & 23{,}353 & 1{,}044 & 4.47  \\
\midrule
Dementia          & 2-year & 271{,}172 & 2{,}021 & 0.75  & 82{,}645 & 1{,}395 & 1.69  \\
Dementia          & 5-year & 53{,}520  & 1{,}685 & 3.15  & 23{,}795 & 786     & 3.30  \\
\midrule
Pancreatic Cancer & 2-year & 283{,}814 & 143     & 0.05  & 84{,}642 & 103     & 0.12  \\
Pancreatic Cancer & 5-year & 55{,}363  & 82      & 0.15  & 24{,}281 & 39      & 0.16  \\
\midrule
Prostate Cancer   & 2-year & 275{,}208 & 887     & 0.32  & 82{,}999 & 623     & 0.75  \\
Prostate Cancer   & 5-year & 53{,}698  & 708     & 1.32  & 23{,}736 & 360     & 1.52  \\
\midrule
Acute MI          & 2-year & 269{,}442 & 2{,}270 & 0.84  & 82{,}488 & 1{,}559 & 1.89  \\
Acute MI          & 5-year & 52{,}816  & 1{,}651 & 3.13  & 23{,}576 & 802     & 3.40  \\
\midrule
Knee OA           & 2-year & 240{,}857 & 5{,}808 & 2.41  & 77{,}497 & 4{,}061 & 5.24  \\
Knee OA           & 5-year & 48{,}225  & 5{,}060 & 10.49 & 22{,}087 & 2{,}484 & 11.25 \\
\bottomrule
\end{tabular}%
}
\end{table}
}

\newpage

\subsection*{Full zero-shot forecasting results}

This supplementary section provides a complete view of RAVEN model performance across seven disease onsets, using two prediction horizons (2-year versus 5-year). We include the full experimental results with 95\% confidence intervals using different regularization strengths, model size, and data budgets. Also, we provide full results for baselines appeared in Table \ref{tab:zs_2y_all_baselines} and \ref{tab:zs_5y_all_baselines}.

\subsubsection*{Regularization strengths}

%

{
\scriptsize
\setlength{\tabcolsep}{3.5pt}
\begin{longtable}{llcccccc}
\caption[Zero-shot forecasting with $\lambda=0.25$ (144M, full data)]{Zero-shot forecasting with $\lambda=0.25$ for the 144M model trained on full data.}
\label{tab:zs_reg025}\\
\toprule
\textbf{Condition} & \textbf{Hor.} & \textbf{AUROC} & \textbf{AUPRC} & \textbf{F1} & \textbf{Precision} & \textbf{Recall} \\
\midrule
\endfirsthead
\multicolumn{7}{c}{\tablename\ \thetable\ -- \textit{continued}} \\
\toprule
\textbf{Condition} & \textbf{Hor.} & \textbf{AUROC} & \textbf{AUPRC} & \textbf{F1} & \textbf{Precision} & \textbf{Recall} \\
\midrule
\endhead
\midrule \multicolumn{7}{r}{\textit{continued on next page}} \\
\endfoot
\bottomrule
\endlastfoot
CHF       & 2y & 0.854 [0.847, 0.861] & 0.078 [0.070, 0.086] & 0.140 [0.131, 0.151] & 0.097 [0.090, 0.105] & 0.254 [0.237, 0.274] \\
          & 5y & 0.826 [0.811, 0.840] & 0.189 [0.170, 0.212] & 0.267 [0.248, 0.286] & 0.190 [0.176, 0.205] & 0.445 [0.411, 0.479] \\
COPD      & 2y & 0.687 [0.678, 0.696] & 0.050 [0.047, 0.054] & 0.096 [0.090, 0.103] & 0.066 [0.062, 0.071] & 0.179 [0.167, 0.191] \\
          & 5y & 0.659 [0.647, 0.673] & 0.155 [0.146, 0.166] & 0.222 [0.210, 0.234] & 0.153 [0.145, 0.162] & 0.404 [0.383, 0.426] \\
Dementia  & 2y & 0.781 [0.769, 0.795] & 0.035 [0.031, 0.040] & 0.082 [0.075, 0.091] & 0.051 [0.046, 0.056] & 0.213 [0.194, 0.235] \\
          & 5y & 0.787 [0.771, 0.803] & 0.121 [0.104, 0.141] & 0.180 [0.160, 0.200] & 0.136 [0.121, 0.152] & 0.268 [0.237, 0.300] \\
Acute MI & 2y & 0.794 [0.782, 0.805] & 0.030 [0.028, 0.033] & 0.068 [0.061, 0.076] & 0.042 [0.038, 0.047] & 0.177 [0.159, 0.196] \\
         & 5y & 0.756 [0.738, 0.772] & 0.078 [0.071, 0.087] & 0.147 [0.134, 0.159] & 0.089 [0.081, 0.097] & 0.413 [0.378, 0.446] \\
Knee OA   & 2y & 0.722 [0.715, 0.729] & 0.053 [0.051, 0.056] & 0.098 [0.093, 0.104] & 0.062 [0.059, 0.065] & 0.242 [0.230, 0.255] \\
          & 5y & 0.687 [0.676, 0.698] & 0.182 [0.174, 0.194] & 0.266 [0.257, 0.276] & 0.172 [0.166, 0.179] & 0.584 [0.565, 0.605] \\
Panc.\ Cancer & 2y & 0.676 [0.620, 0.731] & 0.002 [0.001, 0.005] & 0.010 [0.000, 0.032] & 0.015 [0.000, 0.055] & 0.007 [0.000, 0.022] \\
              & 5y & 0.617 [0.523, 0.721] & 0.010 [0.002, 0.043] & 0.010 [0.000, 0.025] & 0.005 [0.000, 0.014] & 0.061 [0.000, 0.154] \\
Prostate Ca. & 2y & 0.911 [0.905, 0.918] & 0.022 [0.020, 0.026] & 0.031 [0.019, 0.044] & 0.032 [0.019, 0.045] & 0.030 [0.018, 0.044] \\
             & 5y & 0.901 [0.892, 0.909] & 0.071 [0.062, 0.085] & 0.115 [0.096, 0.136] & 0.076 [0.062, 0.090] & 0.242 [0.202, 0.287] \\
\end{longtable}
}


{
\scriptsize
\setlength{\tabcolsep}{3.5pt}
\begin{longtable}{llcccccc}
\caption[Zero-shot forecasting with $\lambda^\star=0.5$ (144M, full data)]{Zero-shot forecasting with $\lambda^\star=0.5$ for the 144M model trained on full data.}
\label{tab:zs_reg050}\\
\toprule
\textbf{Condition} & \textbf{Hor.} & \textbf{AUROC} & \textbf{AUPRC} & \textbf{F1} & \textbf{Precision} & \textbf{Recall} \\
\midrule
\endfirsthead
\multicolumn{7}{c}{\tablename\ \thetable\ -- \textit{continued}} \\
\toprule
\textbf{Condition} & \textbf{Hor.} & \textbf{AUROC} & \textbf{AUPRC} & \textbf{F1} & \textbf{Precision} & \textbf{Recall} \\
\midrule
\endhead
\midrule \multicolumn{7}{r}{\textit{continued on next page}} \\
\endfoot
\bottomrule
\endlastfoot
CHF       & 2y & 0.857 [0.850, 0.864] & 0.078 [0.071, 0.087] & 0.147 [0.135, 0.157] & 0.111 [0.102, 0.120] & 0.216 [0.200, 0.232] \\
          & 5y & 0.821 [0.806, 0.835] & 0.184 [0.165, 0.205] & 0.257 [0.238, 0.276] & 0.186 [0.172, 0.201] & 0.414 [0.384, 0.446] \\
COPD      & 2y & 0.691 [0.683, 0.701] & 0.052 [0.049, 0.056] & 0.100 [0.094, 0.106] & 0.068 [0.063, 0.072] & 0.193 [0.182, 0.205] \\
          & 5y & 0.676 [0.664, 0.689] & 0.164 [0.154, 0.175] & 0.238 [0.226, 0.249] & 0.163 [0.154, 0.171] & 0.441 [0.419, 0.462] \\
Dementia  & 2y & 0.789 [0.775, 0.801] & 0.037 [0.033, 0.043] & 0.083 [0.074, 0.091] & 0.053 [0.048, 0.059] & 0.182 [0.164, 0.202] \\
          & 5y & 0.773 [0.756, 0.791] & 0.108 [0.096, 0.125] & 0.182 [0.163, 0.203] & 0.136 [0.121, 0.152] & 0.274 [0.244, 0.306] \\
Acute MI & 2y & 0.793 [0.782, 0.804] & 0.031 [0.029, 0.034] & 0.075 [0.068, 0.083] & 0.046 [0.042, 0.051] & 0.194 [0.175, 0.214] \\
         & 5y & 0.754 [0.736, 0.769] & 0.080 [0.072, 0.089] & 0.149 [0.136, 0.161] & 0.090 [0.082, 0.098] & 0.423 [0.388, 0.456] \\
Knee OA   & 2y & 0.726 [0.718, 0.733] & 0.057 [0.054, 0.060] & 0.107 [0.102, 0.112] & 0.065 [0.062, 0.068] & 0.299 [0.285, 0.312] \\
          & 5y & 0.697 [0.686, 0.709] & 0.200 [0.190, 0.213] & 0.272 [0.263, 0.283] & 0.183 [0.176, 0.191] & 0.534 [0.515, 0.555] \\
Panc.\ Cancer & 2y & 0.693 [0.637, 0.745] & 0.002 [0.001, 0.004] & 0.007 [0.000, 0.023] & 0.007 [0.000, 0.024] & 0.007 [0.000, 0.022] \\
              & 5y & 0.639 [0.548, 0.732] & 0.004 [0.002, 0.021] & 0.008 [0.004, 0.014] & 0.004 [0.002, 0.008] & 0.134 [0.054, 0.240] \\
Prostate Ca. & 2y & 0.910 [0.904, 0.917] & 0.022 [0.020, 0.026] & 0.043 [0.031, 0.054] & 0.030 [0.022, 0.039] & 0.072 [0.053, 0.091] \\
             & 5y & 0.901 [0.893, 0.910] & 0.073 [0.063, 0.087] & 0.108 [0.082, 0.137] & 0.089 [0.068, 0.113] & 0.138 [0.104, 0.178] \\
\end{longtable}
}

\newpage
{
\scriptsize
\setlength{\tabcolsep}{3.5pt}
\begin{longtable}{llcccccc}
\caption[Zero-shot forecasting with $\lambda=0.75$ (144M, full data)]{Zero-shot forecasting with $\lambda=0.75$ for the 144M model trained on full data.}
\label{tab:zs_reg075}\\
\toprule
\textbf{Condition} & \textbf{Hor.} & \textbf{AUROC} & \textbf{AUPRC} & \textbf{F1} & \textbf{Precision} & \textbf{Recall} \\
\midrule
\endfirsthead
\multicolumn{7}{c}{\tablename\ \thetable\ -- \textit{continued}} \\
\toprule
\textbf{Condition} & \textbf{Hor.} & \textbf{AUROC} & \textbf{AUPRC} & \textbf{F1} & \textbf{Precision} & \textbf{Recall} \\
\midrule
\endhead
\midrule \multicolumn{7}{r}{\textit{continued on next page}} \\
\endfoot
\bottomrule
\endlastfoot
CHF       & 2y & 0.855 [0.847, 0.863] & 0.075 [0.069, 0.083] & 0.138 [0.128, 0.148] & 0.096 [0.088, 0.103] & 0.246 [0.229, 0.263] \\
          & 5y & 0.824 [0.809, 0.839] & 0.189 [0.170, 0.214] & 0.273 [0.252, 0.292] & 0.202 [0.185, 0.219] & 0.418 [0.386, 0.451] \\
COPD      & 2y & 0.686 [0.677, 0.695] & 0.050 [0.047, 0.054] & 0.097 [0.091, 0.105] & 0.072 [0.066, 0.077] & 0.152 [0.141, 0.164] \\
          & 5y & 0.668 [0.655, 0.681] & 0.162 [0.152, 0.173] & 0.231 [0.221, 0.241] & 0.146 [0.140, 0.153] & 0.551 [0.530, 0.573] \\
Dementia  & 2y & 0.766 [0.751, 0.779] & 0.034 [0.030, 0.040] & 0.081 [0.072, 0.091] & 0.057 [0.050, 0.064] & 0.142 [0.125, 0.160] \\
          & 5y & 0.761 [0.743, 0.780] & 0.109 [0.095, 0.127] & 0.182 [0.163, 0.201] & 0.129 [0.115, 0.142] & 0.309 [0.276, 0.342] \\
Acute MI & 2y & 0.790 [0.779, 0.799] & 0.031 [0.028, 0.034] & 0.070 [0.063, 0.076] & 0.041 [0.037, 0.045] & 0.228 [0.208, 0.247] \\
         & 5y & 0.739 [0.722, 0.755] & 0.074 [0.068, 0.083] & 0.134 [0.123, 0.144] & 0.079 [0.073, 0.085] & 0.448 [0.415, 0.479] \\
Knee OA   & 2y & 0.711 [0.704, 0.718] & 0.051 [0.049, 0.054] & 0.098 [0.093, 0.103] & 0.061 [0.058, 0.064] & 0.254 [0.241, 0.267] \\
          & 5y & 0.678 [0.667, 0.689] & 0.177 [0.168, 0.187] & 0.255 [0.246, 0.264] & 0.167 [0.161, 0.174] & 0.538 [0.518, 0.558] \\
Panc.\ Cancer & 2y & 0.701 [0.648, 0.751] & 0.002 [0.001, 0.003] & 0.000 [0.000, 0.000] & 0.000 [0.000, 0.000] & 0.000 [0.000, 0.000] \\
              & 5y & 0.698 [0.619, 0.782] & 0.013 [0.003, 0.064] & 0.010 [0.004, 0.016] & 0.005 [0.002, 0.008] & 0.195 [0.084, 0.326] \\
Prostate Ca. & 2y & 0.910 [0.904, 0.916] & 0.021 [0.019, 0.024] & 0.042 [0.034, 0.050] & 0.025 [0.020, 0.029] & 0.133 [0.109, 0.157] \\
             & 5y & 0.900 [0.891, 0.908] & 0.071 [0.061, 0.088] & 0.118 [0.099, 0.137] & 0.074 [0.061, 0.086] & 0.292 [0.248, 0.340] \\
\end{longtable}
}

{
\scriptsize
\setlength{\tabcolsep}{3.5pt}
\begin{longtable}{llcccccc}
\caption[Zero-shot forecasting with $\lambda=1.0$ (144M, full data)]{Zero-shot forecasting with $\lambda=1.0$ (no reg.) for the 144M model trained on full data.}
\label{tab:zs_reg100}\\
\toprule
\textbf{Condition} & \textbf{Hor.} & \textbf{AUROC} & \textbf{AUPRC} & \textbf{F1} & \textbf{Precision} & \textbf{Recall} \\
\midrule
\endfirsthead
\multicolumn{7}{c}{\tablename\ \thetable\ -- \textit{continued}} \\
\toprule
\textbf{Condition} & \textbf{Hor.} & \textbf{AUROC} & \textbf{AUPRC} & \textbf{F1} & \textbf{Precision} & \textbf{Recall} \\
\midrule
\endhead
\midrule \multicolumn{7}{r}{\textit{continued on next page}} \\
\endfoot
\bottomrule
\endlastfoot
CHF       & 2y & 0.856 [0.849, 0.863] & 0.073 [0.067, 0.081] & 0.139 [0.128, 0.149] & 0.104 [0.096, 0.112] & 0.207 [0.190, 0.223] \\
          & 5y & 0.830 [0.817, 0.843] & 0.186 [0.169, 0.207] & 0.268 [0.247, 0.288] & 0.202 [0.186, 0.218] & 0.396 [0.363, 0.428] \\
COPD      & 2y & 0.692 [0.684, 0.701] & 0.052 [0.049, 0.055] & 0.102 [0.095, 0.109] & 0.071 [0.066, 0.076] & 0.181 [0.169, 0.194] \\
          & 5y & 0.676 [0.663, 0.689] & 0.173 [0.162, 0.185] & 0.240 [0.228, 0.252] & 0.167 [0.158, 0.177] & 0.423 [0.401, 0.445] \\
Dementia  & 2y & 0.772 [0.758, 0.785] & 0.036 [0.033, 0.042] & 0.087 [0.078, 0.098] & 0.059 [0.053, 0.066] & 0.167 [0.149, 0.187] \\
          & 5y & 0.770 [0.753, 0.788] & 0.110 [0.097, 0.127] & 0.179 [0.163, 0.198] & 0.127 [0.115, 0.140] & 0.305 [0.276, 0.337] \\
Acute MI & 2y & 0.775 [0.764, 0.787] & 0.032 [0.029, 0.035] & 0.074 [0.066, 0.082] & 0.050 [0.045, 0.056] & 0.143 [0.128, 0.159] \\
         & 5y & 0.745 [0.726, 0.762] & 0.081 [0.073, 0.091] & 0.145 [0.133, 0.157] & 0.086 [0.079, 0.094] & 0.448 [0.411, 0.482] \\
Knee OA   & 2y & 0.706 [0.699, 0.713] & 0.051 [0.048, 0.053] & 0.097 [0.091, 0.103] & 0.067 [0.063, 0.071] & 0.174 [0.163, 0.185] \\
          & 5y & 0.674 [0.664, 0.685] & 0.181 [0.172, 0.192] & 0.257 [0.248, 0.267] & 0.171 [0.165, 0.179] & 0.513 [0.495, 0.532] \\
Panc.\ Cancer & 2y & 0.692 [0.638, 0.746] & 0.002 [0.001, 0.003] & 0.000 [0.000, 0.000] & 0.000 [0.000, 0.000] & 0.000 [0.000, 0.000] \\
              & 5y & 0.663 [0.572, 0.748] & 0.004 [0.002, 0.007] & 0.007 [0.000, 0.022] & 0.005 [0.000, 0.015] & 0.012 [0.000, 0.042] \\
Prostate Ca. & 2y & 0.908 [0.902, 0.914] & 0.021 [0.018, 0.023] & 0.035 [0.024, 0.046] & 0.027 [0.019, 0.035] & 0.051 [0.035, 0.068] \\
             & 5y & 0.897 [0.889, 0.906] & 0.069 [0.060, 0.082] & 0.108 [0.089, 0.128] & 0.070 [0.057, 0.083] & 0.240 [0.197, 0.284] \\
\end{longtable}
}

\newpage
\subsubsection*{Data budgets and model capacity}

{
\scriptsize
\setlength{\tabcolsep}{4pt}
\begin{longtable}{llccccc}
\caption[Macro-averaged zero-shot forecasting across model sizes]{Macro-averaged zero-shot forecasting across model sizes (full data, $\lambda=0.5$).}
\label{tab:zs_model_size_macro_ci}\\
\toprule
\textbf{Model size} & \textbf{Horizon} & \textbf{AUROC} & \textbf{AUPRC} & \textbf{F1} & \textbf{Precision} & \textbf{Recall} \\
\midrule
\endfirsthead
\multicolumn{7}{c}{\tablename\ \thetable\ -- \textit{continued}} \\
\toprule
\textbf{Model size} & \textbf{Horizon} & \textbf{AUROC} & \textbf{AUPRC} & \textbf{F1} & \textbf{Precision} & \textbf{Recall} \\
\midrule
\endhead
\bottomrule
\endlastfoot
0.69M  & 2y & 0.732 [0.724, 0.740] & 0.0274 [0.0267, 0.0285] & 0.057 [0.055, 0.059] & 0.037 [0.035, 0.038] & 0.128 [0.122, 0.133] \\
       & 5y & 0.706 [0.693, 0.720] & 0.0927 [0.0898, 0.0969] & 0.147 [0.141, 0.153] & 0.101 [0.096, 0.107] & 0.288 [0.278, 0.298] \\
2.81M  & 2y & 0.764 [0.757, 0.771] & 0.0345 [0.0334, 0.0360] & 0.072 [0.069, 0.075] & 0.048 [0.046, 0.050] & 0.158 [0.151, 0.165] \\
       & 5y & 0.748 [0.736, 0.761] & 0.1151 [0.1108, 0.1209] & 0.173 [0.167, 0.179] & 0.120 [0.116, 0.125] & 0.333 [0.321, 0.344] \\
5.81M  & 2y & 0.755 [0.747, 0.763] & 0.0345 [0.0335, 0.0361] & 0.071 [0.068, 0.074] & 0.049 [0.047, 0.051] & 0.150 [0.143, 0.156] \\
       & 5y & 0.725 [0.712, 0.740] & 0.1094 [0.1057, 0.1146] & 0.165 [0.159, 0.171] & 0.117 [0.112, 0.121] & 0.311 [0.295, 0.330] \\
14M & 2y & 0.759 [0.751, 0.768] & 0.0361 [0.0349, 0.0378] & 0.074 [0.071, 0.077] & 0.050 [0.048, 0.052] & 0.166 [0.160, 0.173] \\
       & 5y & 0.748 [0.733, 0.763] & 0.1139 [0.1099, 0.1198] & 0.175 [0.168, 0.182] & 0.118 [0.113, 0.127] & 0.354 [0.343, 0.367] \\
47M    & 2y & 0.775 [0.766, 0.784] & 0.0393 [0.0380, 0.0412] & 0.081 [0.077, 0.085] & 0.053 [0.051, 0.056] & 0.174 [0.167, 0.182] \\
       & 5y & 0.749 [0.735, 0.765] & 0.1149 [0.1106, 0.1224] & 0.169 [0.163, 0.175] & 0.114 [0.110, 0.118] & 0.358 [0.344, 0.374] \\
144M   & 2y & 0.780 [0.771, 0.788] & 0.0398 [0.0386, 0.0417] & 0.080 [0.077, 0.084] & 0.054 [0.052, 0.057] & 0.166 [0.160, 0.173] \\
       & 5y & 0.752 [0.738, 0.766] & 0.1160 [0.1123, 0.1220] & 0.173 [0.167, 0.180] & 0.122 [0.117, 0.127] & 0.337 [0.322, 0.355] \\
446M   & 2y & 0.774 [0.766, 0.782] & 0.0384 [0.0374, 0.0401] & 0.077 [0.074, 0.081] & 0.053 [0.050, 0.055] & 0.156 [0.150, 0.163] \\
       & 5y & 0.754 [0.741, 0.769] & 0.1195 [0.1152, 0.1261] & 0.176 [0.170, 0.182] & 0.120 [0.115, 0.123] & 0.364 [0.349, 0.382] \\
849M   & 2y & 0.781 [0.772, 0.789] & 0.0398 [0.0385, 0.0416] & 0.078 [0.075, 0.081] & 0.052 [0.050, 0.053] & 0.178 [0.172, 0.184] \\
       & 5y & 0.755 [0.742, 0.768] & 0.1176 [0.1136, 0.1238] & 0.174 [0.169, 0.180] & 0.119 [0.115, 0.123] & 0.345 [0.333, 0.357] \\
\end{longtable}
}

{
\scriptsize
\setlength{\tabcolsep}{4pt}
\begin{longtable}{llccccc}
\caption[Macro-averaged zero-shot forecasting across data budgets]{Macro-averaged zero-shot forecasting across data budgets ($\lambda=0.5$).}
\label{tab:zs_data_budget_macro_ci}\\
\toprule
\textbf{Data budget} & \textbf{Horizon} & \textbf{AUROC} & \textbf{AUPRC} & \textbf{F1} & \textbf{Precision} & \textbf{Recall} \\
\midrule
\endfirsthead
\multicolumn{7}{c}{\tablename\ \thetable\ -- \textit{continued}} \\
\toprule
\textbf{Data budget} & \textbf{Horizon} & \textbf{AUROC} & \textbf{AUPRC} & \textbf{F1} & \textbf{Precision} & \textbf{Recall} \\
\midrule
\endhead
\bottomrule
\endlastfoot
10\%         & 2y & 0.755 [0.746, 0.763] & 0.0339 [0.0330, 0.0356] & 0.069 [0.066, 0.072] & 0.046 [0.044, 0.049] & 0.154 [0.147, 0.162] \\
             & 5y & 0.723 [0.710, 0.738] & 0.1028 [0.0991, 0.1079] & 0.159 [0.154, 0.165] & 0.108 [0.104, 0.112] & 0.314 [0.304, 0.325] \\
25\%         & 2y & 0.771 [0.763, 0.779] & 0.0368 [0.0357, 0.0384] & 0.076 [0.072, 0.079] & 0.052 [0.051, 0.056] & 0.157 [0.149, 0.163] \\
             & 5y & 0.739 [0.725, 0.755] & 0.1064 [0.1030, 0.1117] & 0.170 [0.164, 0.176] & 0.114 [0.110, 0.119] & 0.372 [0.356, 0.389] \\
50\%         & 2y & 0.764 [0.755, 0.773] & 0.0378 [0.0367, 0.0395] & 0.077 [0.074, 0.080] & 0.051 [0.049, 0.053] & 0.183 [0.176, 0.190] \\
             & 5y & 0.740 [0.725, 0.754] & 0.1126 [0.1085, 0.1182] & 0.171 [0.165, 0.177] & 0.115 [0.112, 0.120] & 0.338 [0.327, 0.349] \\
100\%        & 2y & 0.780 [0.771, 0.788] & 0.0398 [0.0386, 0.0417] & 0.080 [0.077, 0.084] & 0.054 [0.052, 0.057] & 0.166 [0.160, 0.173] \\
             & 5y & 0.752 [0.738, 0.766] & 0.1160 [0.1123, 0.1220] & 0.173 [0.167, 0.180] & 0.122 [0.117, 0.127] & 0.337 [0.322, 0.355] \\
\end{longtable}
}

\subsubsection*{Baselines}

\begin{table}[ht]
\centering
\caption{Zero-shot forecasting at 2-year horizon (AUROC, 95\% confidence intervals).}
\label{tab:supp_zs_2y_auroc_ci}
\scriptsize
\setlength{\tabcolsep}{4pt}
\resizebox{\linewidth}{!}{%
\begin{tabular}{lcccccc}
\toprule
\textbf{Condition} & \textbf{RAVEN} & \textbf{Multiclass} & \textbf{SeqLoss} & \textbf{EGE} & \textbf{BERT (FT)} & \textbf{MedGemma-27B} \\
\midrule
Dementia          & 0.789 [0.775, 0.801] & 0.687 [0.637, 0.735] & 0.704 [0.654, 0.752] & 0.677 [0.662, 0.694] & 0.731 [0.729, 0.733] & 0.620 [0.592, 0.649] \\
Knee OA           & 0.726 [0.718, 0.733] & 0.627 [0.598, 0.657] & 0.672 [0.644, 0.700] & 0.606 [0.597, 0.614] & 0.744 [0.736, 0.750] & 0.642 [0.624, 0.661] \\
COPD              & 0.691 [0.682, 0.701] & 0.544 [0.509, 0.578] & 0.629 [0.599, 0.661] & 0.602 [0.593, 0.611] & 0.704 [0.695, 0.712] & 0.589 [0.568, 0.607] \\
CHF               & 0.857 [0.850, 0.864] & 0.708 [0.666, 0.747] & 0.733 [0.695, 0.768] & 0.656 [0.646, 0.667] & 0.862 [0.854, 0.869] & 0.736 [0.712, 0.758] \\
Acute MI          & 0.793 [0.781, 0.804] & 0.603 [0.566, 0.647] & 0.579 [0.541, 0.617] & 0.604 [0.592, 0.614] & 0.818 [0.808, 0.828] & 0.748 [0.724, 0.772] \\
Pancreatic Cancer & 0.693 [0.637, 0.745] & 0.496 [0.495, 0.496] & 0.494 [0.493, 0.495] & 0.502 [0.498, 0.509] & 0.607 [0.551, 0.659] & 0.620 [0.446, 0.775] \\
Prostate Cancer   & 0.910 [0.904, 0.916] & 0.828 [0.763, 0.887] & 0.820 [0.751, 0.880] & 0.736 [0.719, 0.754] & 0.905 [0.898, 0.911] & 0.659 [0.628, 0.690] \\
\bottomrule
\end{tabular}%
}
\end{table}

\begin{table}[ht]
\centering
\caption{Zero-shot forecasting at 2-year horizon (AUPRC, 95\% confidence intervals).}
\label{tab:supp_zs_2y_auprc_ci}
\scriptsize
\setlength{\tabcolsep}{4pt}
\resizebox{\linewidth}{!}{%
\begin{tabular}{lcccccc}
\toprule
\textbf{Condition} & \textbf{RAVEN} & \textbf{Multiclass} & \textbf{SeqLoss} & \textbf{EGE} & \textbf{BERT (FT)} & \textbf{MedGemma-27B} \\
\midrule
Dementia          & 0.037 [0.033, 0.043] & 0.026 [0.018, 0.041] & 0.031 [0.019, 0.061] & 0.026 [0.023, 0.030] & 0.050 [0.050, 0.051] & 0.011 [0.009, 0.016] \\
Knee OA           & 0.057 [0.054, 0.060] & 0.039 [0.034, 0.047] & 0.053 [0.045, 0.065] & 0.038 [0.037, 0.041] & 0.064 [0.060, 0.068] & 0.039 [0.036, 0.044] \\
COPD              & 0.052 [0.049, 0.056] & 0.036 [0.027, 0.053] & 0.044 [0.035, 0.060] & 0.040 [0.038, 0.044] & 0.055 [0.052, 0.059] & 0.030 [0.027, 0.033] \\
CHF               & 0.078 [0.071, 0.086] & 0.047 [0.034, 0.073] & 0.048 [0.035, 0.073] & 0.045 [0.039, 0.052] & 0.085 [0.077, 0.094] & 0.036 [0.031, 0.044] \\
Acute MI          & 0.031 [0.029, 0.034] & 0.015 [0.012, 0.022] & 0.013 [0.011, 0.019] & 0.017 [0.015, 0.019] & 0.049 [0.043, 0.057] & 0.026 [0.022, 0.032] \\
Pancreatic Cancer & 0.002 [0.001, 0.004] & 0.000 [0.000, 0.000] & 0.000 [0.000, 0.000] & 0.001 [0.000, 0.001] & 0.001 [0.001, 0.001] & 0.001 [0.000, 0.002] \\
Prostate Cancer   & 0.022 [0.020, 0.025] & 0.018 [0.011, 0.038] & 0.017 [0.010, 0.034] & 0.010 [0.009, 0.011] & 0.021 [0.019, 0.025] & 0.005 [0.003, 0.018] \\
\bottomrule
\end{tabular}%
}
\end{table}

\begin{table}[ht]
\centering
\caption{Zero-shot forecasting at 5-year horizon (AUROC, 95\% confidence intervals).}
\label{tab:supp_zs_5y_auroc_ci}
\scriptsize
\setlength{\tabcolsep}{4pt}
\resizebox{\linewidth}{!}{%
\begin{tabular}{lcccccc}
\toprule
\textbf{Condition} & \textbf{RAVEN} & \textbf{Multiclass} & \textbf{SeqLoss} & \textbf{EGE} & \textbf{BERT (FT)} & \textbf{MedGemma-27B} \\
\midrule
Dementia          & 0.773 [0.756, 0.791] & 0.669 [0.637, 0.700] & 0.693 [0.663, 0.725] & 0.670 [0.652, 0.690] & 0.721 [0.714, 0.726] & 0.609 [0.592, 0.626] \\
Knee OA           & 0.697 [0.686, 0.709] & 0.635 [0.618, 0.651] & 0.671 [0.654, 0.689] & 0.620 [0.608, 0.632] & 0.718 [0.706, 0.729] & 0.617 [0.607, 0.628] \\
COPD              & 0.676 [0.664, 0.689] & 0.531 [0.509, 0.550] & 0.640 [0.622, 0.659] & 0.615 [0.603, 0.628] & 0.699 [0.687, 0.711] & 0.586 [0.576, 0.596] \\
CHF               & 0.821 [0.806, 0.835] & 0.718 [0.690, 0.745] & 0.723 [0.694, 0.749] & 0.666 [0.649, 0.683] & 0.838 [0.825, 0.851] & 0.718 [0.703, 0.732] \\
Acute MI          & 0.754 [0.736, 0.769] & 0.613 [0.587, 0.640] & 0.569 [0.544, 0.595] & 0.623 [0.606, 0.640] & 0.782 [0.766, 0.797] & 0.699 [0.680, 0.717] \\
Pancreatic Cancer & 0.639 [0.548, 0.732] & 0.516 [0.491, 0.558] & 0.513 [0.488, 0.558] & 0.498 [0.498, 0.499] & 0.642 [0.581, 0.702] & 0.597 [0.523, 0.667] \\
Prostate Cancer   & 0.901 [0.893, 0.910] & 0.784 [0.749, 0.821] & 0.803 [0.761, 0.836] & 0.779 [0.756, 0.801] & 0.896 [0.888, 0.905] & 0.642 [0.621, 0.662] \\
\bottomrule
\end{tabular}%
}
\end{table}

\begin{table}[ht]
\centering
\caption{Zero-shot forecasting at 5-year horizon (AUPRC, 95\% confidence intervals).}
\label{tab:supp_zs_5y_auprc_ci}
\scriptsize
\setlength{\tabcolsep}{4pt}
\resizebox{\linewidth}{!}{%
\begin{tabular}{lcccccc}
\toprule
\textbf{Condition} & \textbf{RAVEN} & \textbf{Multiclass} & \textbf{SeqLoss} & \textbf{EGE} & \textbf{BERT (FT)} & \textbf{MedGemma-27B} \\
\midrule
Dementia          & 0.108 [0.096, 0.125] & 0.089 [0.071, 0.112] & 0.102 [0.082, 0.129] & 0.088 [0.075, 0.103] & 0.146 [0.141, 0.150] & 0.041 [0.038, 0.045] \\
Knee OA           & 0.200 [0.190, 0.213] & 0.166 [0.153, 0.181] & 0.190 [0.174, 0.208] & 0.166 [0.157, 0.178] & 0.221 [0.207, 0.235] & 0.146 [0.138, 0.154] \\
COPD              & 0.164 [0.154, 0.174] & 0.124 [0.111, 0.140] & 0.169 [0.151, 0.194] & 0.149 [0.139, 0.162] & 0.189 [0.176, 0.203] & 0.115 [0.109, 0.121] \\
CHF               & 0.184 [0.165, 0.205] & 0.114 [0.096, 0.138] & 0.134 [0.110, 0.165] & 0.127 [0.110, 0.148] & 0.220 [0.195, 0.248] & 0.101 [0.091, 0.112] \\
Acute MI          & 0.080 [0.072, 0.089] & 0.060 [0.048, 0.075] & 0.047 [0.041, 0.057] & 0.059 [0.052, 0.068] & 0.106 [0.093, 0.123] & 0.067 [0.059, 0.076] \\
Pancreatic Cancer & 0.004 [0.002, 0.021] & 0.002 [0.002, 0.005] & 0.004 [0.002, 0.019] & 0.001 [0.001, 0.002] & 0.003 [0.002, 0.005] & 0.002 [0.001, 0.003] \\
Prostate Cancer   & 0.073 [0.063, 0.087] & 0.061 [0.044, 0.095] & 0.049 [0.037, 0.069] & 0.045 [0.038, 0.056] & 0.065 [0.058, 0.076] & 0.017 [0.016, 0.020] \\
\bottomrule
\end{tabular}%
}
\end{table}

\rev{\subsection*{Stratification results with baselines}}

\rev{We obtain stratification results over age\,$\times$\,sex for RAVEN and simple supervised baselines for all prediction horizons and diseases except for pancreatic cancer. For the stratified analysis, strata with fewer than 10 positive cases are omitted, and the $\geq$80-year bands are excluded owing to sparsity; given small per-stratum event counts, we summarize results by the consistency of the direction of effect across strata.}

{%
\scriptsize
\setlength{\tabcolsep}{4pt}
\begin{longtable}{lllr@{\hspace{6pt}}r p{2.55cm} p{2.55cm} p{2.55cm}}
\caption{Age- and sex-stratified zero-shot disease-onset prediction performance (AUROC with 95\% confidence intervals). We omit the 80+ age band and missing-age strata because many cells contained few incident cases. The best AUROC within each row is bolded.}\label{tab:supp_stratified_age_sex_auroc_ci}\\
\toprule
\textbf{Condition} & \textbf{Horizon} & \textbf{Stratum} & \textbf{$n$} & \textbf{Events} & \textbf{RAVEN} & \textbf{LR} & \textbf{GBM} \\
\midrule
\endfirsthead
\toprule
\textbf{Condition} & \textbf{Horizon} & \textbf{Stratum} & \textbf{$n$} & \textbf{Events} & \textbf{RAVEN} & \textbf{LR} & \textbf{GBM} \\
\midrule
\endhead
\midrule
\multicolumn{8}{r}{\emph{Continued on next page}} \\
\endfoot
\bottomrule
\endlastfoot
CHF & 2y & Female, <40 & 59,084 & 76 & \textbf{0.823 [0.754, 0.894]} & 0.643 [0.554, 0.734] & 0.669 [0.588, 0.743] \\
CHF & 2y & Female, 40-60 & 66,316 & 418 & \textbf{0.788 [0.760, 0.816]} & 0.626 [0.593, 0.659] & 0.666 [0.635, 0.697] \\
CHF & 2y & Female, 60-70 & 25,945 & 444 & \textbf{0.771 [0.744, 0.798]} & 0.636 [0.603, 0.668] & 0.641 [0.608, 0.669] \\
CHF & 2y & Female, 70-80 & 10,516 & 403 & \textbf{0.706 [0.674, 0.737]} & 0.590 [0.553, 0.626] & 0.590 [0.557, 0.622] \\
CHF & 2y & Male, <40 & 32,197 & 71 & \textbf{0.855 [0.802, 0.900]} & 0.743 [0.678, 0.803] & 0.728 [0.670, 0.788] \\
CHF & 2y & Male, 40-60 & 41,154 & 518 & \textbf{0.771 [0.745, 0.797]} & 0.655 [0.627, 0.684] & 0.651 [0.624, 0.677] \\
CHF & 2y & Male, 60-70 & 16,851 & 395 & \textbf{0.721 [0.691, 0.749]} & 0.606 [0.571, 0.643] & 0.620 [0.587, 0.652] \\
CHF & 2y & Male, 70-80 & 6,191 & 358 & \textbf{0.689 [0.655, 0.724]} & 0.616 [0.578, 0.653] & 0.611 [0.575, 0.648] \\
\addlinespace[2pt]
CHF & 5y & Female, <40 & 10,026 & 58 & \textbf{0.718 [0.589, 0.852]} & 0.593 [0.429, 0.745] & 0.711 [0.606, 0.798] \\
CHF & 5y & Female, 40-60 & 14,585 & 366 & \textbf{0.749 [0.708, 0.792]} & 0.639 [0.592, 0.681] & 0.646 [0.598, 0.688] \\
CHF & 5y & Female, 60-70 & 5,546 & 362 & \textbf{0.726 [0.685, 0.765]} & 0.603 [0.556, 0.644] & 0.587 [0.543, 0.630] \\
CHF & 5y & Female, 70-80 & 1,993 & 247 & \textbf{0.650 [0.594, 0.706]} & 0.549 [0.498, 0.605] & 0.557 [0.510, 0.613] \\
CHF & 5y & Male, <40 & 5,581 & 55 & \textbf{0.847 [0.764, 0.915]} & 0.723 [0.638, 0.794] & 0.667 [0.586, 0.735] \\
CHF & 5y & Male, 40-60 & 9,120 & 403 & \textbf{0.728 [0.688, 0.768]} & 0.610 [0.570, 0.652] & 0.618 [0.578, 0.655] \\
CHF & 5y & Male, 60-70 & 3,534 & 356 & \textbf{0.662 [0.613, 0.710]} & 0.575 [0.526, 0.621] & 0.580 [0.537, 0.618] \\
CHF & 5y & Male, 70-80 & 1,168 & 238 & \textbf{0.708 [0.652, 0.763]} & 0.594 [0.537, 0.653] & 0.578 [0.527, 0.625] \\
\midrule
COPD & 2y & Female, <40 & 55,257 & 561 & \textbf{0.632 [0.602, 0.661]} & 0.563 [0.531, 0.593] & 0.599 [0.574, 0.625] \\
COPD & 2y & Female, 40-60 & 58,154 & 1,284 & \textbf{0.651 [0.630, 0.669]} & 0.594 [0.576, 0.612] & 0.587 [0.569, 0.605] \\
COPD & 2y & Female, 60-70 & 21,994 & 734 & \textbf{0.641 [0.616, 0.665]} & 0.608 [0.583, 0.633] & 0.611 [0.585, 0.636] \\
COPD & 2y & Female, 70-80 & 9,797 & 396 & \textbf{0.625 [0.590, 0.661]} & 0.584 [0.546, 0.617] & 0.594 [0.557, 0.626] \\
COPD & 2y & Male, <40 & 30,356 & 351 & \textbf{0.622 [0.588, 0.658]} & 0.553 [0.514, 0.592] & 0.610 [0.578, 0.643] \\
COPD & 2y & Male, 40-60 & 36,865 & 992 & \textbf{0.647 [0.627, 0.669]} & 0.596 [0.573, 0.616] & 0.615 [0.595, 0.635] \\
COPD & 2y & Male, 60-70 & 15,423 & 516 & \textbf{0.643 [0.612, 0.672]} & 0.602 [0.572, 0.633] & 0.604 [0.575, 0.635] \\
COPD & 2y & Male, 70-80 & 6,388 & 287 & \textbf{0.617 [0.573, 0.660]} & 0.600 [0.559, 0.640] & 0.592 [0.550, 0.633] \\
\addlinespace[2pt]
COPD & 5y & Female, <40 & 9,419 & 391 & \textbf{0.632 [0.588, 0.675]} & 0.556 [0.509, 0.600] & 0.568 [0.530, 0.608] \\
COPD & 5y & Female, 40-60 & 13,000 & 1,200 & \textbf{0.640 [0.616, 0.664]} & 0.590 [0.563, 0.617] & 0.594 [0.571, 0.616] \\
COPD & 5y & Female, 60-70 & 4,745 & 615 & 0.592 [0.561, 0.627] & \textbf{0.597 [0.566, 0.631]} & 0.595 [0.564, 0.627] \\
COPD & 5y & Female, 70-80 & 1,872 & 308 & 0.555 [0.502, 0.602] & 0.566 [0.517, 0.613] & \textbf{0.581 [0.531, 0.623]} \\
COPD & 5y & Male, <40 & 5,270 & 238 & \textbf{0.655 [0.597, 0.710]} & 0.570 [0.506, 0.631] & 0.615 [0.563, 0.662] \\
COPD & 5y & Male, 40-60 & 8,349 & 868 & \textbf{0.623 [0.598, 0.650]} & 0.595 [0.569, 0.626] & 0.589 [0.564, 0.616] \\
COPD & 5y & Male, 60-70 & 3,256 & 373 & \textbf{0.614 [0.568, 0.657]} & 0.582 [0.537, 0.622] & 0.575 [0.538, 0.611] \\
COPD & 5y & Male, 70-80 & 1,205 & 195 & 0.567 [0.502, 0.632] & \textbf{0.605 [0.534, 0.669]} & 0.504 [0.447, 0.555] \\
\midrule
Knee OA & 2y & Female, <40 & 56,521 & 476 & \textbf{0.796 [0.774, 0.817]} & 0.783 [0.762, 0.803] & 0.771 [0.749, 0.792] \\
Knee OA & 2y & Female, 40-60 & 53,840 & 2,040 & \textbf{0.629 [0.614, 0.644]} & 0.618 [0.603, 0.632] & 0.617 [0.602, 0.631] \\
Knee OA & 2y & Female, 60-70 & 21,135 & 981 & \textbf{0.619 [0.596, 0.642]} & 0.585 [0.562, 0.609] & 0.594 [0.572, 0.616] \\
Knee OA & 2y & Female, 70-80 & 9,968 & 406 & \textbf{0.649 [0.617, 0.681]} & 0.602 [0.567, 0.640] & 0.627 [0.595, 0.661] \\
Knee OA & 2y & Male, <40 & 31,663 & 180 & \textbf{0.804 [0.767, 0.839]} & 0.772 [0.739, 0.805] & 0.769 [0.736, 0.802] \\
Knee OA & 2y & Male, 40-60 & 37,924 & 985 & \textbf{0.629 [0.608, 0.650]} & 0.570 [0.547, 0.594] & 0.587 [0.566, 0.606] \\
Knee OA & 2y & Male, 60-70 & 16,842 & 419 & \textbf{0.610 [0.579, 0.642]} & 0.584 [0.548, 0.618] & 0.605 [0.573, 0.636] \\
Knee OA & 2y & Male, 70-80 & 7,184 & 215 & \textbf{0.638 [0.591, 0.686]} & 0.579 [0.533, 0.625] & 0.628 [0.585, 0.674] \\
\addlinespace[2pt]
Knee OA & 5y & Female, <40 & 9,610 & 454 & \textbf{0.765 [0.732, 0.801]} & 0.759 [0.729, 0.789] & 0.743 [0.711, 0.774] \\
Knee OA & 5y & Female, 40-60 & 12,362 & 1,791 & \textbf{0.616 [0.594, 0.638]} & 0.605 [0.584, 0.625] & 0.607 [0.587, 0.627] \\
Knee OA & 5y & Female, 60-70 & 4,764 & 884 & 0.589 [0.557, 0.622] & 0.599 [0.571, 0.630] & \textbf{0.603 [0.575, 0.630]} \\
Knee OA & 5y & Female, 70-80 & 1,956 & 282 & \textbf{0.661 [0.611, 0.710]} & 0.586 [0.534, 0.633] & 0.611 [0.556, 0.658] \\
Knee OA & 5y & Male, <40 & 5,527 & 173 & \textbf{0.735 [0.667, 0.798]} & 0.730 [0.678, 0.778] & 0.732 [0.678, 0.779] \\
Knee OA & 5y & Male, 40-60 & 8,603 & 895 & \textbf{0.622 [0.592, 0.653]} & 0.574 [0.542, 0.605] & 0.595 [0.565, 0.623] \\
Knee OA & 5y & Male, 60-70 & 3,549 & 389 & 0.571 [0.522, 0.619] & 0.586 [0.537, 0.634] & \textbf{0.622 [0.582, 0.660]} \\
Knee OA & 5y & Male, 70-80 & 1,302 & 157 & \textbf{0.684 [0.605, 0.753]} & 0.523 [0.458, 0.597] & 0.539 [0.473, 0.604] \\
\midrule
Dementia & 2y & Female, <40 & 59,047 & 82 & 0.629 [0.574, 0.690] & 0.615 [0.538, 0.680] & \textbf{0.686 [0.611, 0.754]} \\
Dementia & 2y & Female, 40-60 & 67,244 & 280 & \textbf{0.690 [0.656, 0.727]} & 0.650 [0.609, 0.692] & 0.670 [0.632, 0.707] \\
Dementia & 2y & Female, 60-70 & 26,666 & 382 & \textbf{0.651 [0.616, 0.684]} & 0.594 [0.561, 0.628] & 0.588 [0.554, 0.620] \\
Dementia & 2y & Female, 70-80 & 11,139 & 382 & \textbf{0.593 [0.556, 0.627]} & 0.586 [0.549, 0.622] & 0.549 [0.518, 0.584] \\
Dementia & 2y & Male, <40 & 32,382 & 48 & 0.633 [0.564, 0.709] & \textbf{0.729 [0.645, 0.798]} & 0.665 [0.580, 0.750] \\
Dementia & 2y & Male, 40-60 & 43,632 & 220 & \textbf{0.667 [0.627, 0.709]} & 0.624 [0.574, 0.671] & 0.622 [0.572, 0.665] \\
Dementia & 2y & Male, 60-70 & 18,398 & 263 & \textbf{0.672 [0.634, 0.711]} & 0.619 [0.581, 0.662] & 0.633 [0.597, 0.672] \\
Dementia & 2y & Male, 70-80 & 7,151 & 231 & \textbf{0.604 [0.560, 0.648]} & 0.569 [0.520, 0.618] & 0.571 [0.532, 0.615] \\
\addlinespace[2pt]
Dementia & 5y & Female, <40 & 10,036 & 79 & \textbf{0.686 [0.601, 0.768]} & 0.549 [0.444, 0.652] & 0.679 [0.589, 0.752] \\
Dementia & 5y & Female, 40-60 & 14,775 & 331 & \textbf{0.717 [0.673, 0.760]} & 0.669 [0.623, 0.711] & 0.671 [0.629, 0.710] \\
Dementia & 5y & Female, 60-70 & 5,715 & 368 & \textbf{0.639 [0.598, 0.679]} & 0.628 [0.584, 0.672] & 0.579 [0.530, 0.627] \\
Dementia & 5y & Female, 70-80 & 2,187 & 245 & \textbf{0.587 [0.541, 0.633]} & 0.524 [0.466, 0.579] & 0.540 [0.480, 0.598] \\
Dementia & 5y & Male, <40 & 5,602 & 47 & 0.557 [0.478, 0.657] & \textbf{0.674 [0.528, 0.795]} & 0.636 [0.491, 0.763] \\
Dementia & 5y & Male, 40-60 & 9,539 & 226 & \textbf{0.681 [0.624, 0.735]} & 0.626 [0.573, 0.684] & 0.618 [0.566, 0.675] \\
Dementia & 5y & Male, 60-70 & 3,824 & 217 & \textbf{0.655 [0.599, 0.706]} & 0.615 [0.555, 0.675] & 0.621 [0.567, 0.670] \\
Dementia & 5y & Male, 70-80 & 1,300 & 141 & \textbf{0.579 [0.505, 0.648]} & 0.552 [0.479, 0.633] & 0.575 [0.514, 0.634] \\
\midrule
Acute MI & 2y & Female, <40 & 59,038 & 79 & 0.748 [0.668, 0.818] & 0.747 [0.689, 0.804] & \textbf{0.762 [0.700, 0.823]} \\
Acute MI & 2y & Female, 40-60 & 66,579 & 361 & \textbf{0.705 [0.668, 0.738]} & 0.677 [0.643, 0.712] & 0.683 [0.648, 0.719] \\
Acute MI & 2y & Female, 60-70 & 27,244 & 346 & \textbf{0.694 [0.661, 0.729]} & 0.622 [0.582, 0.659] & 0.636 [0.599, 0.672] \\
Acute MI & 2y & Female, 70-80 & 11,887 & 268 & \textbf{0.662 [0.621, 0.700]} & 0.576 [0.535, 0.615] & 0.605 [0.566, 0.645] \\
Acute MI & 2y & Male, <40 & 32,228 & 73 & \textbf{0.777 [0.716, 0.839]} & 0.768 [0.712, 0.823] & 0.777 [0.721, 0.828] \\
Acute MI & 2y & Male, 40-60 & 41,529 & 449 & \textbf{0.711 [0.680, 0.742]} & 0.654 [0.623, 0.685] & 0.662 [0.630, 0.692] \\
Acute MI & 2y & Male, 60-70 & 17,788 & 321 & \textbf{0.666 [0.631, 0.704]} & 0.610 [0.568, 0.646] & 0.632 [0.593, 0.670] \\
Acute MI & 2y & Male, 70-80 & 7,299 & 251 & 0.591 [0.549, 0.631] & \textbf{0.602 [0.553, 0.648]} & 0.595 [0.550, 0.643] \\
\addlinespace[2pt]
Acute MI & 5y & Female, <40 & 9,999 & 61 & 0.680 [0.584, 0.780] & \textbf{0.726 [0.624, 0.809]} & 0.690 [0.625, 0.754] \\
Acute MI & 5y & Female, 40-60 & 14,594 & 308 & \textbf{0.656 [0.611, 0.695]} & 0.642 [0.600, 0.682] & 0.621 [0.580, 0.660] \\
Acute MI & 5y & Female, 60-70 & 5,825 & 249 & \textbf{0.616 [0.566, 0.671]} & 0.552 [0.495, 0.605] & 0.540 [0.488, 0.591] \\
Acute MI & 5y & Female, 70-80 & 2,196 & 192 & 0.585 [0.532, 0.638] & \textbf{0.590 [0.521, 0.649]} & 0.556 [0.491, 0.624] \\
Acute MI & 5y & Male, <40 & 5,574 & 47 & \textbf{0.809 [0.738, 0.874]} & 0.752 [0.654, 0.831] & 0.722 [0.632, 0.799] \\
Acute MI & 5y & Male, 40-60 & 9,072 & 314 & \textbf{0.676 [0.630, 0.717]} & 0.592 [0.545, 0.637] & 0.588 [0.546, 0.632] \\
Acute MI & 5y & Male, 60-70 & 3,668 & 284 & \textbf{0.617 [0.565, 0.666]} & 0.552 [0.504, 0.610] & 0.562 [0.518, 0.609] \\
Acute MI & 5y & Male, 70-80 & 1,316 & 159 & 0.525 [0.453, 0.590] & \textbf{0.597 [0.525, 0.669]} & 0.596 [0.533, 0.665] \\
\midrule
Prostate cancer & 2y & Male, <40 & 32,726 & 13 & \textbf{0.893 [0.792, 0.973]} & 0.862 [0.759, 0.945] & 0.762 [0.651, 0.862] \\
Prostate cancer & 2y & Male, 40-60 & 41,015 & 416 & \textbf{0.651 [0.622, 0.681]} & 0.614 [0.586, 0.642] & 0.612 [0.582, 0.640] \\
Prostate cancer & 2y & Male, 60-70 & 16,758 & 293 & \textbf{0.592 [0.554, 0.631]} & 0.542 [0.501, 0.583] & 0.505 [0.468, 0.541] \\
Prostate cancer & 2y & Male, 70-80 & 6,980 & 147 & 0.542 [0.484, 0.601] & \textbf{0.570 [0.509, 0.626]} & 0.541 [0.490, 0.595] \\
\addlinespace[2pt]
Prostate cancer & 5y & Male, <40 & 5,669 & 12 & 0.849 [0.794, 0.923] & \textbf{0.890 [0.837, 0.962]} & 0.832 [0.762, 0.928] \\
Prostate cancer & 5y & Male, 40-60 & 8,895 & 414 & \textbf{0.636 [0.605, 0.673]} & 0.608 [0.572, 0.642] & 0.614 [0.584, 0.644] \\
Prostate cancer & 5y & Male, 60-70 & 3,474 & 192 & \textbf{0.575 [0.515, 0.631]} & 0.575 [0.508, 0.638] & 0.557 [0.506, 0.608] \\
Prostate cancer & 5y & Male, 70-80 & 1,232 & 84 & \textbf{0.572 [0.472, 0.665]} & 0.528 [0.456, 0.601] & 0.500 [0.416, 0.581] \\
\end{longtable}
}

\newpage

\subsection*{\rev{Calibration and additional evaluations}}

\begin{figure}[!ht]
\centering
\includegraphics[width=\linewidth]{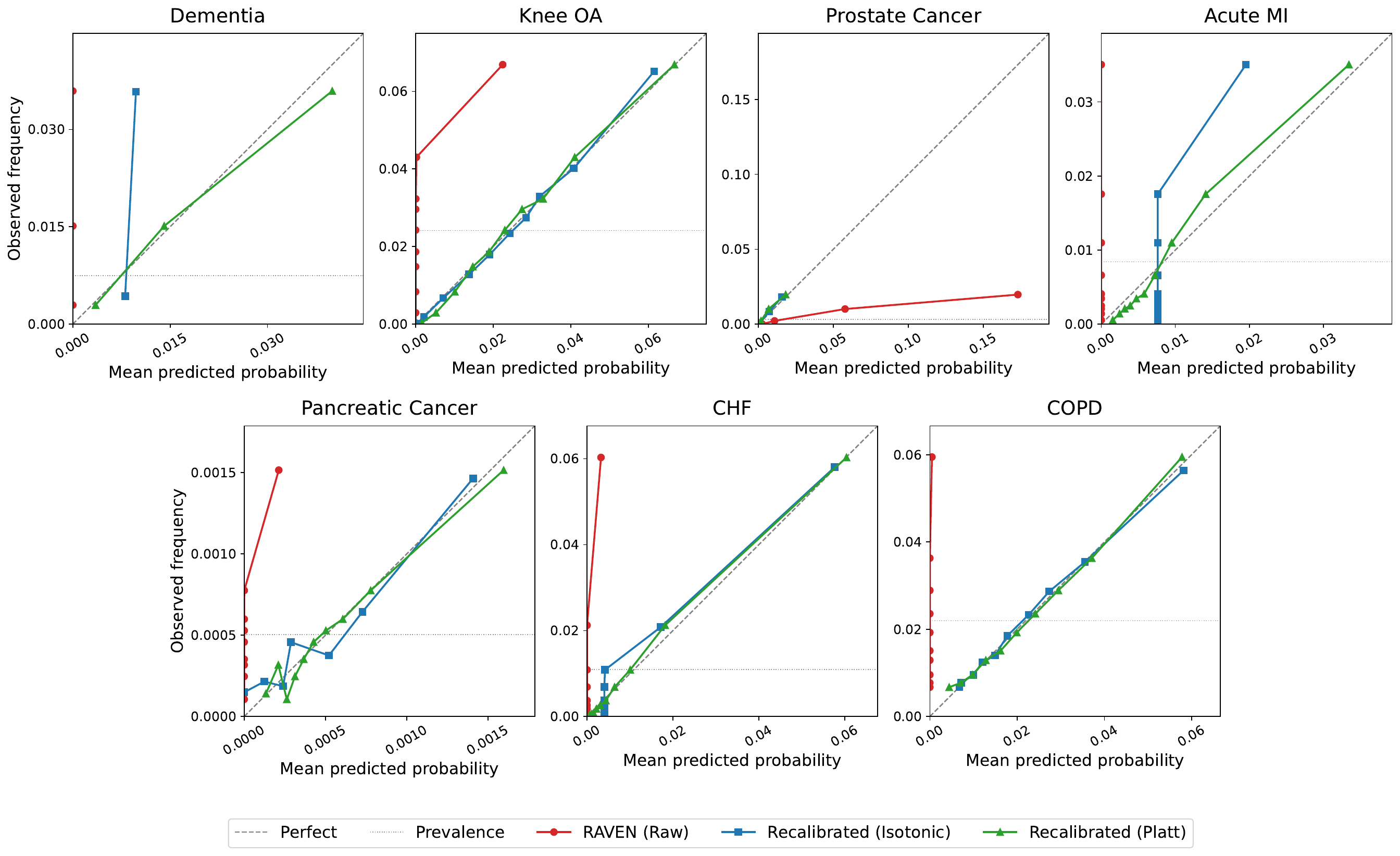}
\caption[RAVEN reliability across diseases at the 2-year horizon]{\textbf{Reliability of zero-shot RAVEN risk scores at the 2-year horizon (144M, $\lambda^\star=0.5$).} For each condition, observed onset frequency is plotted against mean predicted probability for raw scores (red), isotonic regression (blue), and Platt scaling (green); the dashed diagonal indicates perfect calibration, and the dotted line shows the task prevalence. Recalibration maps are fit on the validation split and applied to the test set. Raw scores deviate from the diagonal, whereas Platt-scaled scores track the diagonal closely. Isotonic regression attains similar calibration but collapses tied scores, reducing AUROC for dementia.}
\label{fig:supp_calib_2y}
\end{figure}

\begin{figure}[!t]
\centering
\includegraphics[width=\linewidth]{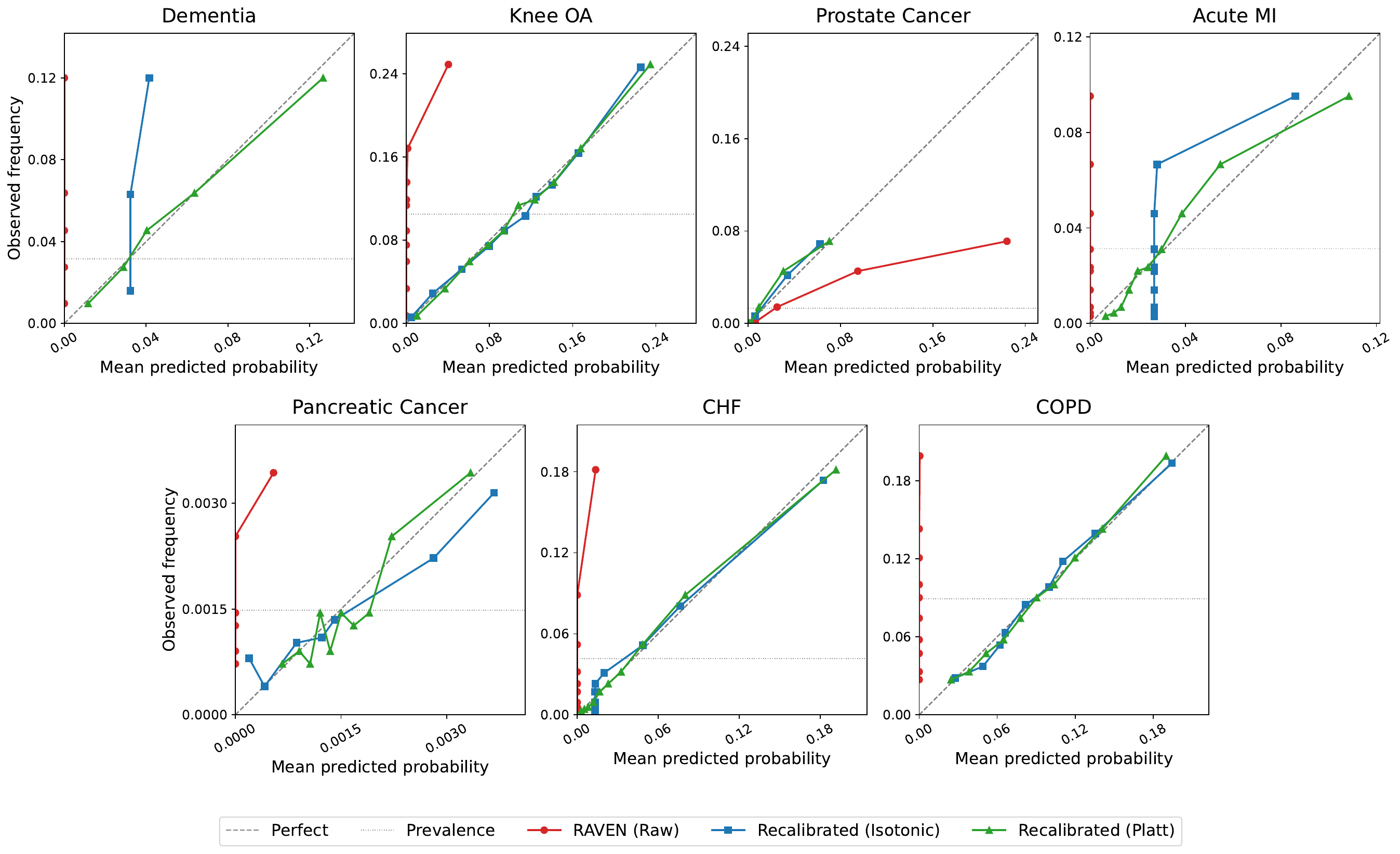}
\caption[RAVEN reliability across diseases at the 5-year horizon]{\textbf{Reliability of zero-shot RAVEN risk scores at the 5-year horizon (144M, $\lambda^\star=0.5$).} Axes, colors, and recalibration protocol are as in Supplementary Fig.~\ref{fig:supp_calib_2y}. As at the 2-year horizon, raw scores are miscalibrated, and Platt scaling restores agreement with the diagonal without changing AUROC, while isotonic regression reduces discrimination on dementia.}
\label{fig:supp_calib_5y}
\end{figure}

\rev{To characterize the probabilistic quality of RAVEN's zero-shot scores, we computed reliability diagrams and calibration summaries for the 144M, $\lambda^\star=0.5$ model at both horizons. For each condition, we binned \nrev{prediction windows} by predicted risk using quantile bins and plotted observed onset frequency against mean predicted probability (Figs.~\ref{fig:supp_calib_2y}--\ref{fig:supp_calib_5y}). We summarized calibration with the Brier score, the Brier skill score (BSS) relative to a constant-prevalence reference, and the expected calibration error (ECE). We report BSS as an additional metric because the raw Brier score is dominated by true negatives at these prevalences and is uninformative in isolation. We additionally evaluated two standard post-hoc recalibration maps, isotonic regression and Platt scaling, each fit on the held-out validation split and applied to the test set, so that no test labels inform the recalibration. All intervals are 95\% patient-level clustered bootstrap intervals. Table~\ref{tab:supp_calibration} reports every calibration metric before and after recalibration.}
 
\begin{figure}[!ht]
\centering
\includegraphics[width=\linewidth]{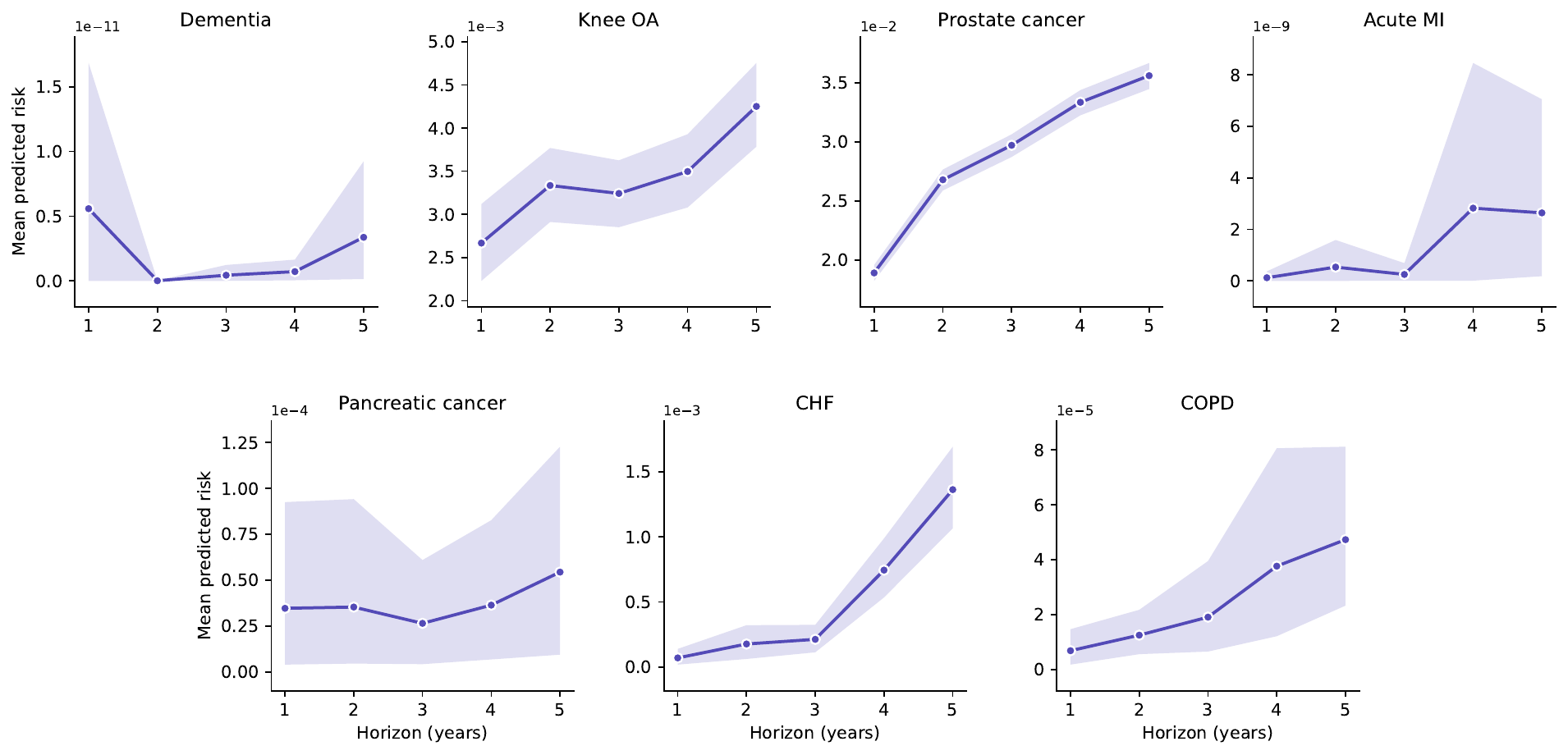}
\caption{\textbf{Mean horizon-conditioned RAVEN scores across prediction horizons (144M, $\lambda^\star=0.5$).} Mean raw zero-shot scores from RAVEN are shown for each timestep queried at $\{1,2,3,4,5\}$-year horizons for the 5-year evaluation cohort. Points denote the mean score across prediction windows and shaded regions denote 95\% patient-level clustered bootstrap intervals. Mean scores generally increase with horizon for several endpoints, but this population-level trend does not imply individual-level monotonicity or calibrated cumulative-incidence estimation.}
\label{fig:supp_mean_risk_by_horizon}
\end{figure}

\rev{Because the future-separator query is evaluated independently at each horizon, the raw zero-shot scores from RAVEN are not constrained to form a monotone cumulative-incidence function. We therefore additionally assessed horizon-wise monotonicity by querying the same main 144M, $\lambda^\star=0.5$ model at $\{1,2,3,4,5\}$-year horizons. Mean predicted scores generally increased with longer horizons for several endpoints, including knee osteoarthritis, prostate cancer, CHF, and COPD (Fig. \ref{fig:supp_mean_risk_by_horizon}). However, individual prediction-window trajectories were not uniformly monotone: across endpoints, the fraction of trajectories that were monotonically non-decreasing ranged from 0.199 to 0.471, and pairwise horizon-violation rates ranged from 0.159 to 0.332 (Table \ref{tab:supp_monotonicity}). A pairwise violation occurs when the score at a longer horizon is lower than the score at a shorter horizon. These results support interpreting the current zero-shot output as a horizon-conditioned ranking score rather than a calibrated cumulative-incidence estimate.}

\rev{Also, we report operating-point and ranking metrics for the same model in Table~\ref{tab:supp_utility}. These metrics are computed on the raw scores and, being rank-based, are invariant to the monotonic recalibration above. We report positive predictive value (PPV) at thresholds achieving 50\% sensitivity, the sensitivity (fraction of incident cases captured) and PPV obtained when the top 1\% and top 5\% of \nrev{prediction windows} by predicted risk are flagged for review, and the precision among the 100 highest-ranked \nrev{windows}. Because every precision-type metric has a no-skill value equal to the task prevalence, we report prevalence alongside each row so that absolute values can be read as enrichment over chance.}
 
\FloatBarrier

{\scriptsize
\setlength{\tabcolsep}{2.2pt}
\renewcommand{\arraystretch}{1.25}
\begin{longtable}{llcccccc}
\caption[Calibration of RAVEN before and after Platt recalibration]{\textbf{Calibration of RAVEN before and after post-hoc Platt recalibration (144M, $\lambda^\star=0.5$).} For each condition and horizon, we report the Brier score, Brier skill score (BSS), and expected calibration error (ECE) for the raw RAVEN scores and after Platt recalibration. Recalibration maps are fit on the validation split and evaluated on the held-out test set. BSS $=1-\mathrm{Brier}/\mathrm{Brier}_{\mathrm{prev}}$, where the reference predicts the task prevalence for every window. Brackets give 95\% patient-level confidence intervals. Because ECE is non-negative and positively biased, for near-perfectly calibrated, low-event cells the percentile-bootstrap interval can sit slightly above the point estimate like pancreatic cancer.}
\label{tab:supp_calibration}\\
\toprule
\textbf{Condition} & \textbf{Hz.}
& \textbf{Brier raw}
& \textbf{Brier Platt}
& \textbf{BSS raw}
& \textbf{BSS Platt}
& \textbf{ECE raw}
& \textbf{ECE Platt} \\
\midrule
\endfirsthead

\multicolumn{8}{l}{\textit{(Continued from previous page)}} \\

\toprule
\textbf{Condition} & \textbf{Hz.}
& \textbf{Brier raw}
& \textbf{Brier Platt}
& \textbf{BSS raw}
& \textbf{BSS Platt}
& \textbf{ECE raw}
& \textbf{ECE Platt} \\
\midrule
\endhead

\midrule
\multicolumn{8}{r}{\textit{(Continued on next page)}} \\
\endfoot

\bottomrule
\endlastfoot
Dementia & 2y & \shortstack{0.0075\\{[0.0073, 0.0076]}} & \shortstack{0.0073\\{[0.0072, 0.0074]}} & \shortstack{-0.008\\{[-0.008, -0.007]}} & \shortstack{0.014\\{[0.010, 0.018]}} & \shortstack{0.0075\\{[0.0073, 0.0076]}} & \shortstack{0.00096\\{[0.00070, 0.0013]}} \\
 & 5y & \shortstack{0.0315\\{[0.0302, 0.0328]}} & \shortstack{0.0292\\{[0.0281, 0.0304]}} & \shortstack{-0.033\\{[-0.034, -0.031]}} & \shortstack{0.042\\{[0.033, 0.052]}} & \shortstack{0.0315\\{[0.0302, 0.0328]}} & \shortstack{0.0025\\{[0.0015, 0.0046]}} \\
\addlinespace
Knee OA & 2y & \shortstack{0.0244\\{[0.0241, 0.0247]}} & \shortstack{0.0232\\{[0.0229, 0.0234]}} & \shortstack{-0.038\\{[-0.044, -0.033]}} & \shortstack{0.016\\{[0.014, 0.017]}} & \shortstack{0.0218\\{[0.0215, 0.0222]}} & \shortstack{0.0012\\{[0.0010, 0.0020]}} \\
 & 5y & \shortstack{0.1038\\{[0.1014, 0.1064]}} & \shortstack{0.0895\\{[0.0877, 0.0915]}} & \shortstack{-0.105\\{[-0.110, -0.099]}} & \shortstack{0.047\\{[0.041, 0.053]}} & \shortstack{0.1007\\{[0.0984, 0.1034]}} & \shortstack{0.0047\\{[0.0033, 0.0087]}} \\
\addlinespace
COPD & 2y & \shortstack{0.0219\\{[0.0217, 0.0222]}} & \shortstack{0.0212\\{[0.0209, 0.0214]}} & \shortstack{-0.022\\{[-0.023, -0.021]}} & \shortstack{0.013\\{[0.011, 0.014]}} & \shortstack{0.0219\\{[0.0216, 0.0222]}} & \shortstack{0.00087\\{[0.00068, 0.0017]}} \\
 & 5y & \shortstack{0.0893\\{[0.0870, 0.0914]}} & \shortstack{0.0786\\{[0.0769, 0.0803]}} & \shortstack{-0.098\\{[-0.101, -0.095]}} & \shortstack{0.033\\{[0.028, 0.038]}} & \shortstack{0.0892\\{[0.0870, 0.0914]}} & \shortstack{0.0039\\{[0.0030, 0.0081]}} \\
\addlinespace
CHF & 2y & \shortstack{0.0110\\{[0.0108, 0.0112]}} & \shortstack{0.0104\\{[0.0102, 0.0106]}} & \shortstack{-0.016\\{[-0.024, -0.008]}} & \shortstack{0.037\\{[0.032, 0.042]}} & \shortstack{0.0106\\{[0.0104, 0.0108]}} & \shortstack{0.00073\\{[0.00059, 0.0012]}} \\
 & 5y & \shortstack{0.0416\\{[0.0401, 0.0431]}} & \shortstack{0.0367\\{[0.0353, 0.0380]}} & \shortstack{-0.042\\{[-0.049, -0.034]}} & \shortstack{0.081\\{[0.067, 0.095]}} & \shortstack{0.0403\\{[0.0387, 0.0419]}} & \shortstack{0.0030\\{[0.0020, 0.0058]}} \\
\addlinespace
Acute MI & 2y & \shortstack{0.0084\\{[0.0083, 0.0086]}} & \shortstack{0.0083\\{[0.0081, 0.0084]}} & \shortstack{-0.008\\{[-0.009, -0.008]}} & \shortstack{0.011\\{[0.009, 0.013]}} & \shortstack{0.0084\\{[0.0083, 0.0086]}} & \shortstack{0.0015\\{[0.0011, 0.0018]}} \\
 & 5y & \shortstack{0.0313\\{[0.0300, 0.0326]}} & \shortstack{0.0296\\{[0.0285, 0.0308]}} & \shortstack{-0.032\\{[-0.034, -0.031]}} & \shortstack{0.022\\{[0.015, 0.028]}} & \shortstack{0.0313\\{[0.0300, 0.0326]}} & \shortstack{0.0054\\{[0.0040, 0.0076]}} \\
\addlinespace
Pancreatic Cancer & 2y & \shortstack{0.00051\\{[0.00048, 0.00054]}} & \shortstack{0.00050\\{[0.00047, 0.00054]}} & \shortstack{-0.012\\{[-0.028, -0.002]}} & \shortstack{0.000\\{[-0.000, 0.001]}} & \shortstack{0.00048\\{[0.00044, 0.00052]}} & \shortstack{0.00005\\{[0.00007, 0.00018]}} \\
 & 5y & \shortstack{0.0015\\{[0.0012, 0.0018]}} & \shortstack{0.0015\\{[0.0012, 0.0017]}} & \shortstack{-0.013\\{[-0.034, -0.001]}} & \shortstack{0.001\\{[-0.000, 0.002]}} & \shortstack{0.0014\\{[0.0012, 0.0017]}} & \shortstack{0.00024\\{[0.00030, 0.00081]}} \\
\addlinespace
Prostate Cancer & 2y & \shortstack{0.0063\\{[0.0062, 0.0064]}} & \shortstack{0.0032\\{[0.0031, 0.0033]}} & \shortstack{-0.962\\{[-1.014, -0.915]}} & \shortstack{0.012\\{[0.010, 0.014]}} & \shortstack{0.0215\\{[0.0211, 0.0220]}} & \shortstack{0.00066\\{[0.00055, 0.00077]}} \\
 & 5y & \shortstack{0.0155\\{[0.0148, 0.0162]}} & \shortstack{0.0125\\{[0.0117, 0.0132]}} & \shortstack{-0.194\\{[-0.234, -0.158]}} & \shortstack{0.040\\{[0.032, 0.047]}} & \shortstack{0.0224\\{[0.0212, 0.0237]}} & \shortstack{0.0025\\{[0.0020, 0.0037]}} \\
\addlinespace
\end{longtable}
\renewcommand{\arraystretch}{1.0}
}
 \newpage

{\scriptsize
\setlength{\tabcolsep}{5pt}
\begin{longtable}{lcc}
\caption{\textbf{Window-level monotonicity of horizon-conditioned zero-shot RAVEN scores.}
For each endpoint, we queried the 144M RAVEN model with recurrence-aware regularization $\lambda^\star=0.5$ at 1-, 2-, 3-, 4-, and 5-year horizons for the 5-year evaluation cohort. We report the fraction of prediction-window trajectories that were monotone non-decreasing across horizons and the pairwise horizon-violation rate. Values in brackets are 95\% patient-level clustered bootstrap intervals.}
\label{tab:supp_monotonicity}\\
\toprule
\textbf{Condition} &
\textbf{Monotone rate} &
\textbf{Pairwise violation rate} \\
\midrule
\endfirsthead

\multicolumn{3}{c}{{\tablename\ \thetable{} -- continued from previous page}}\\
\toprule
\textbf{Condition} &
\textbf{Monotone rate} &
\textbf{Pairwise violation rate} \\
\midrule
\endhead

\midrule
\multicolumn{3}{r}{{Continued on next page}}\\
\endfoot

\bottomrule
\endlastfoot

Dementia & 0.463 [0.456, 0.470] & 0.169 [0.164, 0.172] \\
Knee OA & 0.393 [0.388, 0.398] & 0.197 [0.195, 0.199] \\
Prostate cancer & 0.199 [0.194, 0.203] & 0.332 [0.329, 0.334] \\
Acute MI & 0.353 [0.348, 0.358] & 0.207 [0.206, 0.209] \\
Pancreatic cancer & 0.295 [0.290, 0.299] & 0.246 [0.243, 0.248] \\
CHF & 0.471 [0.466, 0.476] & 0.159 [0.157, 0.160] \\
COPD & 0.324 [0.319, 0.329] & 0.224 [0.222, 0.226] \\

\end{longtable}
}

{\scriptsize
\setlength{\tabcolsep}{2.2pt}
\renewcommand{\arraystretch}{1.25}
\begin{longtable}{llccccccc}
\caption[Operating-point and ranking utility of RAVEN risk scores]{\textbf{Operating-point and ranking utility of zero-shot RAVEN risk scores (144M, $\lambda^\star=0.5$).} Metrics are computed on raw scores, which are invariant to monotone recalibration. PPV at fixed sensitivity reports precision at the threshold achieving 50\% recall. Sensitivity and PPV at fixed alert burden report the fraction of incident cases captured and the precision when the top 1\% or top 5\% of prediction windows by predicted risk are flagged for review. Precision @ top-100 reports the fraction of true onsets among the 100 highest-ranked windows; for context, the no-skill value of every precision-type metric equals the task prevalence. Brackets give 95\% patient-level clustered bootstrap confidence intervals.}
\label{tab:supp_utility}\\
\toprule
\textbf{Condition} & \textbf{Hz.} & \textbf{Prev.\,(\%)}
& \shortstack{\textbf{PPV}\\\textbf{@ Sn=0.5}}
& \shortstack{\textbf{Sens}\\\textbf{@ top-1\%}}
& \shortstack{\textbf{PPV}\\\textbf{@ top-1\%}}
& \shortstack{\textbf{Sens}\\\textbf{@ top-5\%}}
& \shortstack{\textbf{PPV}\\\textbf{@ top-5\%}}
& \shortstack{\textbf{Prec}\\\textbf{@ top-100}} \\
\midrule
\endfirsthead

\multicolumn{9}{l}{\textit{continued from previous page}} \\
\toprule
\textbf{Condition} & \textbf{Hz.} & \textbf{Prev.\,(\%)}
& \shortstack{\textbf{PPV}\\\textbf{@ Sn=0.5}}
& \shortstack{\textbf{Sens}\\\textbf{@ top-1\%}}
& \shortstack{\textbf{PPV}\\\textbf{@ top-1\%}}
& \shortstack{\textbf{Sens}\\\textbf{@ top-5\%}}
& \shortstack{\textbf{PPV}\\\textbf{@ top-5\%}}
& \shortstack{\textbf{Prec}\\\textbf{@ top-100}} \\
\midrule
\endhead

\midrule
\multicolumn{9}{r}{\textit{continued on next page}} \\
\endfoot

\bottomrule
\endlastfoot

Dementia & 2y & 0.75
& \shortstack{0.0350\\{[0.0332, 0.0370]}}
& \shortstack{0.0920\\{[0.0774, 0.1069]}}
& \shortstack{0.0692\\{[0.0582, 0.0803]}}
& \shortstack{0.2909\\{[0.2689, 0.3153]}}
& \shortstack{0.0441\\{[0.0408, 0.0479]}}
& \shortstack{0.1400\\{[0.0600, 0.2200]}} \\

 & 5y & 3.15
& \shortstack{0.1010\\{[0.0934, 0.1090]}}
& \shortstack{0.0617\\{[0.0471, 0.0779]}}
& \shortstack{0.2084\\{[0.1631, 0.2609]}}
& \shortstack{0.2231\\{[0.1956, 0.2533]}}
& \shortstack{0.1449\\{[0.1276, 0.1643]}}
& \shortstack{0.2100\\{[0.1300, 0.3500]}} \\

\addlinespace
Knee OA & 2y & 2.41
& \shortstack{0.0508\\{[0.0492, 0.0524]}}
& \shortstack{0.0422\\{[0.0365, 0.0483]}}
& \shortstack{0.0970\\{[0.0838, 0.1116]}}
& \shortstack{0.1586\\{[0.1482, 0.1689]}}
& \shortstack{0.0766\\{[0.0714, 0.0815]}}
& \shortstack{0.0800\\{[0.0300, 0.1500]}} \\

 & 5y & 10.49
& \shortstack{0.1859\\{[0.1790, 0.1941]}}
& \shortstack{0.0314\\{[0.0261, 0.0376]}}
& \shortstack{0.2944\\{[0.2477, 0.3430]}}
& \shortstack{0.1251\\{[0.1122, 0.1383]}}
& \shortstack{0.2714\\{[0.2469, 0.2969]}}
& \shortstack{0.3100\\{[0.1900, 0.4200]}} \\

\addlinespace
COPD & 2y & 2.19
& \shortstack{0.0447\\{[0.0431, 0.0464]}}
& \shortstack{0.0428\\{[0.0370, 0.0487]}}
& \shortstack{0.1023\\{[0.0890, 0.1157]}}
& \shortstack{0.1597\\{[0.1485, 0.1705]}}
& \shortstack{0.0730\\{[0.0679, 0.0783]}}
& \shortstack{0.1600\\{[0.0900, 0.2600]}} \\

 & 5y & 8.93
& \shortstack{0.1541\\{[0.1468, 0.1616]}}
& \shortstack{0.0306\\{[0.0247, 0.0366]}}
& \shortstack{0.2826\\{[0.2343, 0.3326]}}
& \shortstack{0.1204\\{[0.1083, 0.1331]}}
& \shortstack{0.2152\\{[0.1950, 0.2352]}}
& \shortstack{0.3500\\{[0.2300, 0.4600]}} \\

\addlinespace
CHF & 2y & 1.09
& \shortstack{0.0665\\{[0.0636, 0.0696]}}
& \shortstack{0.1228\\{[0.1090, 0.1362]}}
& \shortstack{0.1362\\{[0.1210, 0.1511]}}
& \shortstack{0.3692\\{[0.3499, 0.3898]}}
& \shortstack{0.0818\\{[0.0772, 0.0865]}}
& \shortstack{0.2700\\{[0.1600, 0.3802]}} \\

 & 5y & 4.16
& \shortstack{0.1677\\{[0.1555, 0.1794]}}
& \shortstack{0.0932\\{[0.0763, 0.1124]}}
& \shortstack{0.3501\\{[0.3008, 0.4073]}}
& \shortstack{0.2721\\{[0.2429, 0.3031]}}
& \shortstack{0.2261\\{[0.2045, 0.2491]}}
& \shortstack{0.3400\\{[0.2400, 0.4600]}} \\

\addlinespace
Acute MI & 2y & 0.84
& \shortstack{0.0296\\{[0.0282, 0.0312]}}
& \shortstack{0.0559\\{[0.0458, 0.0675]}}
& \shortstack{0.0451\\{[0.0373, 0.0547]}}
& \shortstack{0.2485\\{[0.2262, 0.2702]}}
& \shortstack{0.0428\\{[0.0389, 0.0468]}}
& \shortstack{0.0400\\{[0.0100, 0.0900]}} \\

 & 5y & 3.13
& \shortstack{0.0815\\{[0.0748, 0.0880]}}
& \shortstack{0.0309\\{[0.0216, 0.0420]}}
& \shortstack{0.0922\\{[0.0662, 0.1232]}}
& \shortstack{0.1641\\{[0.1390, 0.1887]}}
& \shortstack{0.1033\\{[0.0882, 0.1187]}}
& \shortstack{0.1200\\{[0.0400, 0.2200]}} \\

\addlinespace
Pancreatic Cancer & 2y & 0.05
& \shortstack{0.0010\\{[0.0008, 0.0012]}}
& \shortstack{0.0490\\{[0.0196, 0.0863]}}
& \shortstack{0.0025\\{[0.0010, 0.0044]}}
& \shortstack{0.1818\\{[0.1119, 0.2606]}}
& \shortstack{0.0018\\{[0.0011, 0.0027]}}
& \shortstack{0.0100\\{[0.0000, 0.0300]}} \\

 & 5y & 0.15
& \shortstack{0.0025\\{[0.0016, 0.0036]}}
& \shortstack{0.0610\\{[0.0000, 0.1539]}}
& \shortstack{0.0086\\{[0.0000, 0.0220]}}
& \shortstack{0.1341\\{[0.0543, 0.2400]}}
& \shortstack{0.0042\\{[0.0018, 0.0073]}}
& \shortstack{0.0200\\{[0.0000, 0.0600]}} \\

\addlinespace
Prostate Cancer & 2y & 0.32
& \shortstack{0.0207\\{[0.0189, 0.0224]}}
& \shortstack{0.0970\\{[0.0756, 0.1190]}}
& \shortstack{0.0314\\{[0.0247, 0.0385]}}
& \shortstack{0.3427\\{[0.3067, 0.3773]}}
& \shortstack{0.0222\\{[0.0197, 0.0247]}}
& \shortstack{0.0300\\{[0.0000, 0.0800]}} \\

 & 5y & 1.32
& \shortstack{0.0712\\{[0.0634, 0.0799]}}
& \shortstack{0.0777\\{[0.0525, 0.1064]}}
& \shortstack{0.0934\\{[0.0634, 0.1260]}}
& \shortstack{0.2924\\{[0.2449, 0.3402]}}
& \shortstack{0.0772\\{[0.0655, 0.0903]}}
& \shortstack{0.1400\\{[0.0500, 0.2300]}} \\

\addlinespace
\end{longtable}
\renewcommand{\arraystretch}{1.0}
}

\newpage

\subsection*{\rev{Inference compute scaling of next-token baselines}}

\begin{figure}[h]
    \centering
    \includegraphics[width=\textwidth]{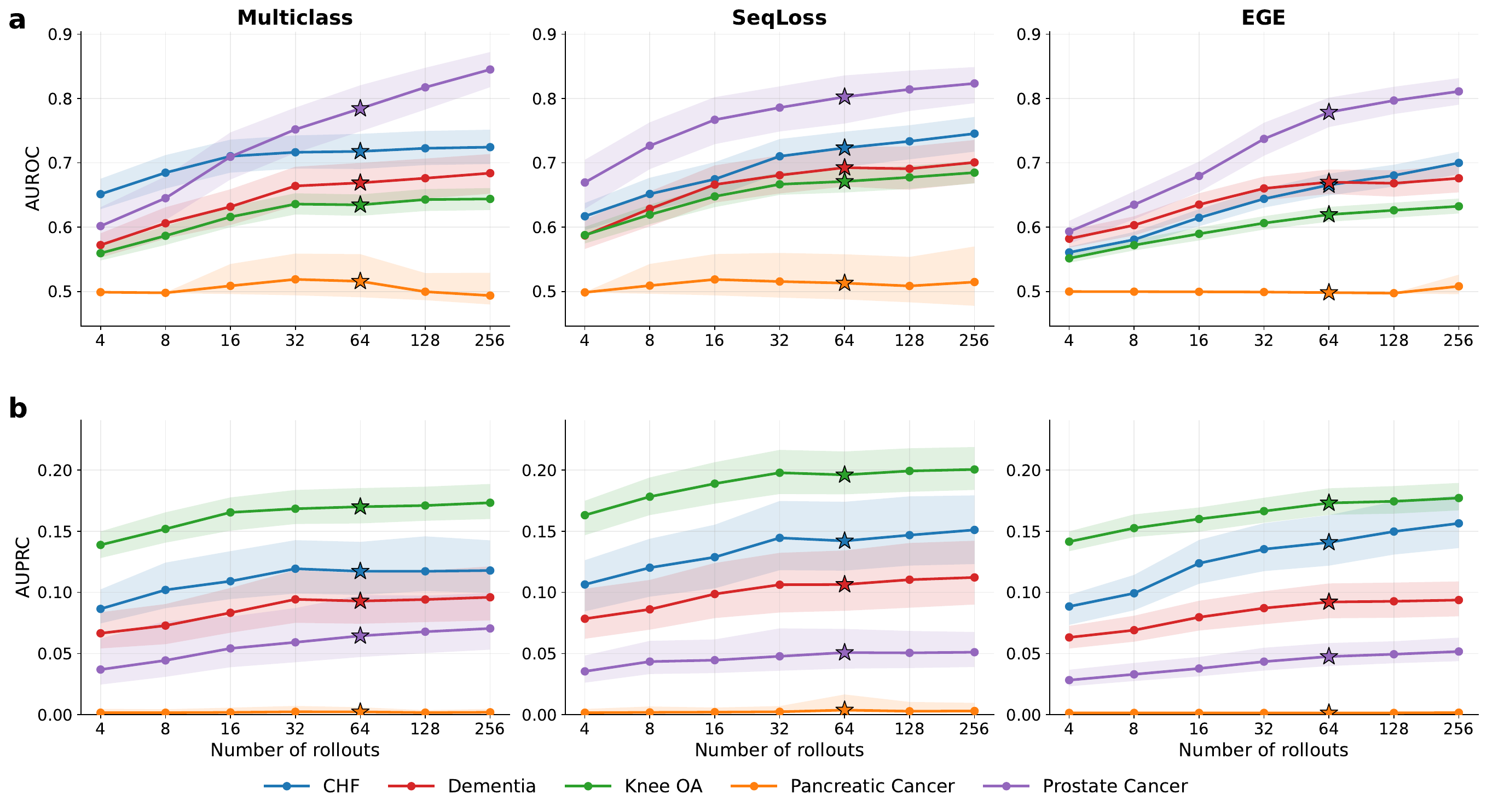}
    \caption{
    \textbf{Rollout-budget sensitivity for the simulation-based models.}
    We evaluated the Multiclass, SeqLoss, and EGE next-token baselines using increasing numbers of simulated rollouts at inference. Panels show AUROC (\textbf{a}) and AUPRC (\textbf{b}) across five representative prediction tasks on the 5-year horizon. Shaded bands are 95\% patient-clustered bootstrap confidence intervals, and stars mark the $R=64$ setting used in the main experiments. Performance generally improves as the rollout budget increases from very small values, but gains diminish substantially by 64 rollouts for most settings. These results suggest that the performance of the next-token baselines is not primarily limited by the rollout budget used in the main comparison.
    }
    \label{fig:rollout_ablation}
\end{figure}

\rev{The simulation-based baselines estimate horizon-level disease risk by sampling $R$ future trajectories per patient window, whereas RAVEN produces a horizon-conditioned score in a single forward pass; their predictions may depend on the number of Monte Carlo rollouts used at inference. To assess whether the observed performance gap is driven primarily by limited sampling, we performed a rollout-budget sensitivity analysis for the Multiclass, SeqLoss, and EGE next-token baselines. We evaluated performance across increasing sampling budgets from 4 to 256 rollouts for five representative outcomes at the 5-year horizon, including both common chronic conditions and rarer diseases (Fig. \ref{fig:rollout_ablation}).} 

\rev{Across baselines and outcomes, AUROC and AUPRC generally improved with additional rollouts but largely plateaued by 32 to 64 rollouts, suggesting that further increasing inference compute provides limited additional gains. This saturation behavior agrees with independent evidence at a larger scale: in a study of simulation-based medical-event models on Epic Cosmos, Waxler et al.\cite{waxler2025generative} report that performance from aggregating Monte Carlo generations plateaus near 64 simulations per patient. Critically, no baseline closes the per-condition gap to RAVEN at any budget we tested. Even at $R{=}256$, the strongest baseline remains below RAVEN on every condition. These results indicate that although very small rollout budgets can underestimate the performance of simulation-based next-token baselines, increasing the budget beyond 64 rollouts yields limited additional improvement in most settings.}

\newpage

\subsection*{Zero-shot evaluations of large language models}

For the MedGemma-27B zero-shot baseline, we serialized each patient history as JSON and prompted the model to predict whether a target condition would occur within a prespecified prediction horizon. The prompt explicitly specified the target condition, the prediction window, and a required JSON output schema so that generations could be parsed into a binary prediction and a confidence score. For each evaluation example, we truncated the serialized history to fit within the model context window by retaining the most recent portion of the record. We then sampled $K=50$ generations per example using temperature sampling ($T=0.7$, top-$p=0.95$) and averaged the parsed confidence values across valid generations to obtain the final risk score. In preliminary validation experiments, we compared two prompting strategies for MedGemma models: direct confidence prediction for the target condition and full future patient-history generation followed by extraction of the target outcome. We evaluated these strategies using MedGemma 1.5 4B and MedGemma 27B, and found that MedGemma 27B with the direct confidence aggregation approach yielded the strongest validation performance on the prostate cancer task. We therefore report this configuration as the primary prompted LLM baseline.

The template below shows the exact structure used for prompting, with placeholders for the patient identifier, target condition, prediction horizon, serialized patient history, and condition-specific JSON key.

\begin{lstlisting}[]
<s>[SYSTEM_PROMPT]
You are an information extraction and forecasting engine for clinical risk prediction. 
Your task is to predict the risk of <CONDITION_NAME> based on patient history.

You MUST follow these rules exactly:
1) Output MUST be EXACTLY one JSON object, and NOTHING else.
2) The JSON MUST have exactly this structure:
{
  "<PATIENT_ID>": {
    "reason": "<brief_1_to_2_sentence_summary>",
    "confidence": <float_0_to_1>,
    "<JSON_KEY>": "<Y_or_N>"
  }
}
3) CRITICAL: The "confidence" value MUST represent the numerical probability P(Y)
   of the condition occurring (e.g., 0.85 for high risk, 0.12 for low risk).
4) Evaluate the patient's records and determine if they will develop 
   <CONDITION_NAME> within the <HORIZON_TEXT>.
5) Provide a concise reason for your prediction based on the evidence in the records.
[/SYSTEM_PROMPT]
[PATIENT_HISTORY]
<PATIENT_HISTORY_JSON>
[/PATIENT_HISTORY]
[QUERY]
Predict the risk of <CONDITION_NAME> for this patient in the <HORIZON_TEXT> 
based on the records provided above.
[/QUERY]
[OUTPUT]
\end{lstlisting}

Here, \texttt{<CONDITION\_NAME>} was replaced with the full disease name (for example, ``Heart Attack'' or ``Dementia''), \texttt{<HORIZON\_TEXT>} with the corresponding prediction window (for example, ``NEXT 730 DAYS'' or ``NEXT 1825 DAYS''), and \texttt{<JSON\_KEY>} with the task-specific output field used for parsing. The model was required to return both a binary prediction (\texttt{Y} or \texttt{N}) and a confidence score in $[0,1]$. Unparseable generations were discarded, and rule-based fallbacks were used when needed to extract the predicted label and confidence from imperfect outputs.

\end{document}